\documentclass[11pt]{article}
\usepackage[latin1]{inputenc}

\usepackage[T1]{fontenc}

\usepackage{amsmath,amssymb,amsthm,mathrsfs}
\usepackage{dsfont}
\usepackage[english]{babel}
\usepackage{graphicx,color}
\usepackage{subfigure,enumitem}
\usepackage[colorlinks]{hyperref}

\hypersetup{
linkcolor=blue,
citecolor=blue,
}
\usepackage[numbers,sort]{natbib}
\usepackage[ruled,vlined]{algorithm2e} %
\usepackage{siunitx} %

\graphicspath{{Figures/}}

\def\argmin{\mathop{\rm arg \; min}\limits}%
\def\argmax{\mathop{\rm arg \; max}\limits}%

\theoremstyle{plain}
\newtheorem{prop}{Proposition}[section]

\newtheorem{fact}{Fact}[section]

\theoremstyle{definition}
\newtheorem{defin}{Definition}[section]
\newtheorem{exe}{Example}[section]

\newtheorem{pf}{Proof}

\newcommand{\E}{{\mathbb E}}
\newcommand{\R}{\mathrm{I\!R}}

\newcommand{\Z}{{\mathbb Z}}

\newcommand{\bL}{{\mathbf L}} %
\newcommand{\bPhi}{{\mathbf \Phi}} %
\newcommand{\bLambda}{{\mathbf \Lambda}} %
\newcommand{\bSigma}{{\mathbf{\Sigma}}} %

\newcommand{\Cov}{\mathrm{Cov}} %

\newcommand{\GG}{\ensuremath{\mathcal G}}
\newcommand{\FFtheo}{\ensuremath{\mathcal F}}
\newcommand{\FF}{\ensuremath{\mathbf F}} %
\newcommand{\X}{\ensuremath{\mathcal X}} %
\newcommand{\Y}{\ensuremath{\mathcal Y}} %

\newcommand{\bx}{\boldsymbol{x}}

\renewcommand{\bL}{\boldsymbol{L}}
\newcommand{\bK}{\boldsymbol{K}}

\renewcommand{\bLambda}{\ensuremath{\boldsymbol{\Lambda}} }
\renewcommand{\bPhi}{\ensuremath{\boldsymbol{\Phi}} }

\def\Id{\mathrm{Id}}

\usepackage{multirow}
\usepackage{tabularx,ragged2e,booktabs,caption}
\newcolumntype{C}[1]{>{\Centering}m{#1}}

\numberwithin{equation}{section}         

\title{Sensor selection on graphs via data-driven node sub-sampling in network time series}

\author{ Yiye Jiang$^{1,2}$, J\'{e}r\'{e}mie Bigot$^{1}$   \& Sofian Maabout $^{2}$\\
\\  $^{1}$Institut de Math\'ematiques de Bordeaux, Universit\'e de Bordeaux \vspace{0.1cm}  \\ $^{2}$Laboratoire Bordelais de Recherche en Informatique, Universit\'e de Bordeaux}

\date{\today}

\begin{document}

\maketitle

\thispagestyle{empty} 

\begin{abstract}
This paper is concerned by the problem of selecting an optimal sampling set of sensors over a network of time series  for the purpose of signal recovery at non-observed sensors with a minimal reconstruction error. The problem is motivated by applications where time-dependent graph signals are collected over redundant networks. In this setting, one may wish to only use a subset of sensors to predict data streams over the whole collection of nodes in the underlying graph. A typical application is the possibility to reduce the power consumption in a network of sensors  that may have limited battery supplies.  We propose and compare various data-driven strategies to  turn off a fixed number of sensors or equivalently to select a sampling set of nodes. We  also relate our approach to the existing literature on sensor selection from multivariate data with a (possibly) underlying graph structure. Our methodology combines tools from multivariate time series analysis, graph signal processing, statistical learning in high-dimension and deep learning. To illustrate the performances of our approach, we report  numerical experiments on the analysis of real data from bike sharing networks in different cities.
\end{abstract}


\section*{Acknowledgments}
J\'er\'emie Bigot is a member of Institut Universitaire de France (IUF), and this work has been carried out with financial support from the IUF. We also gratefully acknowledge Max Halford for providing the bike sharing datasets.

\section{Introduction}

\begin{figure}[h]
    \centering
    \includegraphics[width=0.4\textwidth]{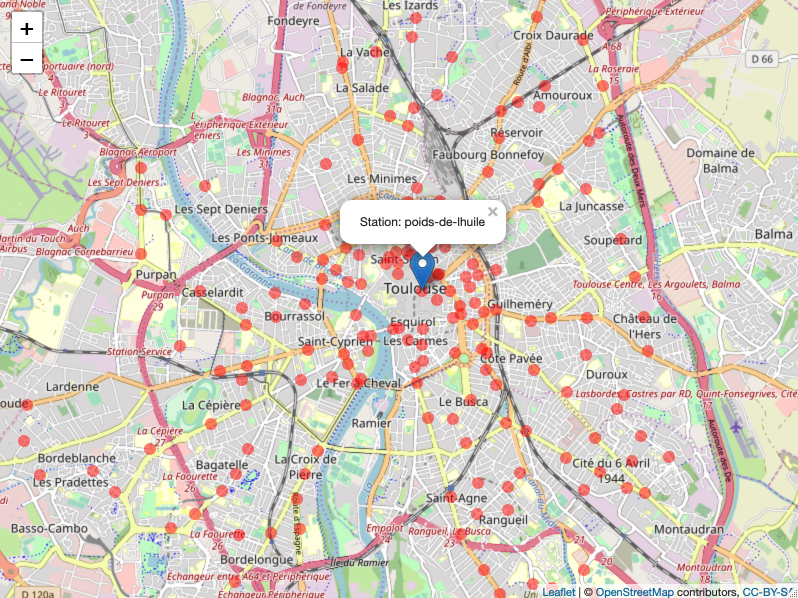}
    \includegraphics[width=0.55\textwidth]{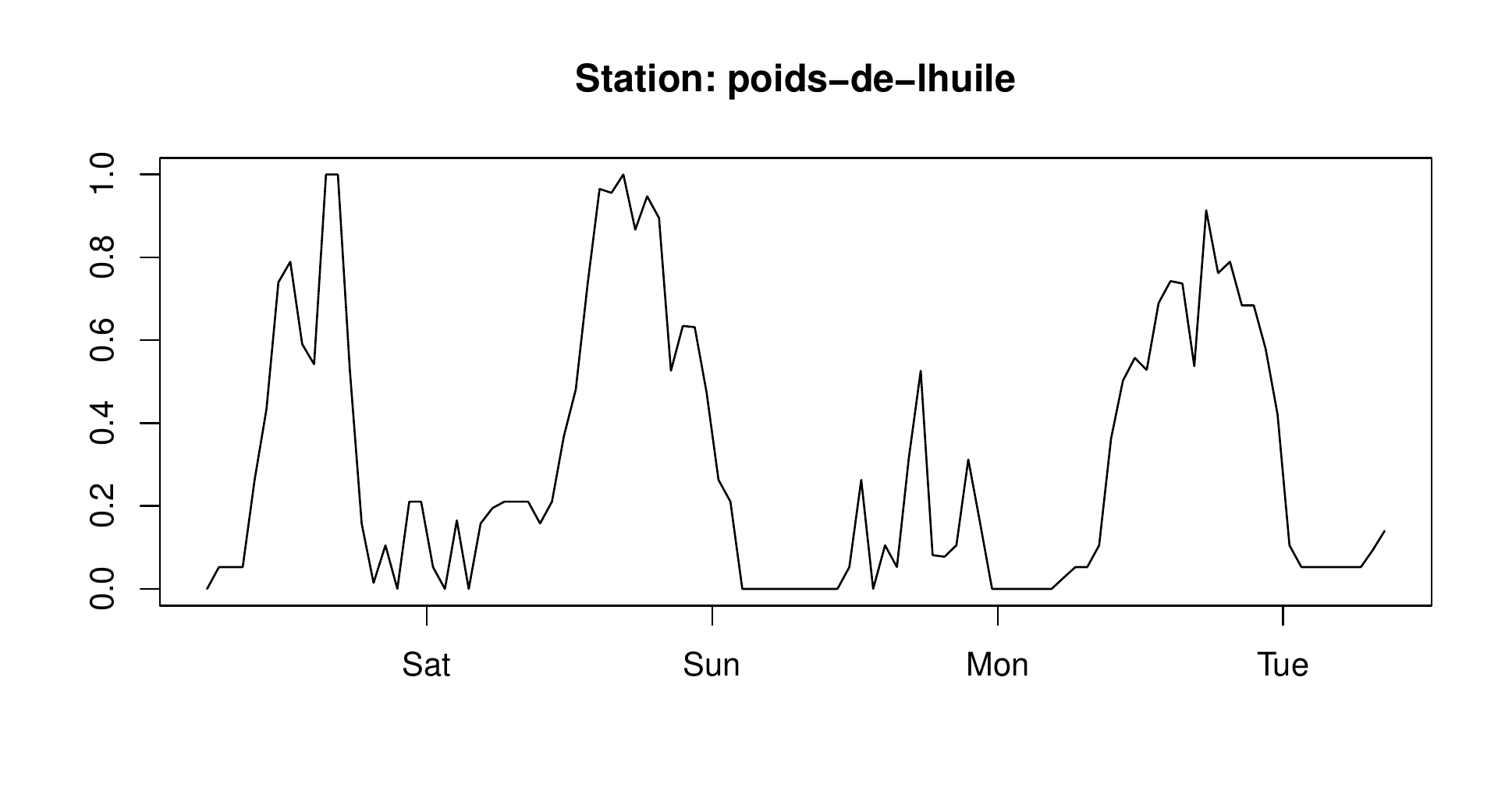}
    \caption{\small \textit{Bike-sharing sensor network of Toulouse city (France).} Time series component $x_{it}$ represents the ratio of bikes at sensor $i$, time $t$ (with display over a few days).}
    \label{fig:velib_toulouse}
\end{figure}

Data recorded over a network of sensors have become increasingly popular in recent years with applications in many areas such as traffic analysis \cite{CrovellaKolaczyk03}, functional brain imaging \cite{HuangBMBRV18}, social \cite{Sandryhaila2014},  or  transport networks \cite{OrtegaFKMV18}. Such data may generally be modeled as  multivariate time series $(x_{it})_{i,t}$ with an underlying graph structure whose nodes represent the sensors (e.g.\ spatial locations) where observations are recorded over time. For example Figure \ref{fig:velib_toulouse} shows data from the bike-sharing network in Toulouse city, which consists of $185$ nodes and $4305$ hours\ after data cleaning. At every bike station, a sensor has been installed to record the number of available bikes and spaces at regular time intervals.  Time series variable $x_{it}$ represents the ratio of bikes to the total number of dock at sensor $i$, time $t$. In this context, there is a natural underlying graph structure, which links two sensors based on a function of their geographical distance (more details on the construction of this graph are given in the section on numerical experiments).

The research fields of signal processing \cite{ShumanNFOV13} and statistical inference \cite{Kolaczyk2009} on graphs are thus currently very active. A reason for the success of these approaches is the ability of capturing relevant relational information from network data. As a consequence, in many situations, sub-groups of sensors exhibit strong relationships with each other which makes time series collected over networks redundant. In this setting, it is of interest to decide which sensors may be kept to approximate unobserved signals at other sensors that have been turned off.  A typical application is the possibility to reduce the power consumption by  node sub-sampling in a network of sensors  that may have limited battery supplies, while still maintaining a satisfactory data reconstruction at turned off sensors. This paper is concerned by the derivation of data-driven sampling schemes to turn off a fixed number of sensors in a network (and thus selecting an optimal subset of sensors to collect data)  for the purpose of signal recovery at non-observed nodes with a minimal reconstruction error.

\subsection{Methodology of sensor selection and main contributions}\label{sec:notation}

Let us now formulate more precisely the task of defining a subset of nodes in a network of sensors to be turned off while still having a good reconstruction of the signals at these sensors from the use of the data at the remaining sensors. 

First of all, we represent a network of sensors as a graph $\mathcal{G} = \{\mathcal{N}, \mathcal{E}, A\}$ consisting of a finite set of nodes (or vertices) $\mathcal{N} = (V_i)_i$ with $|\mathcal{N}| = N$, a set of edges $\mathcal{E} = (i,j)_{i,j}$, and an adjacency matrix $A = (a_{ij})_{i,j}$.
We are interested in analyzing signals which are recorded at the nodes of an undirected, connected, and weighted graph. Therefore, the adjacency matrix $A$ is supposed to be symmetric and  its $(i,j)$-th entry satisfies: $a_{ij} > 0$ \underline{iff} $(i,j) \in \mathcal{E}$; $a_{ij} = 0$ otherwise. Note that if $a_{ij}$ is a strictly decreasing function of the geographical distance between spatial sensors,  then $(i,j)$ necessarily belongs  to $\mathcal{E}$ since, in this case, the graph is presumably complete.  In our numerical experiments, the graph can be sparse as we only connect   each sensor to its $k$ nearest neighbours.

A key hypothesis of our approach is to assume that observations of time-dependent signals $\mathbf{x}_t$ on the graph $\mathcal{G}$ are available for a sufficiently large number of time points $1 \leq t \leq T_0$. At each time $t$, a {\it graph signal} on $\mathcal{G}$ is defined as a mapping $x_t : \mathcal{N} \to \R$ with $x_t(V_i)$ representing the observation at time $t$ and node $V_i$. The collection of data $(\mathbf{x}_t)_{1 \leq t \leq T_0}$ is a multivariate time series that we shall also refer to as a {\it network time series}. Equivalently the signal $\mathbf{x}_t$ may be represented as a vector $\mathbf{x}_t = (x_{1t}, x_{2t}, ..., x_{Nt}) \in \R^N$ with $x_{it} = x_t(V_i)$. Then, after time $T_0$, we wish to turn off a subset $\{ V_i, i \in I\}$ of sensors with given cardinality $|I| = p > 0$, and to reconstruct as accurately as possible its observations over time $T_0 < t \leq T_1$ using the ones from remaining sensors indexed by $I^c$, where $I^c = \mathcal{N}  \backslash I$. We call the observations until $T_0$ the historical data and those after $T_0$ the current data. We denote $\mathbf{x}_{I,t} = (x_{it})_{i \in I}$ and $\mathbf{x}_{I^c,t} = (x_{it})_{i \in I^c}$. 

In practice, we pre-process the data by first extracting their trend, then subtracting the trend from the original data, so that the input of models are considered as stationary time series.
Then, after a preliminary detrend pre-processing step of $\mathbf{x}_{t}$,  the first step in our procedure of sensor selection is to choose a {\it methodology of reconstruction} that is defined as a parametric class of functions
$$
f_{\Theta}^\mathcal{G}  : \R^q \to \R^p, \quad \mbox{where} \quad q = (N-p)(H+1), 
$$
indexed by a set of real parameters $\Theta$ with dimension $d$, representing the degree of flexibility of $f_{\Theta}$. 
The values of  $f_{\Theta}^\mathcal{G}$ may depend on the graph $\GG$ and data from the past up to time lag $H \geq 0$, with $H \ll T_0$. The parameter $H$ represents the amount of past information to be used for signal recovery at unobserved nodes. For instance, $H=0$ means that we want to recover unseen values from just the most recent ones, i.e., those recorded at a given $t > T_0$. We denote by
\begin{equation}
\mathbf{x}_{I^c,t}^H = (\mathbf{x}_{I^c,t}, \mathbf{x}_{I^c,t-1},\dots, \mathbf{x}_{I^c,t - H}) \label{eq:defxH}
\end{equation}
the $\R^q$ vector containing the observed signals from time $t-H$ to $t$.

The simplest class of reconstruction methods is the linear one with $H=0$ that does not make use of the graph $\mathcal{G}$. This corresponds to the setting where 
\begin{equation} \label{eq:reclinH0}
f_{\Theta}^\mathcal{G} = f_{\Theta}, \quad \mbox{with} \quad f_{\Theta}(\bx) = \Theta \bx, \quad \mbox{where} \quad \bx \in \R^{N-p} \mbox{ and } \Theta \in \R^{p \times (N-p)}.
\end{equation}
In a second step (once the class of functions $f_{\Theta}^\mathcal{G} $ is chosen), given a subset $I$ of fixed cardinality $|I| = p$, one trains $f_{\Theta}^\mathcal{G}$  by minimizing the reconstruction error on all historical data $\mathbf{x}_t, t  \leq T_0$ as follows
\begin{equation}\label{eq:recpb}
\FF(I,\hat{\Theta}(I)) = \min_{\Theta \in \R^d} \FF(I,\Theta), \quad \mbox{where}  \quad \FF(I,\Theta) := \frac{1}{T_0} \sum_{t = H + 1}^{T_0}  \|\mathbf{x}_{I,t}  - f_{\Theta}^\mathcal{G}( \mathbf{x}_{I^c,t}^{H}) \|_{\ell_2}^2,
\end{equation}
where $\| \cdot \|_{\ell_2}$ denotes the usual Euclidean norm.
For the example \eqref{eq:reclinH0} of linear reconstruction functions, the minimization problem \eqref{eq:recpb} simplifies to
\begin{equation}\label{eq:recpbLin}
\min_{\Theta \in \R^{p \times (N-p)}} \frac{1}{T_0} \sum_{t = 1}^{T_0}  \| \mathbf{x}_{I,t} - \Theta \mathbf{x}_{I^c,t} \|_{\ell_2}^2,
\end{equation}

Now, assuming that $\hat{\Theta}(I) \in \R^d$ denotes a minimizer of problem \eqref{eq:recpb}, the prediction at time $t > T_0$ at unobserved nodes is given by 
$$
\hat{\mathbf{x}}_{I,t} = \left[  f_{\hat{\Theta}(I)}^\mathcal{G}(\mathbf{x}_{I^c,t}^{H}) \right]_{t},
$$
to which the previously estimated trend may finally be added.
The third step is  to select a subset $\hat{I}$ of cardinality $p$ minimizing the reconstruction error of the historical data that is
\begin{equation}\label{eq:NPpb}
\hat{I} = \argmin_{I \subset \mathcal{N}  \; : \; |I| = p}  \FF(I,\hat{\Theta}(I)).
\end{equation}
In this paper, we consider three popular classes of reconstruction methods, which are \textit{linear regression}, \textit{kernel regression over graphs}, and \textit{graph convolutional neural network}. 
For the first two classes, minimizing the problem \eqref{eq:NPpb} is a delicate combinatorial problem for moderate values of $p$ and $N$ as some its simplest instance using a linear reconstruction approach is known to be NP-complete \cite{Krause2008}. Hence, we shall introduce greedy strategies to approximate a solution to \eqref{eq:NPpb} which consists in considering $p=1$ to select the best node $i_{(1)}$ from $\mathcal{N}$, then to find $i_{(2)}$ from $\mathcal{N}\setminus \{i_{(1)}\}$, and so on. Note that if $p$ and $N$ are both small, it is possible to solve the problem exactly which can be used as a reference to compare exact and greedy strategies in terms of execution time and accuracy. However, the computational cost of an exact approach becomes quickly prohibitive as $p$, $N$ increase.

To circumvent the use of a greedy strategy for the class of graph convolutional neural  network (GCN), we propose to adapt standard neural networks (that are typically used for prediction or classification) to the setting of sensor selection as considered in this paper. We shall refer to such neural networks as {\it selection networks}. One of these networks (that is presented in Subsection \ref{sec:slcNetDP}) is based on the dropout technique from deep learning. Compared to the other selection methods considered in this paper, the direct output of such a GCN (using the dropout method) is not the optimal set $\hat{I}$, but rather a scoring of each sensor which quantifies the  {\it learned predictability} of each node in the graph for the purpose of sensor selection. Some examples of such a scoring  are shown in Figure \ref{fig:learned representation of predictability} and Figure \ref{fig:learned representation of predictability, toulouse} for bike-sharing networks in the cities of Paris and Toulouse. Then, based on the scores of the sensors, it is possible to select $p$ out of $N$ sensors to be turned off, and to train a GCN using as input the remaining sensors and as output the ones that have been removed.

Finally, all of our methods provide not only the selected $p$ sensors, but also their priority to be turned off, denoted by $\hat{I} = \{i_{(1)}, i_{(2)},..., i_{(p)}\}$. Intuitively, the semantic of this priority is that the mean reconstruction accuracy of signals from the sensors indexed by $\{i_{(1)}, \dots, i_{(j)}\}$ is better than that of $\{i_{(1)},\dots, i_{(j)},\dots, i_{(k)}\}$. Moreover, the GCN approach using dropout also implies that the reconstruction accuracy from node $i_{(j)}$ is better than that of node $i_{(j+k)}$.

\subsection{Related literature}

Various algorithms for selecting sensor locations have  been developed in statistics and machine learning to collect observations from a stochastic process. In particular, sensor selection is a well-understood problem in spatial statistics \cite{Cressie93} when spatial phenomena are modeled as multivariate Gaussian  processes. In this setting, many strategies have been proposed to select sampling sensors at the most informative locations such as placing them at locations of highest entropy \cite{Shewry87}, or maximizing the mutual information between the selected sensors and those which are not selected \cite{Krause2008}. These approaches lead to greedy strategies to add sensors one-by-one that are discussed in Section \ref{sec:linest} and Section \ref{sec:kernel}. For further references and discussion on the sensor selection problem beyond  the  case of Gaussian process modeling in spatial statistics we refer to \cite{JoshiB09} where a convex relaxation of such problems is studied. Note that these works do not consider the setting of multivariate time series. Moreover, they do not incorporate any underlying graph structure in the selection of the sensors. 

For graph signals (not depending on time), various approaches have been developed for selecting a sampling set of nodes. They can be classified into deterministic and random methods which are both based on Fourier analysis and sampling theory for graph signals. Deterministic approaches aim at selecting sensors one-by-one such that a cost function is maximized at each step, see e.g.\ \cite{sakiyama2019eigendecomposition,AnisGO16,ChenVSK15}, whereas random methods select sensors according to some probability distributions on the nodes of the graph \cite{puy:hal-01229578,perraudin2018}. For a recent overview of optimal  sampling for graph signals and its connection to sensor selection using Gaussian processes in spatial statistics we refer to \cite{sakiyama2019eigendecomposition}. However, none of these approaches considers the analysis of time-dependent graph signals. Moreover, the resulting  algorithms for nodes selection are not data-driven in the sense that they only depend on the structure of the graph and smoothness assumptions of the signal to be reconstructed.

Sensor selection from time series observed on the nodes of a graph has recently been considered in \cite{AggarwalBS17} in the framework of information networks. The setting in \cite{AggarwalBS17} is somewhat different from the one in this paper as further information is added to the graph structure. In  \cite{AggarwalBS17}  each node of the network is enriched with additional informations such as cost of selection, importance and prediction error. Moreover, each edge $(i,j)$ is associated to a weight depending on the error of predicting sensor $v_j$ from sensor $v_i$  and their importance. The method of predicting one time series at given vertex is based on a linear regression model using  time series at linked vertices as explanatory variables. Finally, the overall procedure in \cite{AggarwalBS17}  allows to incorporate a budget constraint in the selection of most critical sensors. The final criterion to be optimized in \cite{AggarwalBS17}  is shown to lead to a NP-hard problem, and greedy strategies are proposed that progressively add sensors one-by-one.
By contrast, our sensor selection criterion takes into account both time series dependency and graph structure, and it is based on more flexible reconstruction methods.

\subsection{Organization of the paper}

In Section \ref{sec:linest} and Section \ref{sec:kernel}, we consider the linear regression and graph kernel approaches as reconstruction methods, respectively. We interpret the associated sensor selection criteria  through the prism of mutlivariate time series modeling, and we derive the corresponding greedy algorithms. In Section \ref{sec:GCN}, we first give a brief review on  graph convolutional neural networks (GCN). ChebNet, that is one of the most popular GCN for graph-level tasks, is presented with more details. Then, we elaborate the proposed approaches of adapting prediction networks to the data-driven sensor selection problem considered in this work. Lastly, in Section \ref{sec:num}, we evaluate all the sensor selection strategies on real bike-sharing datasets. All proofs are deferred to a technical appendix.

\section{Linear estimators and minimization of partial variance} \label{sec:linest}

Throughout this section, for a given $H \geq 0$, we consider the class of {\it linear estimators} that do not use the underlying graph structure $\GG$. More formally, we consider the class of functions $f_{\Theta}^\mathcal{G} = f_{\Theta}$ defined as
\begin{equation}\label{eq:reclinH}
\left\{
\begin{array}{ccl}
f_{\Theta} : \R^{q} & \to & \R^p, \quad \mbox{where} \quad q = (N-p)(H+1) \\
\bx & \mapsto & \Theta\bx.
\end{array}
\right.
\end{equation}

First of all, we recall the notion of \textit{partial variance}, a well known statistical quantity, that allows to understand how the sensors to be turned off are selected when using  linear estimators.  
Let $\mathbf{x} = (x_1, ..., x_n)$ be a random vector with mean $\mathbf{0}$ and covariance $\bSigma$. We partition $\mathbf{x}$ into two disjoint components, denoted by $\mathbf{x}_A, \mathbf{x}_B$ respectively, where $A, B \subsetneqq \{1, 2, ..., n\}$ are two disjoint sets such that  $A\cup B = \{1, 2, ..., n\}$. Then, we rearrange the rows and columns of $\bSigma$, so that it is the covariance matrix of the random vector $(\mathbf{x}_A, \mathbf{x}_B)$, which we still denote by $\bSigma$, namely
$$
\bSigma = 
    \begin{pmatrix}
    \mathbf{\Sigma_{A}} & \mathbf{\Sigma_{AB}} \\
    \mathbf{\Sigma_{BA}}&\mathbf{\Sigma_{B}}
    \end{pmatrix}.
$$
The \textit{partial variance-covariance matrix} of $\mathbf{x}_A$ given $\mathbf{x}_B$ is defined as (assuming that $\mathbf{\Sigma_{B}}$ is not singular)
\begin{equation}
    \mathbf{\Sigma_{A|B}} = \mathbf{\Sigma_{A}} -  \mathbf{\Sigma_{AB}}\mathbf{\Sigma_{B}}^{-1}\mathbf{\Sigma_{BA}}.
\end{equation}
The diagonal entry of $\mathbf{\Sigma_{A|B}}$ which corresponds to the random variable $x_{i}, i \in A$, is the partial variance of $x_{i}$ given $\mathbf{x}_B$, denoted by $\sigma^2_{i|B}$.  Partial variance can be understood through the following steps. Firstly, regress the variables $\mathbf{x}_A$ on $\mathbf{x}_B$ through the linear regression model $\hat{\mathbf{x}}_A(\mathbf{x}_B) = \E(\mathbf{x}_A|\mathbf{x}_B) = \Theta \mathbf{x}_B$\footnote{Note that, the true conditional expectation $\E(\mathbf{x}_A|\mathbf{x}_B)$ may not be equal to $\Theta \mathbf{x}_B$. In other words, in this case the linear estimator is not the best one. However, such equality holds for several joint distributions of $(\mathbf{x}_A, \mathbf{x}_B)$, for example, multivariate normal and multinomial \cite{2019arXiv190909681O, baba2004partial}.}. Secondly, minimize the \textit{mean squared error (MSE)} of regression to get the best parameter $\Theta^*$, which reads as
$$
\argmin_{\Theta} \E\|\mathbf{X}_A - \Theta \mathbf{X}_B\|^2_{l_2}.
$$
Setting the derivative w.r.t.\ $\Theta$ to zero\footnote{To make sure the derivative and the expectation is interchangeable, we add a technical assumption here, which is $\mathbb{E}|(\mathbf{x}_{A} - \Theta \mathbf{x}_{B})\mathbf{x}_{B}^t| < \infty$, for all $\Theta$. In the rest of this paper, each time we need to exchange these two operations, this assumption is added by defaut.}, leads to $\Theta^* = \mathbf{\Sigma_{AB}}\mathbf{\Sigma_{B}}^{-1}$. 
Lastly, because the regression residual now writes as $\mathbf{x}_A - \Theta^*\mathbf{x}_B = \mathbf{x}_A - \mathbf{\Sigma_{AB}}\mathbf{\Sigma_{B}}^{-1}\mathbf{x}_B$, it is easy to find that the partial covariance $\mathbf{\Sigma_{A|B}}$ is exactly the covariance of residual, and that the minimal value of the problem above equals $\sum_{i \in A} \sigma^2_{i|B}$.
Thus, the partial covariance of $\mathbf{x}_A$ given $\mathbf{x}_B$ can be considered as the covariance among residuals of projections of $\mathbf{x}_A$ onto the linear space spanned by $\mathbf{x}_B$. 
Furthermore, the partial variance $\sigma^2_{i|B}$ reflects the linear explanatory ability of predictors $\mathbf{x}_B$ to $x_{i}$. A lower partial variance $\sigma^2_{i|B}$ reflects a higher explanatory ability of $\mathbf{x}_B$, which leads to a smaller reconstruction error of $x_{i}$.

We are now ready to present our method of sensor selection when the underlying graph structure of the network of sensors is not used, and the class of linear estimators is employed. We first analyze the simplest case where the data is temporally uncorrelated  and $H = 0$. Then, we move onto weakly stationary time series modeling with $H > 0$. In both cases, we start from a population approach analysis, and we end up with the empirical criteria for the best turned-off set $\hat{I}$. We formulate the greedy algorithms accordingly.

\subsection{The case $H=0$}\label{sec:linH=0}
We first assume the observed data is uncorrelated along time, and that it is drawn from the population distribution of a random vector $\mathbf{x} = (x_1,x_2,...,x_N)$, where $x_i$ represents the signal on sensor $i$. For simplicity, we assume that the mean of $\mathbf{x}$ is $\mathbf{0}$. Given a subset of indices $I$ with $|I| = p$, we re-arrange entries of the vector $\mathbf{x}$ as $(\mathbf{x}_{I}, \mathbf{x}_{I^c})$. We denote the covariance matrices $\Cov(\mathbf{x}_{I}), \Cov(\mathbf{x}_{I^c})$ and $ \Cov(\mathbf{x}_{I} \, , \,  \mathbf{x}_{I^c})$ by $\bSigma_{I}, \bSigma_{I^c}$ and $\bSigma_{II^c}$.

We now consider the reconstruction function (\ref{eq:reclinH0}), which gives the recovered signals $\hat{\mathbf{x}}_{I^c,t} = \Theta\mathbf{x}_{I^c}$.
The best parameter $\Theta^*(I)$ is given by the minimizer of
\begin{equation}\label{eq:recpbLintheo}
    \FFtheo(I,\Theta^*(I)) = \min_{\Theta \in \R^{p \times (N-p)}} \mathbb{E}\|\mathbf{x}_I - \Theta \mathbf{x}_{I^c}\|^2_{l_2},
\end{equation}
where $\FFtheo(I,\Theta)$ is the \textit{theoretical (or population) reconstruction error} for a given set $I$ and parameter value $\Theta$. From the above preliminaries, we know that $\Theta^*(I) = \bSigma_{II^c}\bSigma_{I^c}^{-1}$ and
$\FFtheo(I,\Theta^*(I)) = \sum_{i \in I} \sigma^2_{i.I^c}$.
Therefore, the best turned-off set $I$ is given by 
\begin{equation}\label{eq:bestsetlinH=0_theo}
    \argmin_{I \subset \mathcal{N}  \; : \; |I| = p}  \FFtheo(I,\Theta^*(I)) = \argmin_{I \subset \mathcal{N}  \; : \; |I| = p}\sum\limits_{i \in I} \sigma^2_{i.I^c} = \mathop{\rm min}\limits_{I \subset \mathcal{N}  \; : \; |I| = p} tr(\bSigma_I - \bSigma_{II^c}\bSigma_{I^c}^{-1}\bSigma_{I^cI}).
\end{equation}
This procedure aims to find the set of most predictable variables, measured by their partial variances.
It is easy to find that problem (\ref{eq:recpbLintheo}) is consistent with problem (\ref{eq:recpbLin}), and we shall refer to the quantity $\FF(I,\Theta)$ defined in \eqref{eq:recpb} as the \textit{empirical reconstruction error}.
As soon as we replace the submatrices of covariance $\bSigma$ with sample covariance $\hat{\bSigma} = \frac{1}{T_0} \sum_{t = 1}^{T_0} \mathbf{x}_{t}\mathbf{x}_{t}^t$ in $\FFtheo(I,\Theta^*(I))$, we  obtain the minimal empirical reconstruction error $\FF(I,\hat{\Theta}(I))$. We formulate this statement below.

\begin{prop}\label{prop:miniemperror} 
When the reconstruction method is $\hat{\mathbf{x}}_{I,t} = \Theta\mathbf{x}_{I^c,t}$,
the selection result $\hat{I}$ is given by 
\begin{equation}\label{eq:bestsetlinH=0}
    \argmin_{I \subset \mathcal{N}  \; : \; |I| = p} tr(\hat{\bSigma}_{I} -  \hat{\bSigma}_{II^c}\hat{\bSigma}_{I^c}^{-1}\hat{\bSigma}_{I^cI}).
\end{equation}
\end{prop}
The quantity $tr(\hat{\bSigma}_{I} -  \hat{\bSigma}_{II^c}\hat{\bSigma}_{I^c}^{-1}\hat{\bSigma}_{I^cI})
$ is not only the minima of (\ref{eq:recpb}), but also a consistent estimate of $tr(\bSigma_{I} -  \bSigma_{II^c}\bSigma_{I^c}^{-1}\bSigma_{I^cI})$\footnote{However it is biased, because the inverse of sample covariance is not an unbiased estimator of precision matrix. Besides, when the dimension goes up, sample covariance can be ill-conditioned, in this case, its (possibly generalized) inverse is not a good estimate in any sense.}. Therefore in practice, we use the criterion (\ref{eq:bestsetlinH=0}) to derive the greedy Algorithm \ref{algo:H=0} described below to select an optimal subset $\hat{I}$. 
\begin{algorithm}[h]
\SetAlgoLined
\textbf{Input:} $\hat{\bSigma}, \; p$.

\textbf{Initialize:} $n = 0$, $I_{(n)} = \emptyset$, $I^c_{(n)} = \mathcal{N}$.

\For{$n < p$}{

$n \leftarrow n+1$

$i_{(n)} \leftarrow \arg\min_{i \in I^c_{(n-1)}} \hat{\sigma}^2_i - \hat{\bSigma}_{iS}\hat{\bSigma}_{S}^{-1}\hat{\bSigma}_{Si}, \quad \mbox{where} \quad S = I^c_{(n-1)} \backslash i$. 
    
$I_{(n)} \leftarrow I_{(n-1)} \cup i_{(n)}$.
    
$I^c_{(n)} \leftarrow I^c_{(n-1)} \backslash i_{(n)}$.
}
 
\textbf{Output:} $i_{(1)}, i_{(2)}, ..., i_{(p)}$.
 
\caption{Greedy algorithm of sensor selection with linear reconstruction, $H = 0$.}
 \label{algo:H=0}
\end{algorithm}

Starting with an empty turned-off set $I$, at each step, we solve problem (\ref{eq:bestsetlinH=0}) for $p = 1$ and only among the current $I^c$. Then, we add the minimizer $i^*$ to the current set $I$ while removing it from the current $I^c$. From Algorithm \ref{algo:H=0}, it follows that most of the calculation comes from the inversion of the matrix $\hat{\bSigma}_{S}$, and thus the computational complexity is $\mathcal{O}(pN^4)$.

This algorithm coincides with the one induced from another conventional sensor selection method, which is based on Gaussian process assumption and entropy measurement \cite{sakiyama2019eigendecomposition, Shewry87}. In these works, it is assumed that a random signal $\mathbf{x}$ follows a Gaussian joint zero-mean distribution $p(\mathbf{x})$, which reads as
$$
p(\mathbf{x}) = \frac{1}{(2\pi)^{\frac{N}{2}}\mbox{det}[\bSigma]}\exp \left(-\frac{1}{2}\mathbf{x}^t\bSigma^{-1}\mathbf{x}\right).
$$
In the Gaussian framework, the partial covariance and conditional covariance of a subset $A$ of variables given on another subset $B$ are identical. In addition, the linear estimator $\bSigma_{II^c}\bSigma_{I^c}^{-1}\mathbf{x}_{I^c}$ becomes the best estimator of $\mathbf{x}_I$, because $\E(\mathbf{x}_I | \mathbf{x}_{I^c})$ is a linear function of $ \mathbf{x}_{I^c}$. The objective function in \cite{sakiyama2019eigendecomposition, Shewry87} is  the entropy optimal design defined as
\begin{equation}\label{eq:entropyobj}
    \hat{I}^c = \argmax_{I^c \subset \mathcal{N}  \; : \; |I| = p} \mbox{log det}[\bSigma_{I^c}],
\end{equation}
which aims to find out the most informative subset $\hat{I}^c$. From Schur's determinant identity, we know that 
$\mbox{det}[\bSigma_{I^c}] = \mbox{det}[\bSigma]/ \mbox{det}[\bSigma_{I} -  \bSigma_{II^c}\bSigma_{I^c}^{-1}\bSigma_{I^cI}]$. Thus, the objective function (\ref{eq:entropyobj}) amounts to
$$
\hat{I} = \argmax_{I \subset \mathcal{N}  \; : \; |I| = p} \mbox{log det}[\bSigma_{I^c}] = \argmin_{I \subset \mathcal{N}  \; : \; |I| = p} \mbox{det}[\bSigma_{I} -  \bSigma_{II^c}\bSigma_{I^c}^{-1}\bSigma_{I^cI}].
$$
When $p = 1$, $\mbox{det}[\bSigma_{I} -  \bSigma_{II^c}\bSigma_{I^c}^{-1}\bSigma_{I^cI}]$ also equals $\sigma^2_i - \bSigma_{iI^c}\bSigma_{I^c}^{-1}\bSigma_{I^ci}$. Therefore, the resulting greedy algorithms is equal to Algorithm \ref{algo:H=0}. The only difference is that, in our setting, we did not add the Gaussian assumption, allowing the algorithm applied over various population distributions, only if their second moment exists. Note that for many distributions, the linear estimator is not necessarily the best reconstruction function.

\begin{figure}[h]
    \centering
    \includegraphics[width=0.38\textwidth]{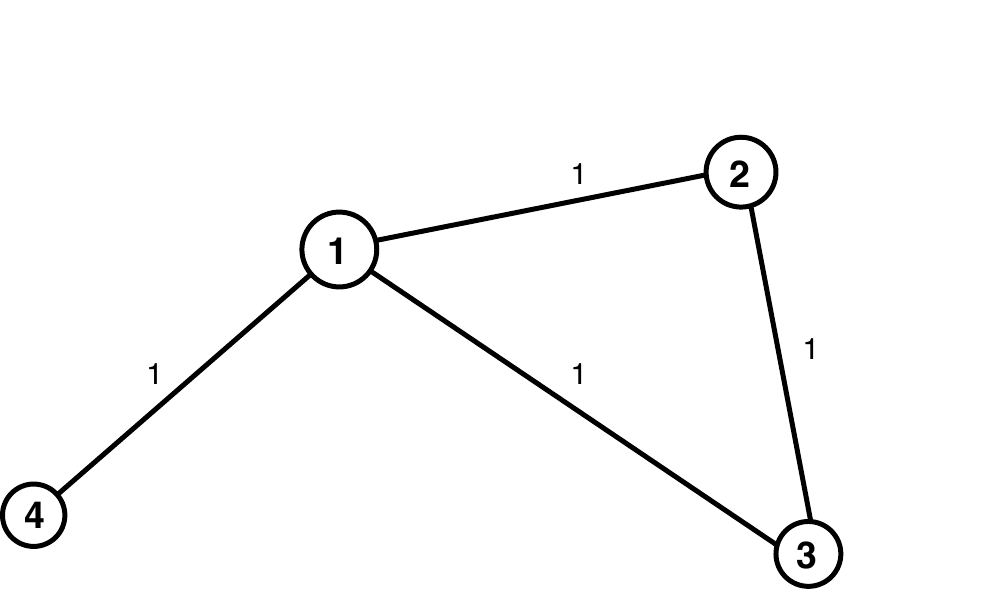}
    \caption{\small When the sensor covariance is $\Sigma = A + D$, the partial variances of each sensor given all the rest are: $1.33$, $1.33$, $1.33$,  $0.57$, from sensor $1$ to $4$ respectively. When the sensor covariance is $D^{-\frac{1}{2}}AD^{-\frac{1}{2}} + \Id$, the partial variances of each sensor given all the rest are: $0.44$, $0.67$, $0.67$, $0.57$, from sensor $1$ to $4$ respectively.}
    \label{fig:simple graph example}
\end{figure}

Criteria (\ref{eq:bestsetlinH=0_theo}) also implies that, the simplest way to select sensors is choosing those of small variance. Indeed, if we furthermore derive the criteria, we have that
$$
tr(\bSigma_{I} -  \bSigma_{II^c}\bSigma_{I^c}^{-1}\bSigma_{I^cI}) = \sum\limits_{i \in I} \sigma^2_i - tr(\bSigma_{II^c}\bSigma_{I^c}^{-1}\bSigma_{I^cI}) \in \left[0 \; , \;  \sum\limits_{i \in I} \sigma^2_i\right].
$$
Hence, the set $I$ which consists of sensors with small variance $\sigma^2_i$ is more likely to have a smaller criteria value, and thus to be selected. We illustrate this fact with the example in Figure \ref{fig:simple graph example}. We assume the observation of a signal on $N=4$ sensors that has a population covariance $\Sigma = A + D$, where $A$ is the adjacency matrix of the graph displayed in Figure \ref{fig:simple graph example}, and $D$ is its degree matrix. Hence, the edges equal to sensor covariance, and the node degree is sensor variance. The results of sensor selection are then easily obtained  through  a direct calculation of criteria (\ref{eq:bestsetlinH=0_theo}). When $p = 1$, the selected sensor is $I = \{4\}$, which has the smallest variance equal to $1$. Nevertheless, if we standardize the variance of the data which amounts to consider the correlation matrix $D^{-\frac{1}{2}}AD^{-\frac{1}{2}} + \Id$ instead of $\Sigma$, the selection process yields the choice of sensor $I = \{1\}$ as is the most predictable one.  This rough way of sensor selection is valid once the selection criteria is induced from mean squared error (MSE). However, whether to scale the data so that all sensors have unit variance $\sigma^2_i = 1$ depends on the application, and this will be discussed in numerical experiments.

\subsection{The case $H>0$}\label{sec:linH>0}
When recovering the missing part $\mathbf{x}_{I,t}$ of a weakly stationary process $(\mathbf{x}_t)_t$, we can use more past information, which brings to  the linear reconstruction method $\Theta \mathbf{x}_{I^c,t}^H$. For the simplicity, we still assume the mean of $(\mathbf{x}_t)_t$ is zero. Then, the theoretical reconstruction error becomes
$$
\FFtheo(I,\Theta) = \mathbb{E}\|\mathbf{x}_{I,t} - \Theta\mathbf{x}_{I,t}^H\|^2_{l_2}.
$$
Because of the stationarity of $(\mathbf{x}_t)_t$, we can use its autocovariance matrix $\mathbf{\Gamma}(l)$ to represent the covariance matrices $\Cov(\mathbf{x}_{I,t}, \,  \mathbf{x}_{I^c,t}^H)$ and $\Cov(\mathbf{x}_{I^c,t}^H)$, denoted by $\beta^H_{II^c}$ and $\alpha^H_{I^c}$ respectively, where $\mathbf{x}_{I^c,t}^H$ is the vector defined by \eqref{eq:defxH}. Using  the same calculations as in the preliminaries of this section, we have that the best turned-off set $I$ is given by 
$$
\argmin_{I \subset \mathcal{N}  \; : \; |I| = p}  \FFtheo(I,\Theta^*(I)) = tr(\bSigma_{I} - [\beta^H_{II^c}][\alpha^H_{I^c}]^{-1}[\beta^H_{II^c}]^t),
$$
where 
\begin{equation}\label{eq:alphaH}
    \alpha^H_{I^c} = 
    \begin{pmatrix}
    \mathbf{\Gamma}_{I^c}(0) & \mathbf{\Gamma}_{I^c}(1) & \cdots & \mathbf{\Gamma}_{I^c}(H)\\
    \mathbf{\Gamma}_{I^c}(-1) & \mathbf{\Gamma}_{I^c}(0) & \cdots & \mathbf{\Gamma}_{I^c}(H - 1)\\
    \vdots & \vdots & \ddots & \vdots\\
    \mathbf{\Gamma}_{I^c}(-H) & \mathbf{\Gamma}_{I^c}(-H + 1) & \cdots & \mathbf{\Gamma}_{I^c}(0)
    \end{pmatrix} \in \R^{q \times q},
\end{equation}

\begin{equation}\label{eq:betaH}
    \beta^H_{II^c} = 
    \begin{pmatrix}
    \mathbf{\Gamma}_{II^c}(0) &  \mathbf{\Gamma}_{II^c}(1) & \cdots & \mathbf{\Gamma}_{II^c}(H)
    \end{pmatrix} \in \R^{p \times q}.
\end{equation}

Similarly, we replace $\bSigma_{I}, \alpha^H_{I^c}$ and $\beta^H_{II^c}$ in $\FFtheo(I,\Theta^*(I))$ with their sample estimates $\hat{\bSigma}_{I}, \hat{\alpha}^H_{I^c}$ and $\hat{\beta}^H_{II^c}$ to get the minimal empirical reconstruction error $\FF(I,\hat{\Theta}(I))$ readily. Note that the sample estimates $\hat{\alpha}^H_{I^c}$ and $\hat{\beta}^H_{II^c}$ are obtained by substituting the sample autocovariance $\hat{\mathbf{\Gamma}}_{I^c}(l)$ and $\hat{\mathbf{\Gamma}}_{II^c}(l)$ block-wise, which are equal to $\frac{1}{T_0} \sum_{t = l + 1}^{T_0} \mathbf{x}_{I^c,t}\mathbf{x}_{I^c,t - l}^t$ and $\frac{1}{T_0} \sum_{t = l + 1}^{T_0} \mathbf{x}_{I, t}\mathbf{x}_{I^c, t - l}^t$,  respectively.
We state these facts into Proposition \ref{prop:miniemperrorH>0}.
\begin{prop}\label{prop:miniemperrorH>0} 
When the reconstruction method is $\hat{\mathbf{x}}_{I,t} = \Theta \mathbf{x}_{I^c,t}^H$, the selection result $\hat{I}$ is given by
\begin{equation}\label{eq:bestsetlinH>0}
    \argmin_{I \subset \mathcal{N}  \; : \; |I| = p} tr(\mathbf{\hat{\Sigma}}_{I} - [\hat{\beta}^H_{II^c}][\hat{\alpha}^H_{I^c}]^{-1}[\hat{\beta}^H_{II^c}]^t).
\end{equation}
\end{prop}
The resulting greedy algorithm is Algorithm \ref{algo:H>0}, which is of computational complexity $\mathcal{O}(pN^4H^3)$ which is acceptable for moderate values of $N \times H$. When the matrix $\hat{\alpha}^H_{I^c}$ is large, in addition to expensive matrix inversion, the ill-condition problem also occurs. In this case, we can build linear ridge regression instead, which is a particular case of the selection method discussed in the following section. In this setting, Algorithm \ref{algo:kernelH>0, CG}  (described later on) will be useful, as it speeds up matrix inversion using conjugate gradient method.
\begin{algorithm}[h]
\SetAlgoLined
\textbf{Input:} $(\hat{\mathbf{\Gamma}}(j))_{j = 0,1,...,H}, \, p$.

\textbf{Initialize:} $n = 0$, $I_{(n)} = \emptyset$, $I^c_{(n)} = \mathcal{N}$.

\For{$n < p$}{

$n \leftarrow n+1$

$i_{(n)} \leftarrow \arg\min_{i \in I^c_{(n-1)}} \hat{\sigma}^2_i - [\hat {\mathbf{\beta}}^H_{iS}][\hat{\alpha}^H_{S}]^{-1}[\hat{\mathbf{\beta}}^H_{iS}]^t, \quad \mbox{where} \quad S = I^c_{(n-1)} \backslash i$. 
    
$I_{(n)} \leftarrow I_{(n-1)} \cup i_{(n)}$.
    
$I^c_{(n)} \leftarrow I^c_{(n-1)} \backslash i_{(n)}$.

}
 
 \textbf{Output:} $i_{(1)}, i_{(2)}, ..., i_{(p)}$.
 
 \caption{Greedy algorithm of sensor selection with linear reconstruction, $H > 0$.}
 \label{algo:H>0}
\end{algorithm}

\section{Graph kernel approaches} \label{sec:kernel}

In this section, for a given $H \geq 0$, we employ the class of {\it reconstruction methods}  that incorporates  the underlying graph structure of the network of sensors as part of its construction. More specifically, we consider the class of functions built from a \textit{reproducing kernel Hilbert space (RKHS)}. In our setting, the graph structure will be the core of the design of the underlying kernel.
The layout of this section is then as follows. In Subsection \ref{sec:Bg}, we present the preliminaries of kernel regression required to build our models. In Subsections \ref{sec:kernelH=0} and \ref{sec:kernelH>0}, we derive the proposed sensor selection criteria based on kernel reconstruction approaches, yet without relying on the specific kernels. Lastly in Subsection \ref{sec:kerneldesigns}, we specify  the kernel design in our context.

\subsection{Backgrounds on reproducing kernel Hilbert space and kernel regression}\label{sec:Bg}
We start by defining the notion of a kernel, and we recall very well known facts on kernel regression.
\begin{defin}\label{def:kernel}

Let $\Y = \R$\footnote{In a general setting, $\Y$ is a real Hilbert space, a kernel is a mapping from $\X \times \X$ to $\mathcal{L(Y)}$, which is the set of all bounded linear operators from $\Y$ to itself. For example, the kernel value $k(x, y)$ can be a vector. Because throughout our models, a real scalar kernel value is sufficient, we only present the definition for $\Y = \R$. For more details see \cite{micchelli2005learning}.}, $\X$ be a set. Then 
$k: \X \times \X \rightarrow \Y$ is a kernel on $\X$ if  $k$ is symmetric: $k(x, y) = k(y, x)$, and  $k$ is positive definite, that is $\forall \, x_1, x_2, ..., x_n \in \X$, the \textit{Gram matrix} $K$ defined by $K_{ij} = k(x_i, x_j)$ is positive semi-definite (PSD).
\end{defin}

We recall two classical examples of valid kernels, which will be referred to later on.

\begin{exe}(Linear kernel)
For $\X  =   \R^m$, the mapping $k_{lin}$ defined by
\begin{equation}\label{eq:linearkernel}
\left\{
\begin{array}{ccl}
k_{lin} : \R^m \times \R^m & \to & \R \\
(\mathbf{x},\textbf{y}) & \mapsto & k_{lin}(\mathbf{x},\textbf{y}) = \frac{1}{m} \mathbf{x}^t\textbf{y}
\end{array}
\right.
\end{equation}
 is a kernel on $\R^m$.
\end{exe}
Linear kernel is commonly used to induce the space of linear functions. Note that if inputs $x_1, x_2, ..., x_p \in \R^m$ represent the random variables with mean zero, then the linear kernel value is actually the sample covariance, together with the Gram matrix to be the corresponding sample covariance matrix.

\begin{exe}(Gaussian kernel)
For $\X  =   \R^m$, the mapping $k_{rbf}$ defined by
\begin{equation}\label{eq:rbfkernel}
\left\{
\begin{array}{ccl}
k_{rbf} : \R^m \times \R^m & \to & \R \\
(\mathbf{x},\textbf{y}) & \mapsto & k_{rbf}(\mathbf{x},\textbf{y}) = \exp\left[-\gamma\|\mathbf{x}-\textbf{y}\|^2_{\ell^2}\right]
\end{array}
\right.
\end{equation}
is a kernel on $\R^m$, where $\gamma > 0$ is a scaling parameter. 
\end{exe}

$k_{rbf}$ is also called a \textit{radial basis function (RBF)} kernel, which is a popular kernel function used in various kernelized learning algorithms. 
A kernel  uniquely defines a \textit{reproducing kernel Hilbert space (RKHS)}, $\mathcal{H}_k = \{f(\cdot) = \sum_{i = 1}^n \alpha_ik(x_i, \cdot) | \alpha_i \in \R\}$, where to search a predictor from the observations $(x_1, y_1)$, $(x_2,y_2)$, $..., (x_n,y_n) \in \X \times \R$.
To accomplish this, we recall that \textit{kernel ridge regression} is the following optimization problem:
\begin{equation}\label{eq:kernelridgeregression}
    f^* = \argmin\limits_{f \in \mathcal{H}_k} \sum\limits_{i = 1}^n \|y_i - f(x_i) \|^2_{\ell^2} + \lambda \|f\|_{\mathcal{H}_k}^2, \mbox{ for some } \lambda \geq 0,    
\end{equation}
and that by the representer theorem \cite{scholkopf2001generalized} the above minimizer has the form $f^* = \sum_{i = 1}^n \alpha_{i}^*k(x_i, \cdot),$
with the vector of optimal coefficients given by
\begin{equation}
\begin{aligned}
    \alpha^* &= \argmin\limits_{\mathbf{\alpha \in {\rm I\!R^{n}}}} \|\mathbf{y} - K_n\alpha\|_{\ell^2}^2 + \lambda \mathbf{\alpha}^tK_n\mathbf{\alpha} \\
    &= (K_n + \lambda \Id)^{-1}\mathbf{y}
\end{aligned}
\end{equation}
where $\mathbf{\alpha} = (\alpha_1,...,\alpha_n)^t$, $\mathbf{y} = (y_1,...,y_n)^t$ and $K_n$ is the Gram matrix of inputs $x_1, x_2, ..., x_n$. Therefore the resulting predictor is $f^*(z) = \sum_{i = 1}^n \alpha_{i}^*k(x_i, z)$, with $\|\hat{f}\|_{\mathcal{H}_k}^2 = \mathbf{y}^t(K_n + \lambda \Id)^{-1}K_n(K_n + \lambda \Id)^{-1}\mathbf{y}$.
In Subsections \ref{sec:kernelH=0} and \ref{sec:kernelH>0}, we present the proposed sensor selection strategies without using an explicit kernel form. These strategies are applicable to all the kernel designs which fit the general framework. In addition, as in the previous section, we assume  that $(\mathbf{x}_{t})_t$ is a weakly stationary multivariate process with zero mean.

\subsection{The case $H = 0$}\label{sec:kernelH=0}

Since we do not consider the temporal factor, we set the input set $\X = \mathcal{N}$. Given the values $\mathbf{x}_{I^c,t}$, we build upon kernel ridge regression  to define the reconstruction of the missing values $\mathbf{x}_{I,t}$ as follows:
\begin{equation}\label{eq:kernelregressionH=0}
    \alpha^*_t = \argmin\limits_{\mathbf{\alpha \in {\rm I\!R^{N-p}}}} \|\mathbf{x}_{I^c,t} - K_{I^c}\alpha\|_{\ell^2}^2 + \lambda \mathbf{\alpha}^tK_{I^c}\mathbf{\alpha},
\end{equation}
where $K_{I^c}$ is the Gram matrix of input $I^c$.
From the preliminaries, we know that $\alpha^*_t = (K_{I^c} + \lambda \Id)^{-1}\mathbf{x}_{I^c,t}$, and thus the reconstruction 
has the form
$$
\hat{\mathbf{x}}_{I,t} = K_{II^c}\alpha^*_t = K_{II^c}(K_{I^c} + \lambda \Id)^{-1}\mathbf{x}_{I^c,t},
$$
where $K_{II^c} \in \R^{p \times (N-p)}$ is the matrix with entries equal to $k(i,j)$ for $i \in I, j \in I^c$.
Therefore the reconstruction error (\ref{eq:recpb}) in this case writes as 
\begin{equation}\label{eq:kernelslovelambdaH=0}
    \frac{1}{T_0} \sum\limits_{t = 1}^{T_0} \|\mathbf{x}_{I,t} - \hat{\mathbf{x}}_{I,t} \|_{\ell_2}^2 = \frac{1}{T_0} \sum\limits_{t = 1}^{T_0} \|\mathbf{x}_{I,t} -  K_{II^c}(K_{I^c} + \lambda \Id)^{-1}\mathbf{x}_{I^c,t} \|_{\ell_2}^2.
\end{equation}
By a slight abuse of  notation we denote the quantity  $K_{II^c}(K_{I^c} + \lambda \Id)^{-1}$ by $\hat{\Theta}_{\lambda}(I)$. Then, the best turned-off set $I$ is given by
\begin{equation}\label{eq:bestsetkernelH=0}
\begin{aligned}
\hat{I} = \argmin_{I \subset \mathcal{N}  \; : \; |I| = p} \FF(I,\hat{\Theta}_{\lambda}(I)) &= \argmin_{I \subset \mathcal{N}  \; : \; |I| = p} \frac{1}{T_0} \sum\limits_{t = 1}^{T_0} \|\mathbf{x}_{I,t} - \hat{\Theta}_{\lambda}(I)\mathbf{x}_{I^c,t} \|_{\ell_2}^2 \\
&= \argmin_{I \subset \mathcal{N}  \; : \; |I| = p} tr(\hat{\bSigma}_{I} - 2\hat{\bSigma}_{II^c}\hat{\Theta}_{\lambda}(I)^t + \hat{\Theta}_{\lambda}(I)\hat{\bSigma}_{I^c}\hat{\Theta}_{\lambda}(I)^t).
\end{aligned}
\end{equation}
Now, recall that the sensor selection criterion in Subsection \ref{sec:linH=0} writes as 
$$\min_{I \subset \mathcal{N}  \; : \; |I| = p} tr(\hat{\bSigma}_{I} -  \hat{\bSigma}_{II^c}\hat{\bSigma}_{I^c}^{-1}\hat{\bSigma}_{I^cI}),$$
and it thus aligns with criterion (\ref{eq:bestsetkernelH=0}). If we compare the quantity $\hat{\Theta}_{\lambda}(I)$ with the corresponding one from the linear case when $H=0$ (which is equal to  $\hat{\bSigma}_{II^c}\hat{\bSigma}_{I^c}^{-1}$), we  find that the Gram matrix now plays the role of the sample covariance matrix\footnote{The term $\lambda \Id$ in the inversion can be thought of as to condition the square matrix $K_{I^c}$.}.

Moreover, if one chooses the linear kernel defined by \eqref{eq:linearkernel}, then we are essentially using linear ridge regression as a reconstruction method. In this case, all Gram matrices in formula (\ref{eq:bestsetkernelH=0}) will become the corresponding sample covariances. Furthermore, when $\lambda = 0$, then the two criteria will be exactly identical, which indicates that the sensor selection criterion (\ref{eq:bestsetkernelH=0}) from the kernel approach can correspond the one in the linear case (\ref{eq:bestsetlinH=0}) when $H = 0$ and $K$ built from  the linear kernel \eqref{eq:linearkernel}. The same correspondence holds in the case $H > 0$ as detailed in the next subsection. 

We finally derive the following greedy Algorithm \ref{algo:kernelH=0} from criterion (\ref{eq:bestsetkernelH=0}). Its computational complexity is $\mathcal{O}(pN^4)$.
\begin{algorithm}[h]
\SetAlgoLined
\textbf{Input:} $\hat{\bSigma}, \, K_{\mathcal{N}}, \, p, \, \lambda$.

\textbf{Initialize:} $n = 0$, $I_{(n)} = \emptyset$, $I^c_{(n)} = \mathcal{N}$.

\For{$n < p$}{

$n \leftarrow n+1$

$i_{(n)} \leftarrow \arg\min_{i \in I^c_{(n-1)}} \hat{\sigma}^2_i - 2\hat{\bSigma}_{iS}\hat{\Theta}_{\lambda}(i)^t + \hat{\Theta}_{\lambda}(i)\hat{\bSigma}_{S}\hat{\Theta}_{\lambda}(i)^t, \quad \mbox{where} \quad \hat{\Theta}_{\lambda}(i) = K_{iS}(K_{S} + \lambda \Id)^{-1}, \quad S = I^c_{(n-1)} \backslash i$. 

\vspace{0.02in}
    
$I_{(n)} \leftarrow I_{(n-1)} \cup i_{(n)}$.
    
$I^c_{(n)} \leftarrow I^c_{(n-1)} \backslash i_{(n)}$.
}
 
\textbf{Output:} $i_{(1)}, i_{(2)}, ..., i_{(p)}$.
 
\caption{Greedy algorithm of sensor selection with kernel ridge regression reconstruction, $H=0$.}
\label{algo:kernelH=0}
\end{algorithm}

\subsection{The case $H > 0$}\label{sec:kernelH>0}

We now leverage $\mathbf{x}_{I^c,t}^H$ to recover $\mathbf{x}_{I,t}$. Thus, the input of kernel should take input values in time as well. Therefore, we expand the input set $\X$ from $\mathcal{N}$ to $\mathcal{N} \times \Z$. Because the process $(\mathbf{x}_{t})_t$ is stationary, we require the kernel value to only depend on the time lag $l$ rather than specific time stamps, meaning that $k\left[(i,t),(j,t-l) \right] = k\left[(i,t'),(j,t'-l) \right]$. Hence, we can denote the kernel value $k\left[(i,t),(j,t-l) \right]$ by $k(i,j,l)$. With this framework, we set up kernel ridge regression (\ref{eq:kernelregressionH>0}) to define a reconstruction function by solving
\begin{equation}\label{eq:kernelregressionH>0}
    \alpha^*_t = \argmin\limits_{\mathbf{\alpha \in {\rm I\!R^q}}} \|\mathbf{x}_{I^c,t}^H - K_{I^c}^H\alpha\|_{\ell^2}^2 + \lambda \mathbf{\alpha}^tK_{I^c}^H\mathbf{\alpha},
\end{equation}
where
$$
K_{I^c}^H = 
\begin{pmatrix}
    K_{I^c}(0) & K_{I^c}(1) & \cdots & K_{I^c}(H)\\
    K_{I^c}(-1) & K_{I^c}(0) & \cdots & K_{I^c}(H-1)\\
    \vdots & \vdots & \ddots & \vdots\\
    K_{I^c}(-H) & K_{I^c}(-H+1) & \cdots & K_{I^c}(0)
\end{pmatrix} \in \R^{q \times q},
$$
which is the Gram matrix of input $I^c \times (t,t-1,...,t-H)$.
Matrix block $K_{I^c}(l) \in \R^{(N-p) \times (N-p)}, \, l \in \{0,\pm 1,...,\pm H\}$, whose  entries equal $k(i,j,l)$ for  $i,j \in I^c$. From the preliminaries, we know that $\alpha^*_t = (K_{I^c}^H + \lambda \Id)^{-1}\mathbf{x}^H_{I^c,t}$,  and thus the reconstruction  has the form 
$$
\hat{\mathbf{x}}_{I,t} = K_{II^c}^H\alpha^*_t = K^H_{II^c}(K_{I^c}^H + \lambda \Id)^{-1}\mathbf{x}_{I^c,t}^H \, ,
$$
where
\begin{equation*}
    K_{II^c}^H = 
    \begin{pmatrix}
    K_{II^c}(0) &  K_{II^c}(1) & \cdots & K_{II^c}(H)
    \end{pmatrix} \in \R^{p\times q},
\end{equation*}
with entries equal to $k(i,j,l)$ for $i \in I, j \in I^c$. Thus, the best turned-off set $I$ is given by
\begin{equation}\label{eq:kernelslovelambdaH>0}
\begin{aligned}
\hat{I} = \argmin_{I \subset \mathcal{N}  \; : \; |I| = p}  \FF(I,\hat{\Theta}_{\lambda}(I)) & = \argmin_{I \subset \mathcal{N}  \; : \; |I| = p} \frac{1}{T_0} \sum\limits_{t = H+1}^{T_0} \|\mathbf{x}_{I,t} -  K^H_{II^c}(K_{I^c}^H + \lambda \Id)^{-1}\mathbf{x}_{I^c,t}^H \|_{\ell_2}^2 \\
& = \argmin_{I \subset \mathcal{N}  \; : \; |I| = p}  tr(\hat{\bSigma}_{I} - 2\hat{\mathbf{\beta}}^H_{II^c}\hat{\Theta}_{\lambda}(I)^t + \hat{\Theta}_{\lambda}(I)\hat{\mathbf{\alpha}}_{I^c}^H\hat{\Theta}_{\lambda}(I)^t),
\end{aligned}
\end{equation}
where $\hat{\mathbf{\alpha}}_{I^c}^H$,  $\hat{\mathbf{\beta}}_{II^c}^H$ are the sample estimates of (\ref{eq:alphaH}) (\ref{eq:betaH})\footnote{Recall that $H \ll T_0$.}, and $\hat{\Theta}_{\lambda}(I)$ denotes $K^H_{II^c}(K_{I^c}^H + \lambda \Id)^{-1}$. 
Similarly, if we define the kernel value as
\begin{equation}
k(i,j,l) = \frac{1}{T_0} \sum_{t = l + 1}^{T_0} x_{it}x_{j,t - l}, \label{eq:kernelautocov}
\end{equation}
which is the sample autocovariance between $(x_{it})_t$ and $(x_{jt})_t$, then $K^H_{I^c} = \hat{\mathbf{\alpha}}_{I^c}^H$ and $K^H_{II^c} = \hat{\mathbf{\beta}}_{II^c}^H$. The kernel ridge regression becomes linear ridge regression again with $\lambda$ set to be $0$. Hence, for this choice of kernel, the criterion (\ref{eq:kernelslovelambdaH>0}) is equivalent to the one (\ref{eq:bestsetlinH>0}) in  the linear case for $H > 0$.

Interestingly, in the linear case, $\lambda = 0$ can be interpreted as an optimal choice of the regularization parameter (over the training set) as shown by the following result.

\begin{prop}\label{prop:bastlambda} 
Assume that the kernel is taken as the sample autocovariance, namely its entries are given by \eqref{eq:kernelautocov}. Then, the function $\lambda \mapsto  \FF(I,\hat{\Theta}_{\lambda}(I))$  is monotone increasing on $[0, + \infty)$, and its unique minimizer is $\lambda = 0$.
\end{prop}

We can induce the resulting greedy algorithm in the same way as previously, which will lead to $\mathcal{O}(pN^4H^3)$ complexity. However when $N \times H$ is large, the algorithm is too costly from a computational point of view.  Nevertheless,  $\lambda \Id$ is a natural matrix conditioner, and we can thus calculate $\hat{\Theta}_{\lambda}(I)$ using a \textit{conjugate gradient method} \cite{shewchuk1994introduction}, instead of a direct inversion. Then we can reach the computational complexity $\mathcal{O}(pN^3H^2\sqrt{\kappa})$, where $\kappa$ is the condition number of $K_{I^c}^H + \lambda \Id$. When $\sqrt{\kappa} \ll NH$, conjugate gradient method helps to reduce the time complexity considerably. We summarize the selection procedure in Algorithm \ref{algo:kernelH>0, CG}.
Besides, when dealing with a large network, namely $N$ itself is great, Algorithm \ref{algo:kernelH=0} can be modified using a conjugate gradient method in the same way.

\begin{algorithm}[h]
\SetAlgoLined
\textbf{Input:} $(\hat{\mathbf{\Gamma}}(j))_{j = 0,1,...,H}, \, K_{\mathcal{N}}^H, \, p, \, \lambda, \, \epsilon$.

\textbf{Initialize:} $n = 0$, $I_{(n)} = \emptyset$, $I^c_{(n)} = \mathcal{N}$.

\For{$n < p$}{

$n \leftarrow n+1$

$i_{(n)} \leftarrow \arg\min_{i \in I^c_{(n-1)}} \hat{\sigma}^2_i - 2\hat{\mathbf{\beta}}^H_{iS}\hat{\Theta}_{\lambda}(i)^t + \hat{\Theta}_{\lambda}(i)\hat{\mathbf{\alpha}}_{S}^H\hat{\Theta}_{\lambda}(i)^t$, where $\hat{\Theta}_{\lambda}(i)^t = cg^*(\, K_{S}^H + \lambda \Id, \, [K^H_{iS}]^t, \, \epsilon)$, $S = I^c_{(n-1)} \backslash i$. 

\vspace{0.02in}
    
$I_{(n)} \leftarrow I_{(n-1)} \cup i_{(n)}$.
    
$I^c_{(n)} \leftarrow I^c_{(n-1)} \backslash i_{(n)}$.
}
 
\textbf{Output:} $i_{(1)}, i_{(2)}, ..., i_{(p)}$.

\vspace{0.06in}

\text{\small $cg(A, b, \epsilon)$ denotes the conjugate gradient approximation of $A^{-1}b$ with tolerance $\epsilon$.}
 
\caption{Greedy algorithm of sensor selection with kernel ridge regression reconstruction, $H>0$.}
\label{algo:kernelH>0, CG}
\end{algorithm}

Lastly, we discuss about how to tune the hyperparameters in the reconstruction method $f_{\Theta}^\mathcal{G}$. The analysis so far is based on the fact that $f_{\Theta}^\mathcal{G}$ is given, which means all of its hyperparameters (denoted by $c$) are known ahead of the sensor selection procedure. The set of hyperparameters   is $c = \{H\}$ in linear case, while it is $c = \{H, \lambda \}$, possibly together with the hyperparameters introduced in the definition of $k$ in kernel methods. When  computational time permits, we can select the best turned-off set $\hat{I}_{c}$ for each hyperparameter setting and calculate the corresponding reconstruction error $\FF(\hat{I}_{c},\hat{\Theta}(\hat{I}_{c}))$. Given a searching grid $C$, the best hyperparameters are given by 
\begin{equation}\label{eq: hyperparameter selection}
    c^* = \argmin_{c \in C}  \FF(\hat{I}_{c},\hat{\Theta}(\hat{I}_{c})),
\end{equation}
with the sensor selection result $\hat{I}_{c^*}$. In practice, we split our data set into training set and test set to simulate the historical and current data respectively. To reduce overfitting, we furthermore take a  proportion of training set as validation set, so that for a given $c$ we can first select sensor set $\hat{I}_{c}$ and fit the reconstruction function $f^{\GG}_{\hat{\Theta}(\hat{I}_{c})}$ on the training set, and then compute the reconstruction error  $\FF(\hat{I}_{c},\hat{\Theta}(\hat{I}_{c}))$ using validation data. Finally, we obtain $c^*$ and $\hat{I}_{c^*}$ through (\ref{eq: hyperparameter selection}). The test set is only used when evaluating the performance of methods. 

\subsection{Kernel designs}\label{sec:kerneldesigns}
We firstly recall the definitions of important quantities in graph signal processing.

\begin{defin}(Graph Laplacians)\label{def:graph laplacians}
In undirected graphs, the most commonly used Laplacian is \textit{combinatorial Laplacian}, which is defined as
$\bL = D - A,$
where $A$ is the adjacency matrix and $D \in \R^{N \times N}$ is the \textit{degree matrix}. 
One can normalize the combinatorial Laplacian to obtain the \textit{symmetric normalized Laplacian}, which is defined as
$
\bL_{sym} = D^{\frac{1}{2}}\bL D^{\frac{1}{2}}.
$
\end{defin}

\begin{defin}(Graph Fourier basis)
The eigen-decomposition of the Laplacian matrix $\bL$ yields the construction of the Fourier basis in vertex domain as follows:
$$
\mathbf{L} (or \, \, \mathbf{L}_{sym}) = \bPhi
    \begin{bmatrix}
    \lambda_0 & & & & \\
    & \lambda_1 & & & \\
    & & \ddots & & \\
    & & & & \lambda_{N-1}
    \end{bmatrix}
    \bPhi^t := \mathbf{\Phi\Lambda\Phi}^t
$$
where $0 = \lambda_0 < \lambda_1 \leq ... \leq \lambda_{N-1}$ are the eigenvalues of $\bL$, and 
$\bPhi = (\phi_0,\phi_1,...,\phi_{N-1})$ are called the \textit{Fourier basis} of graph $\GG$.
\end{defin}

 Given that the multiplicity of eigenvalue $\lambda_0 = 0$ equals the number of independent components of graph $\GG$ (see e.g.\ \cite{brouwer2012distance}), and since we consider a connected graph,  there is only $\lambda_0 = 0$ as the smallest eigenvalue with multiplicity one. 


\begin{defin}(Graph convolution) The graph convolution of the signal $\mathbf{h}_{in} \in \R^N$ by the filter $\mathbf{g} \in \R^N$ is
\begin{equation}\label{eq:GConv}
    \mathbf{h}_{out} = \bPhi
    \begin{bmatrix}
    \hat{g}(0)& & & & \\
    & \hat{g}(\lambda_1) & & & \\
    & & \ddots & & \\
    & & & & \hat{g}(\lambda_{N-1})
    \end{bmatrix}
    \bPhi^t \mathbf{h}_{in},
\end{equation}
where $\hat{g}(\bLambda) := [\hat{g}(\lambda_k) ]_{0 \leq k \leq N-1}= \bPhi^t\mathbf{g}$. 
We denote Equation (\ref{eq:GConv}) as $\mathbf{h}_{out} = \bPhi\hat{g}(\bLambda)\bPhi^t \mathbf{h}_{in}$.
\end{defin}
The two kernels used in our application are the following ones. 

\begin{defin}(Laplacian kernel)
The graph Laplacian kernel $k_{\GG}: \mathcal{N} \times \mathcal{N} \to \R$ is defined as $k_{\GG}(i,j) = \bK_{ij}$, where $\bK$ is the following matrix.
\begin{equation}
\bK =  \bPhi r(\bLambda) \bPhi^t, \label{eq:LapKern}
\end{equation}
and $r :  \R_{+} \to \R_{+}$ is a given mapping (chosen by modelers).
A typical example of $r$ is 
$$
r(\lambda) = 
\bigg\{
\begin{array}{ccl}
\lambda^{-1}, \quad \lambda \neq 0\\
0, \quad \mbox{otherwise}
\end{array},
$$
for which the  matrix $\bK = \bL^{-} $ is the Moore-Penrose inverse of $\bL$.
\end{defin}

\begin{defin}(Spatial-temporal kernel) The kernel $k_{\GG T}: \X \times \X \to \R$ is defined as
\begin{equation}\label{eq:STKern}
    k_{\GG T}\left[\, (i,t),(j,t') \, \right] =  k_{\GG}(i,j)\cdot k_{rbf}(t,t').
\end{equation}
\end{defin}
In Equation \eqref{eq:STKern}, we consider the restriction of the Gaussian kernel $k_{rbf}$ defined on $\R$ by (\ref{eq:rbfkernel})  to the subset $\Z$. Because $k_{\GG}$ and $k_{rbf}$ are both valid kernels on $\mathcal{N}$ and $\Z$ respectively, from the product rule, it follows that $k_{\GG T}$ is a valid kernel on $\mathcal{N} \times \Z$ (see Fact \ref{fact:prodkernel} in the Appendix). 

\section{Graph convolutional neural network and its adaption to sensor selection problem} \label{sec:GCN}

In this section, we consider graph convolutional neural networks (GCN) as the reconstruction method. GCN as one subclass of the graph neural network family, is mainly used in supervised and semi-supervised learning, such as graph-level classification \cite{zhang2018end, henaff2015deep} and node-level classification \cite{kipf2016semi, velivckovic2017graph}. Its key network architecture are graph convolution layers, which aims at exacting high-level node representations. On each layer, the graph convolution can be defined spectrally as Equation (\ref{eq:GConv}). This leads to filtering the output from previous layer, where $\hat{g}(\lambda)$ becomes a set of learnable network parameters $\mathbf{\Theta}$, which yields to spectral CNN \cite{bruna2013spectral}. Graph convolution can also be performed on the vertex domain directly by summing up the signals of neighbouring nodes with weights, as in \cite{micheli2009neural}. The GCN's variants essentially arise from different ways to approximate Equation (\ref{eq:GConv}), and to define the rules of weight-sharing among neighbours.

Apart from GCN, \cite{wu2019comprehensive} provides a comprehensive overview on all types of graph neural networks, among which the subclass \textit{spatial-temporal graph neural networks (STGNN)} can also be of interest as the reconstruction method. STGNNs sew graph convolution layers into network as spatial analyzing module, and interleave them with temporal analyzing modules, such as LSTM unit \cite{seo2018structured}, or CNN layer in temporal dimention \cite{yan2018spatial}. They can be useful when modelling dynamic node inputs at multiple time stamps. 

In the sensor selection context of this paper, our goal is not to propose novel network architectures to improve baseline performance on benchmark datasets. We rather study the way of transforming a prediction network into a selection network which adapts to its own prediction capability. To our knowledge, we have not found related work to this aim. We take the example of ChebNet \cite{defferrard2016convolutional} to illustrate the transfomation, which is used in our application. We first review the architecture of ChebNet. Then, we present its  adaptation to the sensor selection problem.

\subsection{ChebNet in missing graph signal reconstruction}

\begin{figure}[h]
    \centering
   \includegraphics[width=0.9\textwidth]{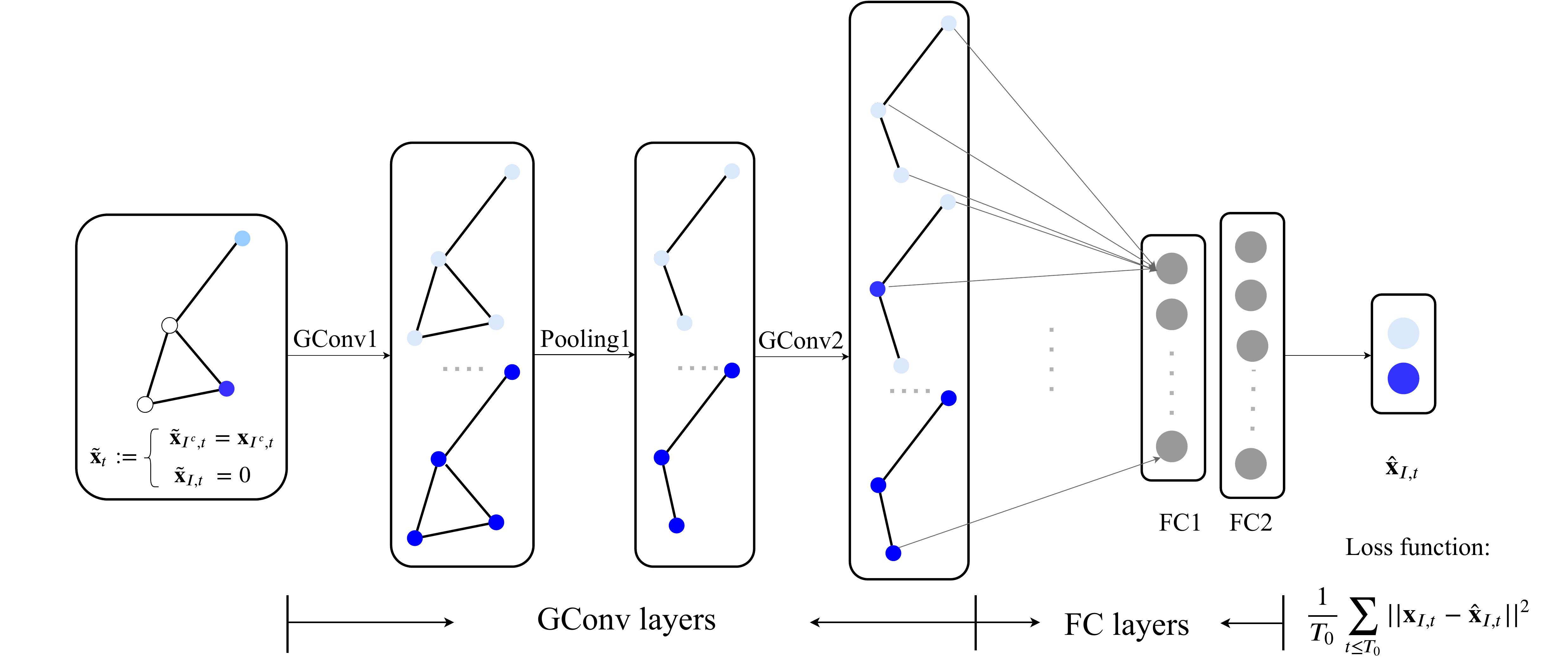}
    \caption{\small \textit{A spectral-based GCN in missing graph signal reconstruction.} In this illustrative architecture, $\hat{\mathbf{x}}_{I,t} = \mathbf{W}_3\sigma_2(\mathbf{W}_2\sigma_1(\mathbf{W}_1\mathbf{h}_{out} + \mathbf{b}_1) + \mathbf{b}_2) + \mathbf{b}_3$, where $\mathbf{h}_{out} = vec(\mathbf{H}_{out})$ is the flatten output from last GConv layer. $\mathbf{H}_{out} = GConv(Pool(\,GConv(\Tilde{\mathbf{x}}_{t}, \mathbf{L}; \mathbf{\Theta}_1)\,), \mathbf{L}_{pool}; \mathbf{\Theta}_2) \in \R^{N_{pool} \times F_{out}}$. $\sigma_1(\cdot), \sigma_2(\cdot)$ are layer activation functions, applied element-wise. $\mathbf{W}_1, \mathbf{b}_1, \mathbf{W}_2, \mathbf{b}_2, \mathbf{\Theta}_1, \mathbf{\Theta}_2$ are network trainable parameters, which will be learned by optimizing the loss function. $\mathbf{L}, \mathbf{L}_{pool}$ are graph Laplacians, which are also network input for spectral-based GCNs.}
    \label{fig:A spectral-based GCN in missing graph signal reconstruction}
\end{figure}

ChebNet falls into the category of spectral-based GCN, whose typical architecture is shown in Figure \ref{fig:A spectral-based GCN in missing graph signal reconstruction}. The graph convolution (GConv) layers filter the input graph signals and output multiple meaningful graph signals, whose components of a same node composes the learned node representation/embedding. These node embeddings serve as predictors in node/graph-level prediction/classification, their functional dependency with target quantities is approximated by fully-connected (FC) layers.

In the first work of spectral-based GCN \cite{bruna2013spectral}, $GConv(\mathbf{H}^{{m-1}}, \mathbf{L}; \mathbf{\Theta}^{(m)})$ is defined as 
$$
\mathbf{H}_{:,j}^{(m)} = \sigma_m(\sum\limits_{i = 1}^{F_{m-1}}\bPhi\mathbf{\Theta}_{i,j}^{(m)}\bPhi^t\mathbf{H}_{:,i}^{(m-1)}),
$$
where $m$ is the layer index, $\mathbf{H}^{(m-1)} \in \R^{N \times F_{m-1}}$ is the layer input,  $\mathbf{H}^{(m)} \in \R^{N \times F_{m}}$ is the layer output, $\mathbf{H}^{(0)}$ is the network input, $F_{m-1}$, $F_{m}$ are the number of input, output graph signals respectively, $N$ is the number of nodes, $\mathbf{L} \in \R^{N \times N}$ is the graph Laplacian, $\bPhi$ is its eigenvector matrix, $\mathbf{\Theta}_{i,j}^{(m)}$ is a diagonal matrix filled with $N$ learnable parameters whose diagonal elements represent the graph filter $\hat{g}(\mathbf{\Lambda})$, $\sigma_m(\cdot)$ is the layer activation function.

Due to the expensive computation of eigen-decomposition, in the follow-up works, ChebNet approximates $\hat{g}(\mathbf{\Lambda})$ by $\sum_{k = 0}^{K_m} \theta_kT_k(\Tilde{\mathbf{\Lambda}})$, where $\Tilde{\mathbf{\Lambda}} = 2\mathbf{\Lambda}/\lambda_{N-1} - \Id$,  $\lambda_{N-1}$ is the largest eigenvalue of the  Laplacian $\mathbf{L}$, and $T_k(x)$ is the $k$-th Chebyshev polynomial defined recursively as $T_k(x) = 2x T_{k-1}(x) - T_{k-2}(x), ~ T_0 = 1, ~ T_1 = x$. Note that $T_k(\Tilde{\mathbf{\Lambda}})$ applies the function element-wise on the entries of $\Tilde{\mathbf{\Lambda}}$. Then, the eigen-decomposition step is eliminated thanks to the following relations
$$
\bPhi\hat{g}(\mathbf{\Lambda})\bPhi^t\mathbf{H}_{:,i}^{(m-1)} = \bPhi\sum_{k = 0}^{K_m} \theta_kT_k(\Tilde{\mathbf{\Lambda}})\bPhi^t\mathbf{H}_{:,i}^{(m-1)} = \sum_{k = 0}^{K_m} \theta_k \bPhi T_k(\Tilde{\mathbf{\Lambda}})\bPhi^t\mathbf{H}_{:,i}^{(m-1)} = \sum_{k = 0}^{K_m} \theta_k T_k(\Tilde{\mathbf{L}})\mathbf{H}_{:,i}^{(m-1)}, 
$$
where $\Tilde{\mathbf{L}} = 2\mathbf{L}/\lambda_{N-1} - \Id$.
Because $T_k(\Tilde{\mathbf{L}})$ is the polynomial up to order $k$ of the normalized graph Laplacian, each term $\theta_k T_k(\Tilde{\mathbf{L}})\mathbf{H}_{:,i}^{(m-1)}$ computes a graph signal, whose node component is the weighted sum of its neighbours within $n$-hop neighbourhood, with the weight associated with $\theta_k$ and shared across all nodes. Thus, the convolution operation $\sum_{k = 0}^K \theta_k T_k(\Tilde{\mathbf{L}})\mathbf{H}_{:,i}^{(m-1)}$ can be regarded as a direct feature aggregation on vertex domain, and it aims to extract the features which are local and stationary over the graph.
The graph convolution $GConv(\mathbf{H}^{{m-1}}, \mathbf{L}; \mathbf{\Theta}^{(m)})$ in ChebNet is defined as
\begin{equation}\label{eq:GConv ChebNet}
\mathbf{H}_{:,j}^{(m)} = \sigma_m(\sum\limits_{i = 1}^{F_{m-1}}\sum_{k = 0}^{K_m} \theta_{k,i,j}^{(m)} T_k(\Tilde{\mathbf{L}})\mathbf{H}_{:,i}^{(m-1)} + \mathbf{b}^{(m)}_j),
\end{equation}
where $\mathbf{b}^{(m)}_j \in \R^N$ is the bias at layer $m$ for output signal $\mathbf{H}_{:,j}^{(m)}$, which can be a vector of $N$  learnable parameters, or chosen as $b^{(m)}_j\mathbf{1}$ with only one learnable parameter. The use of ChebNet also allows to perform graph signal pooling. To this aim, the initial graph is coarsed into multiple levels by the \textit{Graclus algorithm}, and the nodes are rearranged so that the ones left at each level form a binary tree. Graph signals are aggregated along this tree from the bottom to the top, which allows to build an optional pooling layer denoted by {\it Pool} in what follows. A set of forward propagation in the GConv block is thus defined as
\begin{equation}\label{eq:GConv pool}
\left\{
\begin{array}{ccl}
\mathbf{H}^{(m)} & = & GConv(\Tilde{\mathbf{H}}^{(m-1)}, \mathbf{L}^{(m)}; \mathbf{\Theta}^{(m)}) \\
\Tilde{\mathbf{H}}^{(m)} & = & Pool(\mathbf{H}^{(m)}, \GG^{(m+1)}) \, ,
\end{array}
\right.
\end{equation}
where $\mathbf{L}^{(m)} \in \R^{N_{m} \times N_{m}}$ is the graph Laplacian of $\GG^{(m)}$ and $\Tilde{\mathbf{H}}^{(m-1)} \in \R^{N_{m} \times F_{m-1}}$.

Let us now specify how spectral-based GCNs like ChebNet can be used in graph signal reconstruction with missing nodes in a given subset $I$ for $H=0$. Since these networks operate convolution on the whole graph signal, we first insert $0$ at missing nodes $(\mathbf{x}_{I,t})_t$ to build a complete graph signal $(\Tilde{\mathbf{x}}_t)_t$ defined as
$$
\Tilde{\mathbf{x}}_t := 
\left\{
\begin{array}{ccl}
\Tilde{\mathbf{x}}_{I^c,t} & = & \mathbf{x}_{I^c,t} \\
\Tilde{\mathbf{x}}_{I,t} & = & 0 \, ,
\end{array}
\right.
$$
and we input $(\Tilde{\mathbf{x}}_t)_t$ to a GCN network. The completion with $0$ is   reasonable as graph convolution amounts to a weighted signal summation, thus $0$ values will not contribute into the sum on the first {\it GConv} layer.  The target (ground truth) of output $\hat{\mathbf{x}}_{I,t}$ which corresponds to the input sample $\Tilde{\mathbf{x}}_t$ should be $\mathbf{x}_{I,t}$. Thus, we set the output dimension as $|I| = p$, and define the loss component from sample $t$ as $\|\mathbf{x}_{I,t} - \hat{\mathbf{x}}_{I,t}\|_{\ell_2}^2$. In this way,  the reconstruction problem is cast as prediction at a graph-level. 


When using longer history $\mathbf{x}_{I^c,t}^H$ (with $H > 0$) to recover $\mathbf{x}_{I,t}$, we construct a complete signal input $\Tilde{\mathbf{x}}^H_t$ in a similar way, by defining 
$$
\Tilde{\mathbf{x}}^H_t := (\Tilde{\mathbf{x}}_t, \Tilde{\mathbf{x}}_{t-1}, ..., \Tilde{\mathbf{x}}_{t-H}) \in \R^{N \times (H+1)}, \quad \mbox{where} \quad \Tilde{\mathbf{x}}_{t-l} :=
\left\{
\begin{array}{ccl}
\Tilde{\mathbf{x}}_{I^c,t-l} & = & \mathbf{x}^H_{I^c,t-l} \\
\Tilde{\mathbf{x}}_{I,t-l} & = & 0
\end{array}
\right.
, l = 0, 1, ..., H.
$$

In our application, adding temporal analyzing modules, such as long short-term memory layer or gated recurrent unit, does not improve the model performances. 
Besides, in practice, we split the dataset chronologically into three time intervals $t = 1, ..., T_{tv}$, $t = T_{tv} + 1, ..., T_0$, and $t = T_0 + 1, ..., T_1$ as training, validation, and test set, denoted by $S_{tn}$, $S_{vd}$, and $S_{tt}$, respectively. The training set is used to update the network parameters, by sequentially minimizing the training loss 
$$\frac{1}{|S_{tn}|} \sum_{t \in S_{tn}} \|\mathbf{x}_{I,t} - \hat{\mathbf{x}}_{I,t}\|_{\ell_2}^2,$$ while the validation set is used to determine  early stopping, meaning that training  is stopped when the validation loss $\frac{1}{|S_{vd}|} \sum_{t \in S_{vd}} \|\mathbf{x}_{I,t} - \hat{\mathbf{x}}_{I,t}\|_{\ell_2}^2$ starts to increase. More details of early stopping will be given in Subsection \ref{sec:slcNetDP} and \ref{sec:netConfig}. The test set is only used to evaluate the model performance at the last step. We call such a network with $p$ output neurons (trained by  input turned-on sensors and output turned-off sensors) a {\it  prediction network with missing set $I$}, that is denoted by $f^{\GG}_{\Theta}$. 

In the following subsections, we now introduce two types of {\it selection networks} denoted by $f'^{\GG}_{\Theta}$ for the purpose of choosing an optimal set $\hat{I}$ of sensors to be turned off.

\subsection{Regularization by $\ell_1$ norm in sensor selection}
\begin{figure}[!htb]
    \centering
    \includegraphics[width=0.8\textwidth]{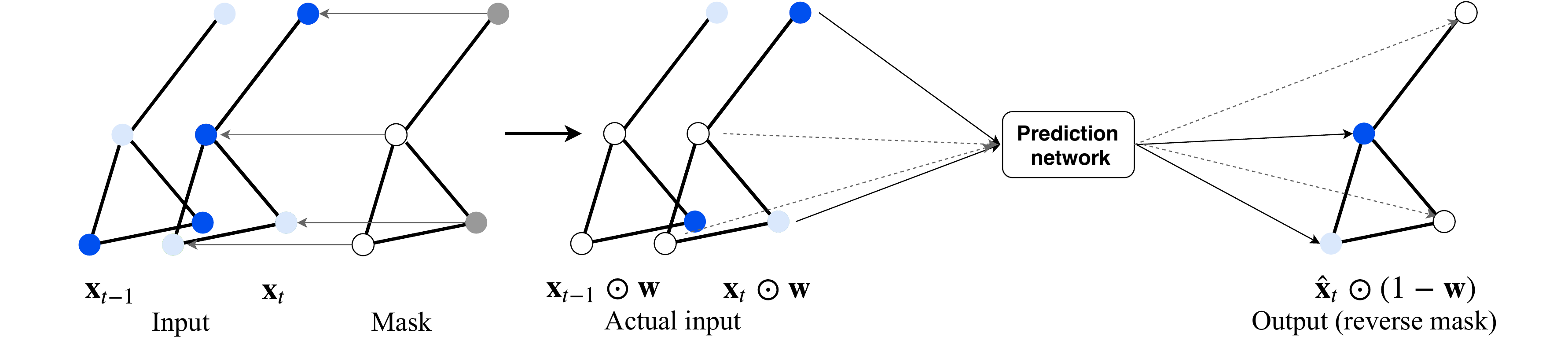}
    \caption{\small \textit{The mechanism of masking with $\ell_1$ regularization.} Mask weights $\mathbf{w} = (w_1, w_2, ..., w_N)$ are trainable network parameters playing the role of a selection vector whose sparsity is controlled by $\ell_1$ regularization.}
    \label{fig:masking}
\end{figure}

The first proposed selection network is based on $\ell_1$ norm regularization, which is a standard penalty term in statistics for variable selection in the so-called {\it Lasso} regression. Figure \ref{fig:masking} illustrates its use in transforming a prediction network into a selection network.

The prediction network $f^{\GG}_{\Theta}$ that we consider takes the complete graph signals as input, and outputs an arbitrary numbers of flexible targets. 
Thus, to transform $f^{\GG}_{\Theta}$ into a selection network, we first change the output dimension from $p$ to $N$. Then, we input $\mathbf{x}_{t}^H =  (\mathbf{x}_t, \mathbf{x}_{t-1}, ..., \mathbf{x}_{t-H}) \in \R^{N \times (H+1)}$ to the network, with observed data at {\it all sensors}. To add a selection rule, we insert a mask layer between the input layer and the first GConv layer. This layer consists of $N$ trainable parameters, referred to as mask weights, and denoted by $\mathbf{w} = (w_1, w_2, ..., w_N)$. We define the forward propagation on this layer as 
$$
\mathbf{x}_{t-l} \odot \mathbf{w}, \, l = 0, 1, ..., H, 
$$
where $\odot$ is the \textit{Hadamard product}, which performs element-wise multiplication between two vectors. Thus, $(\mathbf{x}_t\odot \mathbf{w}, \mathbf{x}_{t-1}\odot \mathbf{w}, ..., \mathbf{x}_{t-H}\odot \mathbf{w}) \in \R^{N \times (H+1)}$ becomes the  input to the prediction network of $f^{\GG}_{\Theta}$, and every mask weight $w_i$ is associated to a single node $V_i$. We denote the transformed network by $f'^{\GG}_{\Theta}$. To make $\mathbf{w}$ a {\it selection vector}, we impose two weight constraints, which read as
\begin{equation}\label{eq:l0 norm and binary constraint}
\|\mathbf{w}\|_{\ell_0} = p \quad \mbox{and} \quad \mathbf{w} \in \{0,1\}^{N}.
\end{equation}
Additionally, we define the loss function as 
\begin{equation}\label{eq:loss function0 masking}
\frac{1}{|S_{tn}|} \sum\limits_{t \in S_{tn}} \|\mathbf{x}_t\odot(1-\mathbf{w}) - \hat{\mathbf{x}}_t\odot(1-\mathbf{w})\|_{\ell_2}^2,
\end{equation}
where $\hat{\mathbf{x}}_t = f'^{\GG}_{\Theta}(\mathbf{x}_t^H) \in \R^N$ is the network output. 
In this way, it is expected that minimizing the objective function (\ref{eq:loss function0 masking}) over both the GCN parameters and mask weights satisfying the constraints \eqref{eq:l0 norm and binary constraint}, leads to  mask weights which selects $p$ sensors having a low reconstruction error. Accordingly, the trained selection vector $\mathbf{w}^*$ indicates an optimal set of sensors to be turned off as $\hat{I} = \{i :   w_i^* = 0\}$. 

Nevertheless, as discussed in Section \ref{sec:notation}, the constraints (\ref{eq:l0 norm and binary constraint}) make the optimization problem combinatorial. Thus, following a standard constraint relaxation, see e.g. \cite{chepuri2017graph, JoshiB09}, we use box and $\ell_1$ norm constraints instead, which read as
\begin{equation}\label{eq:l1 norm and box constraint}
\|\mathbf{w}\|_{\ell_1} = t, \quad \mathbf{w} \in [0,1]^{N}.
\end{equation}
Because $\|\mathbf{w}\|_{\ell_1} = p$ does not amount to $p$ zero weights,  we replace $p$ with a hyperparameter $t > 0$, and we discuss the way to find the $p$ optimal selected sensors later on.
In practice, there are three main ways to implement weight constraints for neural networks, which are as follows.
\begin{enumerate}
\item \textit{Constraint enforcement.} After each optimization step, enforce the updated weight to satisfy the constraints. This can be used, for instance, with a \textit{max-norm} constraint $\|\mathbf{w}\|_{\ell_2} \leq c$, by projecting the weight vector $\mathbf{w}$ onto the surface of a ball of radius $c$, every time its updated norm is larger than $c$. 

\item \textit{Lagrangian.} Eliminate the constraint by adding the corresponding penalty term weighted by a \textit{Lagrange multiplier} in objective function.

\item \textit{Parametrization.} Parametrize the weight to avoid the constraint. For example, the two constraints $\sum_i w_i = 1, \; w_i \geq 0$ can be eliminated by parametrizing $w_i$ using a \textit{softmax} transform defined as $w_i = \frac{\exp(u_i)}{\sum_i \exp(u_i)}$ where the $u_i$'s are unconstrained parameters.
\end{enumerate}

We have chsoen to enforce the box constraint using the projection function $max(min(w, 1), 0)$, and we move the $\|\mathbf{w}\|_{\ell_1}$ constraint into objective function (\ref{eq:loss function0 masking}) with a Lagrange multiplier $\lambda > 0$. This results into the following loss function:
\begin{equation}\label{eq:loss function masking}
\begin{aligned}
    &\frac{1}{|S_{tn}|} \sum\limits_{t \in S_{tn}} \|\mathbf{x}_t\odot(1-\mathbf{w}) - \hat{\mathbf{x}}_t\odot(1-\mathbf{w})\|_{\ell_2}^2 + \lambda \|\mathbf{w}\|_{\ell_1} \\
    = \quad &\frac{1}{|S_{tn}|} \sum\limits_{t \in S_{tn}} \sum\limits_{i = 1}^N (1-w_i)^2(x_{it} - \hat{x}_{it})^2 + \lambda \sum\limits_{i = 1}^N |w_i|.   
\end{aligned}
\end{equation}

Thanks to the $\ell_1$ penalty in the loss function (\ref{eq:loss function masking})  some of entries in the trained selection vector $\mathbf{w}^*$ will be equal to zero. These entries are then interpreted  as the sensors to be turned-off. 
In practice, it is hard to find a single value of $\lambda$  which shrinks exactly $p$ weights $w_i^*$ to $0$. Thus, we rather rely of the notion of \textit{Lasso path} to find a {\it ranked vector} of $p$ selected sensors. To do this, we first train the selection net $f'^{\GG}_{\Theta}$ independently with a grid of $\lambda$'s, denoted by $\lambda_1 < \lambda_2 < ... < \lambda_n$. Then, we collect the trained vector of mask weights for each  value of $\lambda$ in this grid, that we denote by $\mathbf{w}^*(\lambda), \; \lambda = \lambda_1, ..., \lambda_n$. The $N$ curves $-\log(\lambda) \mapsto w_i^*(\lambda)$ for $i = 1, ..., N$ define the so-called \textit{Lasso path}.

As $\lambda$ increases the weights $(w_i^*(\lambda))_i$ will tend to zero. Hence, the first $p$ weights reaching  zero correspond to the first $p$ selected sensors to be turned off. We do not visualize the lasso path and select sensors manually. Instead, for each sensor $i$, we compute the number of elements in the set $F_i := \{\lambda: w_i^*(\lambda) < \epsilon_0, \; \lambda = \lambda_1, ..., \lambda_n\}$, where $\epsilon_0$ is a small value close to zero. The weights  smaller than $ \epsilon_0$ are considered to be zero in numerical experiments.
The first $p$ selected sensors $i_{(1)}, ..., i_{(p)}$ are the ones satisfying $F_{i_{(1)}} \geq ... \geq F_{i_{(p)}}$. This trick is robust against the randomness from network training, and it works in the situation where no $\lambda$ selects exactly $p$ sensors. 

Lastly, when training the selection network, we do not use a validation set to indicate early stopping. The main concern is to obtain weights close zero, the training loss (\ref{eq:loss function masking}) has to approach its local minima w.r.t. all the network parameters. Hence, the validation loss defined either as 
$$
\frac{1}{|S_{vd}|} \sum\limits_{t \in S_{vd}} \|\mathbf{x}_t\odot(1-\mathbf{w}) - \hat{\mathbf{x}}_t\odot(1-\mathbf{w})\|_{\ell_2}^2 + \lambda \|\mathbf{w}\|_{\ell_1}
$$
or as
$$
\frac{1}{|S_{vd}|} \sum\limits_{t \in S_{vd}} \|\mathbf{x}_t\odot(1-\mathbf{w}) - \hat{\mathbf{x}}_t\odot(1-\mathbf{w})\|_{\ell_2}^2$$
will easily pass its minimal value leading to overfitting. Thus, we stop training at a pre-given maximal epoch\footnote{An epoch amounts to a series of optimization steps which traverse all samples in training set once.} (the same for all $\lambda$'s) which is essential another hyperparameter. The whole transform and selection procedure is class {\it masking with $\ell_1$ regularization}. 

\subsection{Dropout in sensor selection}\label{sec:slcNetDP}
Let us now describe another way to construct a selection network which does not bring any additional hyperparameter as in the previous case with $\ell_1$ regularization. Rather than outputting the optimal set $\hat{I}$, the primary goal of this method is to  score the {\it learned predictability} of all sensors which represent, for each sensor, a scoring of its potential reconstruction error by a prediction network. The selection of the sensors to be turned-off is then induced by these scores.

The challenge of sensor selection in the regression models described previously is that searching over all possible set $I$ is not feasible in reasonable time. However, by leveraging dropout technique, we can realize this exhaustive search to a large extent, at the cost of insufficient training in the reconstruction of each $I$. We first review  dropout \cite{srivastava2014dropout} which is a regularization technique designated to prevent overfitting. 

The best way to reduce overfitting is model combination, which means  fitting all possible models first, and then, for each test sample, to aggregate the predictions from all trained models as the final prediction. Obviously, for a class of complex models such neural networks, independent training and explicit aggregation are not realistic. Starting from this motivation, dropout makes a single trained neural network combine exponentially many different network models w.r.t. the number of neurons  in an approximate way, and  it outputs the averaged result.

To be more precise, we illustrate this technique with an example. 
\begin{figure}[h]
    \centering
    \includegraphics[width=0.9\textwidth]{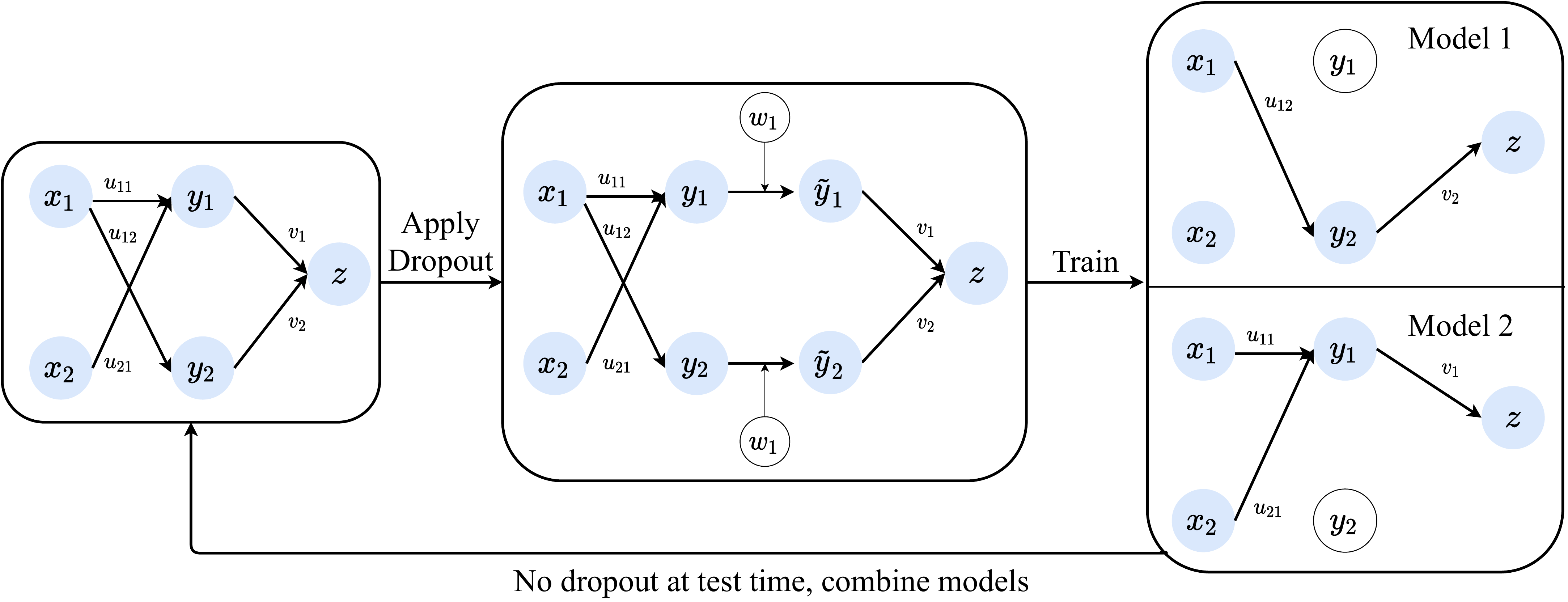}
    \caption{\small Training a network with dropout can be viewed as training a collection of network models which share weights. At test time, all neurons are turned on. The final trained net without dropout outputs the aggregated prediction from these models.}
    \label{fig:dropout}
\end{figure}
Figure \ref{fig:dropout} is a small neural network, which consists in one input layer, one hidden layer and one output layer, whose corresponding neurons are denoted by $\mathbf{x} = (x_1, x_2)$, $\mathbf{y} = (y_1, y_2)$ and $z$, respectively. Without dropout, the network forward propagation is 
$$
z = \sigma_2(v_1y_1 + v_2y_2), \quad y_1 = \sigma_1(u_{11}x_1 + u_{21}x_2), \quad y_2 = \sigma_1(u_{12}x_1), 
$$
where $v_1, v_2, u_{11}, u_{21}, u_{12}$ are weights, which are trainable parameters.
In this example, we omit bias terms.
If we apply dropout on the hidden layer, the forward propagation becomes
$$
z = \sigma_2(v_1\Tilde{y}_1 + v_2\Tilde{y}_2),\quad \Tilde{y}_1 = w_1y_1, \quad \Tilde{y}_2 = w_2y_2, \quad y_1 = \sigma_1(u_{11}x_1 + u_{21}x_2), \quad y_2 = \sigma_1(u_{12}x_1),
$$
where $\mathbf{w} = (w_1, w_2)$ is the dropout vector. The components of dropout vector are independent Bernoulli random variables, each of which has probability $q$ of being $0$, which is referred to as the \textit{dropout rate}. Whether to update the weights connecting to the neurons with dropout is determined by the realization of $\mathbf{w}$ and gradient calculation. We denote the loss component of a sample as $J(z)$. Then, the associated gradient components of trainable parameters are computed using the \textit{chain rule} as follows:
$$
\frac{\partial J(v_1)}{\partial v_1} = \frac{\partial J(z)}{\partial z} \frac{\partial z(v_1)}{\partial v_1} = \frac{\partial J(z)}{\partial z} \sigma_2'(v_1\Tilde{y}_1 + v_2\Tilde{y}_2)w_1y_1
$$
$$
\frac{\partial J(u_{i1})}{\partial u_{i1}} = \frac{\partial J(z)}{\partial z} \frac{\partial z(y_1)}{\partial y_1}\frac{\partial y(u_{i1})}{\partial u_{i1}} = \frac{\partial J(z)}{\partial z} \sigma_2'(v_1\Tilde{y}_1 + v_2\Tilde{y}_2)v_1w_1 \frac{\partial y(u_{i1})}{\partial u_{i1}}, \, i = 1,2. 
$$
The total loss of an optimization step is essentially the sum of loss components of all samples used, and the gradient is accordingly the sum of all the associated components. Thus, if $w_1$ is sampled as $0$ at certain optimization steps, all three gradients above are $0$ and the weights connecting to $y_1$ will not be updated. The neuron $y_1$ is thus \textit{turned off} at this training phase. 
Sampling $\mathbf{w} = (0, 1)$ amounts to sampling and training model $1$ in Figure \ref{fig:dropout}. Similarly, $\mathbf{w} = (1, 0)$ samples model $2$. With less probability, the training with dropout can still explore the model with no hidden layer neurons, or with both. This mode trains each model insufficiently in practice. To see this, we furthermore denote the neurons of the three layers for each sample $t$ by $\mathbf{x}_t$, $\mathbf{y}_t$ and $z_t$. Then, the network loss to be optimized at each updating step of the parameters  is
$$
\frac{1}{|S|}\sum\limits_{t \in S}J(z_t) = \frac{1}{|S|}\sum\limits_{t \in S}C(z_t,z_t^0),
$$
where $S$ denotes the actual training set used at each step with $|S|$ samples, commonly a subset of the entire training set, which literature refers to as \textit{batch}, $C(\cdot, \cdot)$ is some cost function, $z_t^0$ is the ground truth value to be compared with prediction $z_t$. The weights  $\mathbf{w}$ being re-sampled at each optimization step, each batch contributes to one of the correspongin sampled model. The parameters of new model are initialized as the results of preceding training phase. The ones connecting to turned-on neurons are shifted towards the potential independent training results of this model. By doing this, the training needs to find the optimal weight values which are shared across models and gives the aggregated model prediction the best performance. When using the network in prediction (test time), all neurons are present with scaled-down weights, which indicates multiplying the outgoing weights of the dropout neurons by $1-q$. This ensures the expected neuron output in training remains the same as the actual one at test time. The training with dropout and the test without dropout implicitly combine considerably many models into one single net.

\begin{figure}[h]
    \centering
    \includegraphics[width=\textwidth]{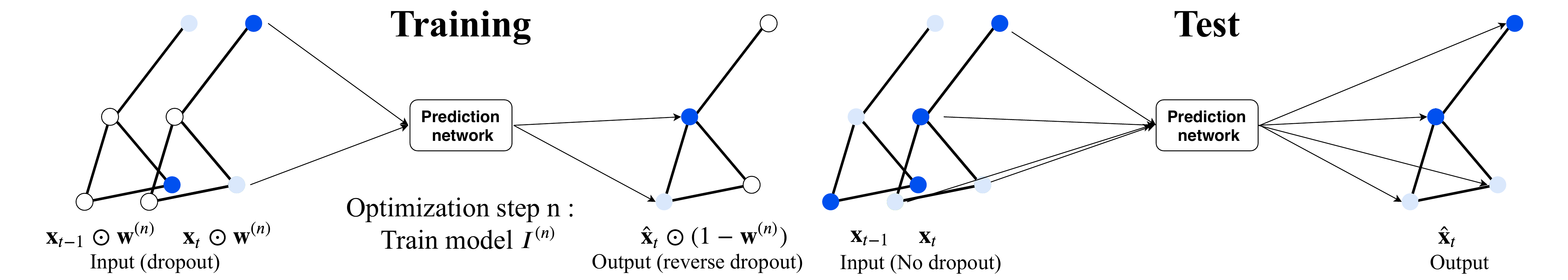}
    \caption{\small \textit{The mechanism of dropout used in sensor selection,} where $\mathbf{w}^{(n)}$ is the dropout vector sampled at the optimization step $n$. The training phase is to change the network parameters to their optimal values based on gradient descent. The test phase is to use these optimal parameters on unseen inputs and compute outputs to target nodes. Thus, the prediction on validation and test sets consists in testing trained networks.}
    \label{fig:dropout used in sensor selection}
\end{figure}

Back to the sensor selection problem, we first align some terms with the model combination situation. For a given prediction network (such as ChebNet) for  graph signal reconstruction with missing nodes, we consider it as a given model (referred to as reconstruction model $I$) when it is used in the reconstruction for a given missing set $I$. Thus, the best way of selecting sensors is to train all possible models first, and then to select the best one $\hat{I}$ with the best prediction performance. This amounts to train $\mathcal{O}(N^p)$ models independently, which is not feasible. Let us now explain how dropout can be used to combine such models for the purpose of sensor selection. Figure \ref{fig:dropout used in sensor selection} illustrates how to use dropout to transform a prediction network into a selection network. 

For the transformation of dropout method, we also first change the output dimension of a prediction network from $p$ to $N$, and the full signal $\mathbf{x}_{t}^H$ is set as input. We denote this network by $f'^{\GG}_{\Theta}$. The main difference is that the selection vector $\mathbf{w}$ becomes a random (and thus non-trainable) dropout vector $\mathbf{w}^{(n)}$ at each iteration $n$ of the learning process.  When training the selection net $f'^{\GG}_{\Theta}$, we apply dropout on the input layer as
$$
\mathbf{x}_{t-l} \odot \mathbf{w}^{(n)}, \, l = 0, 1, ..., H, 
$$
where $\mathbf{x}_{t-l} \in \R^N$ are the input signals, and $\mathbf{w}^{(n)} = \left(w_1^{(n)}, w_2^{(n)}, ..., w_N^{(n)}\right) \in \R^N$ is the dropout vector. We define the loss for a given $\mathbf{w}^{(n)}$ as
$$
\frac{1}{|S_n|} \sum\limits_{t \in S_n} \left\|\mathbf{x}_t\odot\left(1-\mathbf{w}^{(n)}\right) - \hat{\mathbf{x}}_t\odot\left(1-\mathbf{w}^{(n)}\right)\right\|_{\ell_2}^2, 
$$
where $S_n$ is the batch used at step $n$, and $\hat{\mathbf{x}}_t = f'^{\GG}_{\Theta}(\mathbf{x}_t^H) \in \R^N$ is the network output. 
This definition amounts to applying the exact reverse dropout at the output layer. Each component of the dropout vector corresponds to a single node in the graph $\GG$.
At each training phase $n$ the sensors $V_i$ corresponding to $w_i^{(n)} = 0$ are turned off, while reversely the  neurons outputting the predictions for the sensors $V_j$ associated to $w_j^{(n)} = 1$ are turned off. Thus, at phase $n$, one trains a prediction network with missing set $I^{(n)} = \{i : w_i^{(n)} = 0\}$. The dropout rate $q$ is fixed as the missing rate $p/N$, and this transformation does not introduce any extra hyperparameter (given the known value $p$).
In practice, the sampled model changes across batches,
and we use gradient descent as the optimization algorithm, which does not use the previously computed gradients to update network parameters as in other algorithms such as \textit{Adam}. 

At the test phase, to aggregate all sampled reconstruction models $I^{(n)}$,  we do not apply dropout in the network $f'^{\GG}_{\Theta}$ so as to turn-on all sensors. Because the trained network takes the input $\mathbf{x}_t^H$ and outputs the prediction of $\mathbf{x}_t$, it becomes an autoencoder when $H = 0$ whose output is supposed be the recovery of input. Nevertheless, $f'^{\GG}_{\Theta}$ does not copy the input as the corresponding output due to the dropout and reverse dropout processes. 
Instead, it will output the  learned predictability of sensors meaning that the outputs $\hat{x}_{it}$ at some sensors  will be good recovery of their target values $x_{it}$. To the contrary, for other sensors the target signals are not properly recovered  by the trained selection network. 


To be more precise, we propose to {\it score the sensors} as follows. We input  graph signals $\mathbf{x}_t^H$ from  the validation set to the trained selection network, and we obtain as output the prediction $\hat{\mathbf{x}}_t = f'^{\GG}_{\Theta}(\mathbf{x}_t^H)$. For each sensor $i, \; V_i \in \mathcal{N}$, the value 
$
C((x_{it})_t , (\hat{x}_{it})_t), \; t \in S_{vd},
$
quantifies its {\it learned predictability} where $C(\cdot, \cdot)$ is some measure of accuracy. In our application, we use the $R^2$ measure defined as
\begin{equation}\label{eq:R2formula}
R^2_i = 1 - \frac{\sum_{t \in S_{vd}}(x_{it} - \hat{x}_{it})^2}{\sum_{t \in S_{vd}}(x_{it} - \bar{x}_{i})^2},
\end{equation}
where $\bar{x}_{i} = \sum_{t \in S_{vd}} x_{it}$. Since $\hat{x}_{it}$ is not the fitted value from ordinary least squares regression and given that the data used to estimate the model parameters are not the one used to compute the score, the value of $R^2_i$ ranges between $-\infty$ to $1$ (instead of $0$ to $1$ as in standard linear regression). A higher value of $R^2_i$ means that the network recovers the signal of sensor $i$ with a lower error, thus it is better predictable. GCN with dropout method thus yields a sequence of sorted $R^2$ scores $R^2_{i_{(1)}} \geq R^2_{i_{(2)}} \geq ... \geq R^2_{i_{(N)}}$ which indicates a ranking of priority to turn-off sensors, with the top $p$ sensors $i_{(1)}, ..., i_{(p)}$ as the resulting optimal set of sensors to be turned-off.  When $C(\cdot, \cdot)$ is the MSE measure, that is
$$
MSE_i = \frac{\sum_{t \in S_{vd}}(x_{it} - \hat{x}_{it})^2}{|S_{vd}|},
$$
lower scores indicates better predictable sensors. 


Besides, there are two technical details to elaborate.
First, as in the training of prediction network, we also use early stopping in the dropout method. At every epoch, we test once the network (with the current optimized values for its parameters) on the validation set,  we compute its loss $\frac{1}{|S_{vd}|} \sum_{t \in S_{vd}} \|\mathbf{x}_t - \hat{\mathbf{x}}_t\|_{\ell_2}^2$,  and we stop training when the validation loss starts to increase. Thus, the number of missing set $I$'s that is explored during training is equal to  the training set size times the batch size divided by the number of epochs.
Second, during the test phase, we do not multiply the input $\mathbf{x}_t^H$ (or/and output $\hat{\mathbf{x}}_t$) by $1 - q/N$ (or/and $q/N$) to keep the expected sum of inputs (or/and outputs) in training the same as the one in testing. Indeed, we have found that such additional operations deteriorates the performances of reconstruction on selected sensors.

\subsection{Final training after sensor selection}

Lastly, we discuss the common features between these two ways  of building a selection network (using either $\ell_1$ regularization or dropout). Both networks have $N$ output targets, and their input and output are complete graph signals with no zeros inserted. In practice, we use the training and validation sets to learn the parameters of these networks. After this training step, an optimal set $\hat{I}$ of sensors to be turned-off can be determined. Then, we set the signals of  those selected sensors in training and validation sets to zero, and we  train the prediction network corresponding to the missing set $\hat{I}$. To evaluate its performances, we finally use data from the test set in a similar way.

\section{Numerical experiments} \label{sec:num}
In this section, we evaluate our sensor selection approches with real bike sharing datasets. We first introduce the datasets and the experiment settings. Then, we report  numerical results.

\subsection{Datasets}\label{sec:datasets}
\begin{figure}[h]
    \centering
    \includegraphics[width=0.8\textwidth]{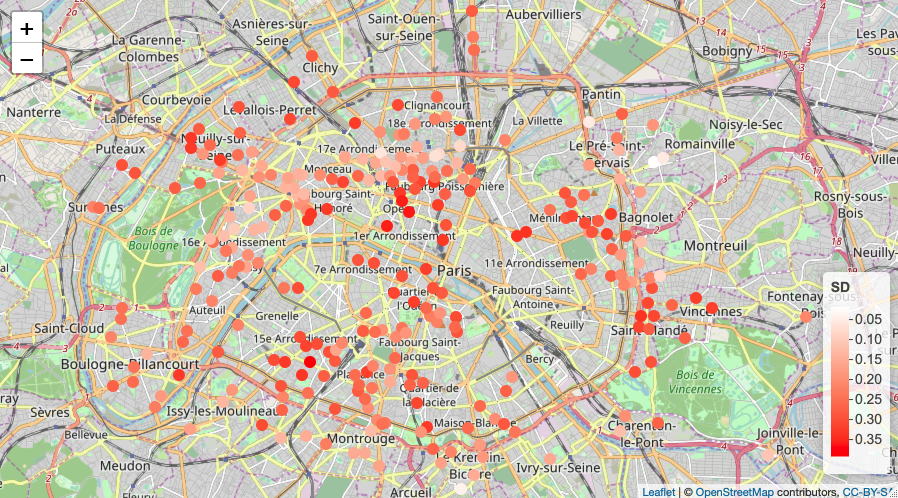}
    \caption{\small \textit{Bike-sharing sensor network in Paris.} Network time series profile: $274$ nodes, $3373$ edges, $4417$ hours. The time series of darker sensors have higher standard deviation.}
    \label{fig:Paris sensor network}
\end{figure}

\begin{figure}[h]
    \centering
    \includegraphics[width=0.8\textwidth]{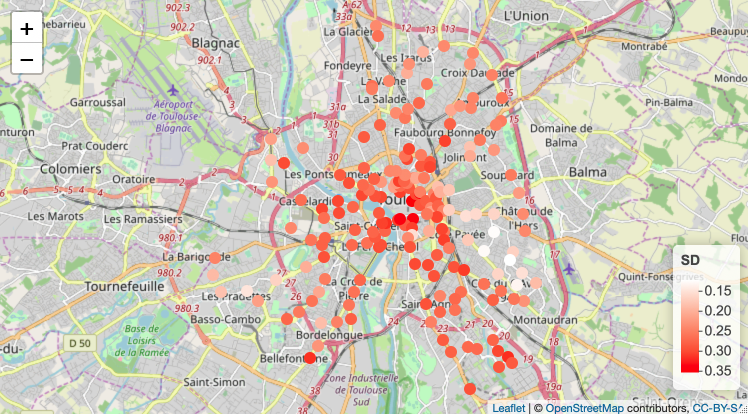}
    \caption{\small \textit{Bike-sharing sensor network in Toulouse.} Network time series profile: $185$ nodes, $2318$ edges, $4305$ hours. The time series of darker sensors have higher standard deviation.}
    \label{fig:Toulouse sensor network}
\end{figure}

The data come from the bike-sharing networks in two French cities, Paris and Toulouse\footnote{The original datasets are available at \url{https://maxhalford.github.io/blog/a-short-introduction-and-conclusion-to-the-openbikes-2016-challenge/}. The clean datasets as well as the codes of this paper can be found at \url{https://github.com/yiyej}}. In each network, the dataset contains the information on all bike stations (each attached a sensor), which includes their latitudes and longitudes, as well as two time series named \textit{bikes} and \textit{spaces} both of which are indexed by \textit{moment} with specific date and time. The variables \textit{bikes} and \textit{spaces} record the number of unused and used bikes of a station respectively. We first delete the bike stations with too few observations. 
Then, for each station, we collect all values of (\textit{bikes}+\textit{spaces}), with its highest value denoted by \textit{max bikes}. We calculate the correction rate defined as the frequency of \textit{max bikes} in   (\textit{bikes}+\textit{spaces}). We keep the stations whose correction rates are larger than the threshold $r_c$, where $r_c$ is $0.9$ for Paris dataset and $0.8$ for Toulouse dataset.
For the remaining stations, we construct the new time series \textit{bikes}/\textit{max bikes} which thus represents the ratio of available bikes at each station. On the other hand, because the recording moments differ across stations, we define the unified time interval from the $0.995$ quantile of all starting moments, to the $0.005$ quantile of all ending moments, and create time stamp every hour within this interval. We interpolate all new time series at these time stamps. Finally, the network time series of Paris consists of $274$ nodes with $4417$ hours, and the one of Toulouse consists of $185$ nodes with $4305$ hours. The spatial locations of the sensors are displayed in Figure \ref{fig:Paris sensor network} and Figure \ref{fig:Toulouse sensor network} with colors related to the standard deviation of each time series.


\subsection{Graph construction}
For both graph kernel ridge regression and the ChebNet, we build  the  graph $\GG$ by taking the $k$-nearest neighbors of each sensor using their geographical coordinates. More specifically, we first compute the Euclidean distance between nodes (sensors) using the latitude and longitude, denoted as $d(\cdot, \cdot)$. Then, for each node $V_i$, we take its $k_0$ nearest neighbors, denoted as $V_{i_1}, V_{i_2}, ..., V_{i_{k_0}}$, with the associated distances $d(V_i, V_{i_1}), d(V_i, V_{i_2}), ..., d(V_i, V_{i_{k_0}})$. We add edges $(i, i_s), \, i = 1, ..., N, s = 1, ..., {k_0}$ into $\mathcal{E}$, with weights calculated as 
$$
a_{i i_s} = \exp\left(-\frac{d^2(V_i, V_{i_s})}{\sigma_i\sigma_{i_s}}\right)
$$
where $\sigma_j$ is the local scale of node $V_j$, that is chosen as $\sigma_j = d(V_j, V_{j_{k_1}})$, with $V_{j_{k_1}}$the $k_1$-th nearest neighbour of $V_j$. This self-tuning approach is proposed in \cite{zelnik2005self}. In our experiments, we took $k_0 = 20$ and $k_1 = 7$. For the pairs $(i, j) \notin \mathcal{E}$, $a_{ij}$ is defined as $0$. Thus, we have built the adjacency matrix $A = (a_{ij})_{i,j}$ which defines the combinatorial Laplacian for all methods.

\subsection{Stationary time series}
The models in Section \ref{sec:linest} and Section \ref{sec:kernel} assume that the network time series $\mathbf{x}_t$ are stationary, which is usually not satisfied by real data. Therefore, we detrend the time series beforehand. Also, we are interested in the recognition of  patterns, and we thus scale the detrened time series as discussed at the end of Subsection \ref{sec:linH=0}.
Let $T_1$ denote the total recorded hours. We split the whole time series  into three intervals, denoted as $1 \leq t \leq T_{tv}$, $T_{tv} < t \leq T_0$ and $T_0 < t \leq T_1$,  that are used as training , validation  and test sets, respectively. The trend and scale are computed from the training set, then applied to all three sets. We evaluate the performance of models on the pre-processed test set, but we visualize the estimators of time series with the trends and scales added and multiplied back.

\begin{figure}[h]
    \centering
    \includegraphics[width=\textwidth]{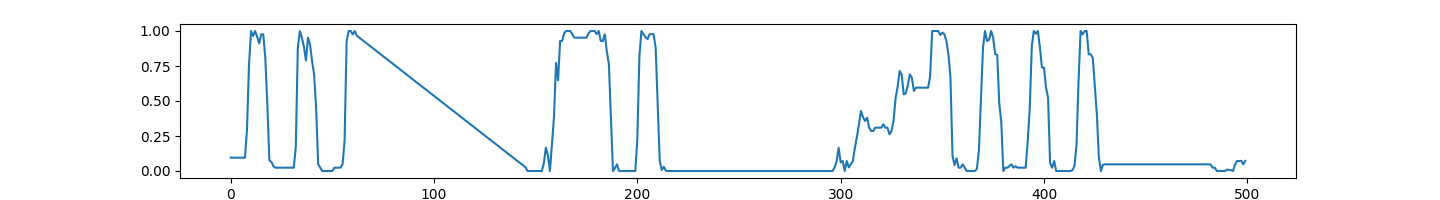}
    \includegraphics[width=\textwidth]{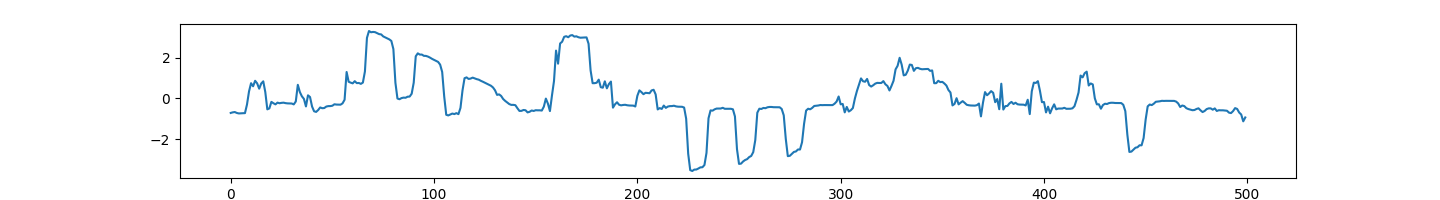}
    \caption{\small \textit{Data preprocessing: detrend and scale.} Top: the original time series. Bottom: processed time series.}
    \label{fig:data preprocessing}
\end{figure}

Because after interpolation, the network time series $(x_{it})_{i,t}$ is recorded every hour, we extract its weekly profile as trend, which reads as
$$
P_i(m) = \frac{\sum\limits_{n = 0}^{N_m-1} x_{i(m+168n)}}{N_m}, \quad m = 1, 2, ..., 168,
$$
where $N_m = \max\limits_n\{n: m+168n \leq T_{tv}\}$, which is the number of moment $m$ contained in the training set. We subtract the weekly profile from each individual time series as $x_{it} - P_i(t \, \mbox{mod} \, 168), \, t = 1, 2, ..., T_1$. Then, we divide the detrended series by its standard deviation to obtain the final pre-processed network time series, still denoted as $x_{it}$.  Figure \ref{fig:data preprocessing} gives an example of a comparison between the original and pre-processed data. We use $x_{it}$ in all models including the neural networks.

\subsection{Searching grid of hyperparameters $H, \gamma, \lambda$}

We use the strategy (\ref{eq: hyperparameter selection}) which is mentioned at the end of Section \ref{sec:kernel} to select hyperparameters in the reconstruction methods. 
We firstly discuss the definition of the searching grid for the graph kernel approach for  $H > 0$ which involves three hyperparameters $H, \gamma$ and $\lambda$. 
However, we do not search optimal values of each parameter independently.
Indeed, the effect of a larger value for $H$ should not be cancelled out by a large value $\gamma$ which makes the similarity decays faster in time in Gaussian kernel $k_{rbf}(t,t') = \exp\left[-\gamma(t -t')^2\right]$, namely $k_{rbf}(0,H) \approx 0$. Therefore, we define the strategy for searching optimal values of $H$ and $\gamma$ as
\begin{enumerate}
\item set the searching values of time lag $H$ as $1, 5, 10$,
\item fix $k_{rbf}(0,H)$ as $r_s$, which leads to $\gamma = -\log(r_s)/H^2$, where $r_s$ is $0.5$ for Paris dataset and $0.3$ for Toulouse dataset.
\end{enumerate}
On the other hand, when $H$ increases, the matrix $K_{\mathcal{S}}^H$ in Algorithm \ref{algo:kernelH>0, CG} becomes larger and more unstable in the sense that it has a greater condition number. This requires taking a larger value of $\lambda$ to speed up the matrix inversion by a better preconditioning. Given that the condition number of $K_{\mathcal{S}}^H + \lambda I_d$ is upper bounded by the one of $K_{\mathcal{N}}^H + \lambda I_d$ (denoted by $\frac{\lambda_{max} + \lambda}{\lambda_{min} + \lambda}$) which is furthermore upper bounded\footnote{For the kernel design (\ref{eq:STKern}), the smallest eigenvalue of $K_{\mathcal{N}}^H$ is exactly $0$. Thus, $\frac{\lambda_{max} + \lambda}{\lambda_{min} + \lambda} = \frac{\lambda_{max}}{\lambda} + 1.$} by $\frac{\lambda_{max} + \lambda}{\lambda} = \frac{\lambda_{max}}{\lambda} + 1$, we use a \textit{power method} to estimate $\lambda_{max}$. This means that after fixing $H$, we set the searching value of $\lambda = a_i\lambda_{max}$, where $a_i = 0.001, 0.00325, 0.0055, 0.00775, 0.01$ for both datasets.

For the kernel approach with $H = 0$ we define the searching values of $\lambda$ in a similar way  with $\lambda_{max}$ defined as the largest eigenvalue of $K_{\mathcal{N}}$. The linear reconstruction model with $H > 0$ is regarded  as a special case of the kernel approach, we use linear ridge regression. The searching grid for the two hyperparameters $H, \lambda$ is defined by the same rules as the kernel case.

\subsection{Neural network configuration and training settings}\label{sec:netConfig}
The ChebNet used in the experiments consists of $1$ GConv layer with no pooling, followed by $3$ FC layers and the output layer. The hyperparameter setting on the first GConv layer in forward propagation (\ref{eq:GConv ChebNet}) is: $K_1 = 50$, $F_1 = 16$, $\mathbf{b}^{(1)}_j$ are all vectors of $N$ independent parameters for $j = 1, 2, ..., F_1$. The layer activation function is \textit{elu}, namely $\sigma(x) = \exp(x) - 1, \; if \; x < 0 \; ; \, x , \; o.w.$. The FC layers contain respectively $128$, $500$, $64$ neurons, all with bias terms. The activation functions on FC layers are \textit{leaky relu} with $\alpha = 0.2$, whose formula is $\sigma(x) = \alpha x$ if  $x < 0 \; ; \, x , \; o.w.$. 

\textit{Early stopping:} we use early stopping in training the prediction network and the selection network with dropout method. Since their actual training set are batches, we test the nets on validation set at every epoch, and collect the sequence of validation losses. For dropout method, at the end of every $5$ epochs, we compute the validation loss mean, and if it is larger than the preceding mean, we stop the training. For the prediction net, at every epoch, we average the current validation loss and its last one, if the mean increases, training stops. We still set the maximal epoch in these cases to avoid long training process. The rest of training settings are as follows:
\begin{description}
\item[-] \textit{Selection net, masking with $\ell_1$ regularization} Optimization algorithm: Adam. Batch size: $50$. Learning rate: $0.005$. $\epsilon_0$: $0.01$. $\lambda$ grid (Paris dataset): $\lambda_1 = 0.0500$, $\lambda_2 = 0.0658$, ..., $\lambda_{20} = 0.3500$. $\lambda$ grid (Toulouse dataset): $\lambda_1 = 0.0200$, $\lambda_2 = 0.0295$, ..., $\lambda_{20} = 0.2000$. Maximal epoch: $50$.
\item[-] \textit{Selection net, dropout method.} Optimization algorithm: gradient descent. Batch size (Paris dataset): $50$. Batch size (Toulouse dataset): $200$. Learning rate (Paris dataset): $0.05$. Learning rate (Toulouse dataset): $0.001$. Maximal epoch: $500$.
\item[-] \textit{Prediction net.} Optimization algorithm: Adam. Batch size (Paris dataset): $1000$. Batch size (Toulouse dataset): $2000$. Learning rate: $0.001$. Maximal epoch: $50$.
\end{description}

We do not use other regularization. We test input history lengths, $H = 0$ and $H = 5$ for the dropout method, and history length $H = 0$ for the masking method.

\subsection{Other experiment settings}
The number $p$ of sensors to be turned off is set as $10\%$ of $N$, which is $27$ for Paris network and $18$ for Toulouse network. Training, validation and test set size are $3776$, $199$ and $442$ hours respectively for Paris dataset, $3288$, $361$ and $641$ hours respectively for Toulouse dataset. 

\subsection{Experiment results}

We use training and validation set to select sensors and fit their reconstruction model to recover the signals from test set. The resulting reconstruction error is measured by MSE. Tables \ref{tab:reconstruction error paris} and \ref{tab:reconstruction error toulouse} report this error for all model classes analyzed in the paper, with different $H$'s.

Meanwhile, to demonstrate the performance of our sensor selection strategy, we compare it to a {\it random sampling approach} that  selects the set  of sensors to be turned off by picking $p$ out of $N$ sensors in a random way. To this end, for each class of reconstruction method, we consider the problem of recovering  $p$ randomly selected turned-off sensors on the test set, and we average the resulting reconstruction error over  $100$ such random sets. The values in parenthesis in Table \ref{tab:reconstruction error paris}, \ref{tab:reconstruction error toulouse} represent these averaged reconstruction errors. For the linear and kernel classes and each value of $H$, we only indicate the lowest error value obtained by the best choice of other hyperparameters tested. For GCN class, we have two selection methods, masking with $\ell_1$ regularization and dropout. We evaluate the dropout method with each of the four values for $H$ on the validation set. As significant differences have not been observed,  we only report experiments related to $H = 0$ and $H = 5$ in the two tables. Besides, since the output of selection network training contains some randomness, we report the results of two independent runs. After having selected the sensors, we again train the prediction network $10$ times, and we report the average of these reconstruction errors. The reported errors are from sensors selected using the $R^2$ measure defined in (\ref{eq:R2formula}). As for the masking method, because its performance is not satisfactory, we report its reconstruction error as well as selection result lastly in Tables \ref{tab:selection results of GCN (MSE) and GCN (Masking)} and \ref{tab:selection results of GCN (MSE) and GCN (Masking), toulouse}, where we also show the results from dropout method using MSE as score. The full list of selected sensors and the hyperparameter values used in the tables can be found in the Appendix.

\subsubsection{Paris Network}\label{sec:paris network}

\begin{minipage}[h]{\linewidth}
\bigskip
\captionof{table}{\small Reconstruction error over the test set of the $27$ (10\% of $274$) selected sensors in Paris network. For each method, the values in parenthesis are the mean reconstruction error from $100$ random sampling sets $I$ of cardinality $p=27$.} \label{tab:reconstruction error paris} 
\begin{tabular}{ C{1.50in} *4{C{0.90in}}}\toprule[1.5pt]
$H$& $0$ & $1$ & $5$ & $10$ \\
\midrule
\bf Linear models &  32.53 (41.04)   & 23.81 (30.45)  &  25.80 (29.60) &  25.85 (30.89)\\
\bf Kernel models &  22.34 (28.90)  & 22.02 (27.83)  &  23.15 (27.57) &  23.18 (27.74)\\
\bf GCNs ($R^2$, run $1$) &  19.31 (29.65) &  / &  18.74 (28.53) & / \\
\bf GCNs ($R^2$, run $2$) &  17.41 (29.65) & /  &  19.56 (28.53) & / \\
\bottomrule[1.25pt]
\end {tabular}\par
\bigskip
\end{minipage}

The result from Table \ref{tab:reconstruction error paris} clearly indicates that GCN with the dropout method yields the best performances in term of reconstruction error on the test set. Moreover, all sensor selection approaches defeat the random sampling strategy (which corresponds to randomly selecting the set  of sensors to be turned off) by a large percentage. These results also indicate that using  more past values (that is increasing $H$) may improve the reconstruction performances.


\begin{minipage}[h]{\linewidth}
\bigskip
\captionof{table}{\small Sensor selection results of the best models in three classes of reconstruction functions. The sensors are indexed from $0$ to $273$ as the order of their columns in \textit{X\_hour\_paris.csv} file.} \label{tab:selection results of 3 best models} 
\begin{tabular}{ C{0.50in} C{1.50in} *4{C{0.70in}}} 
\toprule[1.5pt]
 & \bf \small Selected sensors (in order) & \bf \small Overlap of linear, kernel & \bf \small Overlap of kernel, GCN & \bf \small Overlap of linear, GCN & \bf \small Overlap of the three \\
\hline
\bf Linear & \scriptsize 5, 209, 112, 30, 73, 61, 224, 155, 117, 239, 257, 34, 229, 31, 42, 114, 240, 205, 74, 227, 186, 17, 72, 255, 113, 132, 226. & \multirow{2}{4em}{\small 5, 30, 34, 42, 74, 112, 113, 114, 117, 155, 186, 229. } & & \multirow{2}{4em}{\small 5,  30,  34,  42,  61, 112, 113, 114, 117, 229, 240, 257. } & \multirow{3}{2em}{\small 5,  30,  34,  42, 112, 113, 114, 117, 22. }\\ \cline{2-2}
\bf Kernel & \scriptsize 34, 197, 186, 112, 229, 42, 74, 189, 3, 113, 225, 232, 193, 5, 20, 238, 114, 30, 11, 190, 70, 18, 228, 267, 117, 93, 155. &  & \multirow{2}{4em}{\small 3,   5,  30,  34,  42, 112, 113, 114, 117, 189, 225, 229. } & &\\ \cline{2-2}
\bf GCN & \scriptsize 5, 34,  61, 112, 58, 108, 225, 221, 49, 42, 229, 3, 257, 117, 113, 169, 110, 266, 240, 30, 109, 2, 189, 95, 114, 248, 92. &  & & &\\
\bottomrule[1.25pt]
\end{tabular}\par
\bigskip
\end{minipage}

\begin{figure}[!htb]
    \centering
    \includegraphics[width=0.8\textwidth]{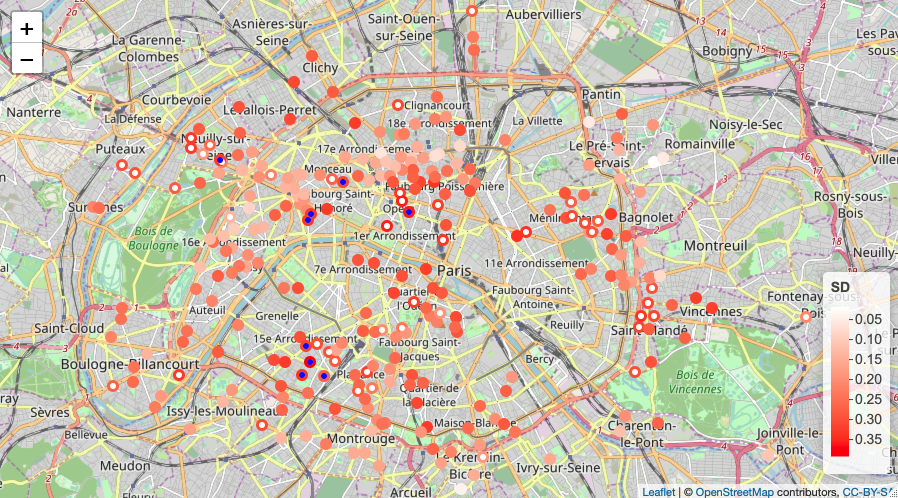}
    \caption{\small \textit{Display of the selected sensors from results in Table \ref{tab:selection results of 3 best models}.} The intersection and union of the three sets of selected sensors (for each of the three reconstruction method) are marked by circles with inside blue and white color respectively.}
    \label{fig:sensor_selection_results_Paris}
\end{figure}

Secondly, in Table \ref{tab:selection results of 3 best models}, we compare the best sets $\hat{I}$ of selected sensors for the following three  reconstruction models which are: linear ridge regression ($H = 1, \lambda = 0.740$), kernel ridge regression ($H = 1, \lambda = 0.428, \gamma = 0.693$), ChebNet ($H = 0$, dropout method with $R^2$, run 2). The results show that some sensors  are always selected across all methods. We visualize these so-called \textit{important sensors} together with all the other selected sensors on a map displayed in Figure \ref{fig:sensor_selection_results_Paris}. We observed that our selection strategies tend to choose  sensors of higher standard deviation in their original data scale, which are likely to contain meaningful patterns. In addition, even though the selected sensors of different models are not exactly the same, they appear in the same neighbourhood of the graph. These facts also support the effectiveness of our sensor selection approaches.

\begin{figure}[!htb]
    \centering
    \includegraphics[width=0.8\textwidth]{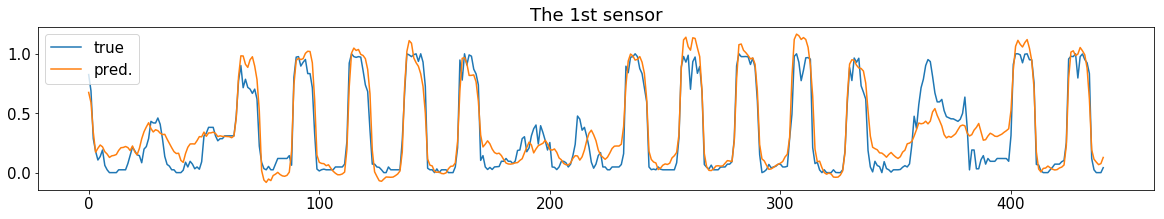}
    \includegraphics[width=0.8\textwidth]{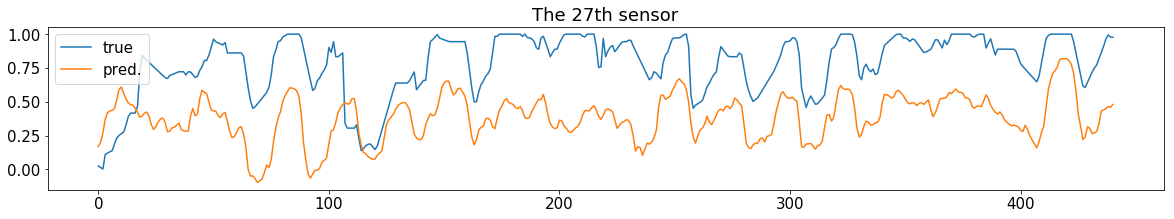}
    \caption{\small \textit{Reconstruction of selected sensors for linear ridge regression ($H = 1, \lambda = 0.740$).}}
    \label{fig:recovered signal lin}
\end{figure}

\begin{figure}[!htb]
    \centering
    \includegraphics[width=0.8\textwidth]{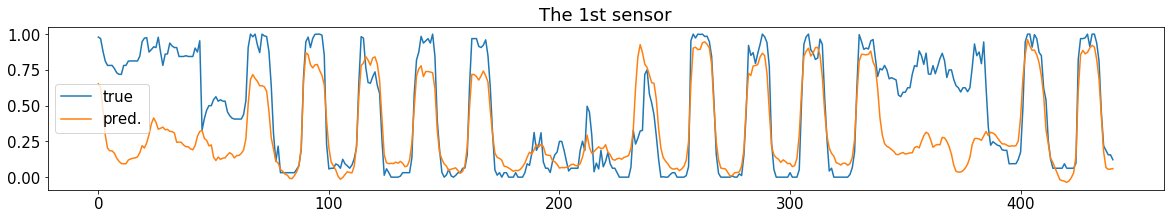}
    \includegraphics[width=0.8\textwidth]{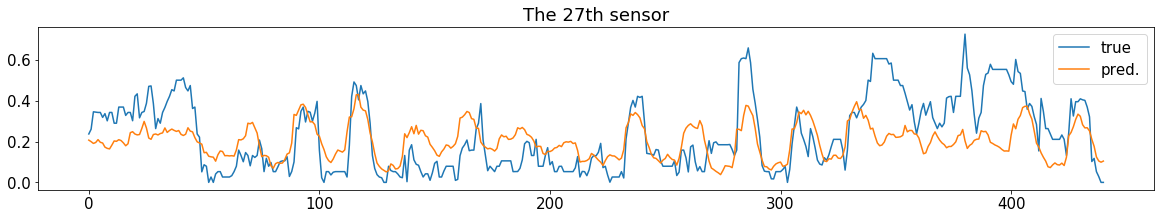}
    \caption{\small \textit{Reconstruction of selected sensors for kernel ridge regression ($H = 1, \lambda = 0.428, \gamma = 0.693$).}}
    \label{fig:recovered signal ker}
\end{figure}

\begin{figure}[!htb]
    \centering
    \includegraphics[width=0.8\textwidth]{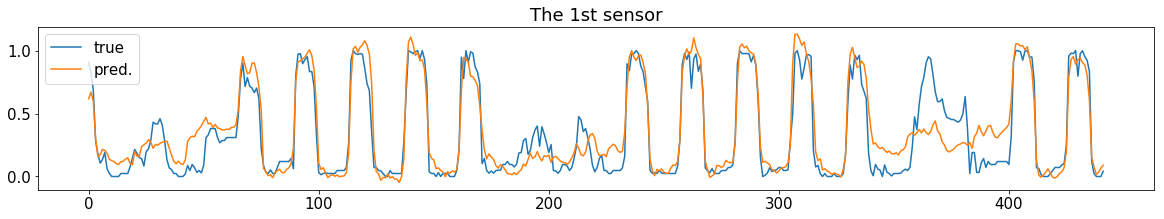}
    \includegraphics[width=0.8\textwidth]{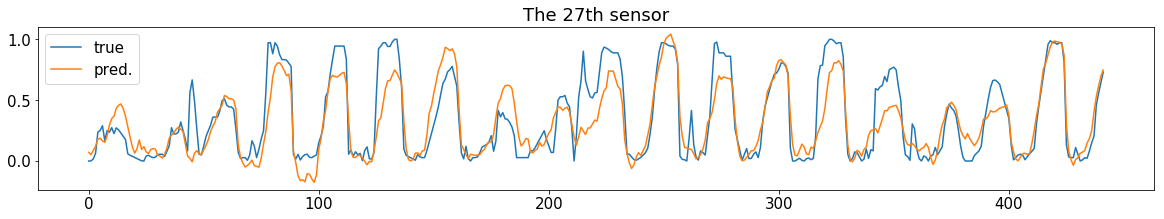}
    \caption{\small \textit{Reconstruction of selected sensors for ChebNet ($H = 0$, dropout method with $R^2$, run 2).}}
    \label{fig:recovered signal gcn}
\end{figure}

Figure \ref{fig:recovered signal lin}, \ref{fig:recovered signal ker} and \ref{fig:recovered signal gcn} show the recovered signals of the $1$st and $27$th (last one) selected sensors on the test set for the three reconstruction models. We can see that the two first selected sensors (vertices $5$ and $34$), are both meaningful and regular time series, whose original scales are larger than the $27$th selected sensors of linear ridge regression and kernel ridge regression, with the latter ones being noisier.

\begin{figure}[!htb]
    \centering
    \includegraphics[width=0.45\textwidth]{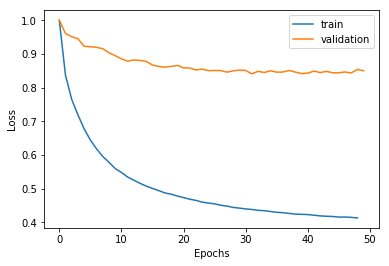}
    \includegraphics[width=0.45\textwidth]{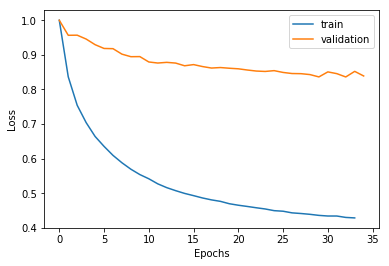}
    \caption{\small \textit{Training of selection net, dropout method, $H = 0$.} Left: run 1, stops at epoch $50$. Right: run 2, stops at epoch $35$. The number of missing set $I$'s explored in two trainings are $3750$ and $2625$, respectively.}
    \label{fig:training selection net}
\end{figure}

\begin{figure}[!htb]
    \centering
    \includegraphics[width=0.45\textwidth]{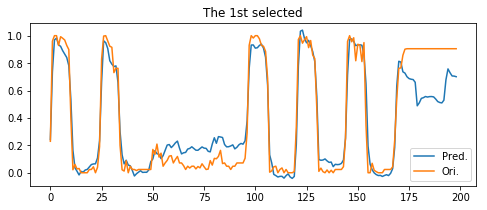}
    \includegraphics[width=0.45\textwidth]{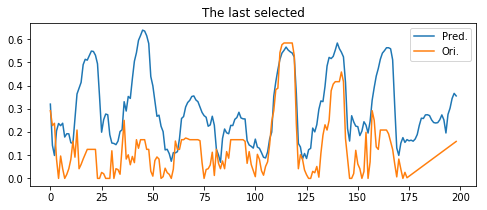}
    \includegraphics[width=0.45\textwidth]{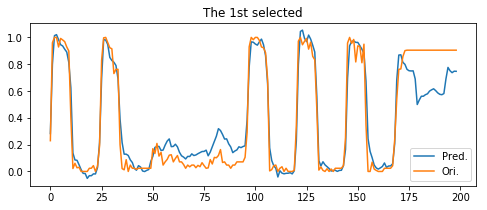}
    \includegraphics[width=0.45\textwidth]{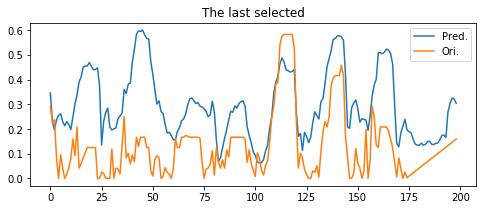}
    \caption{\small \textit{Testing of selection net, dropout method, $H = 0$.} Top: run 1. Bottom: run 2. Figures show the output of the first and last ($N$-th) selected sensors from trained selection net, testing on validation set, with the trend and scale added and multiplied back.}
    \label{fig:output of selection net}
\end{figure}

\begin{figure}[!htb]
    \centering
    \includegraphics[width=0.8\textwidth]{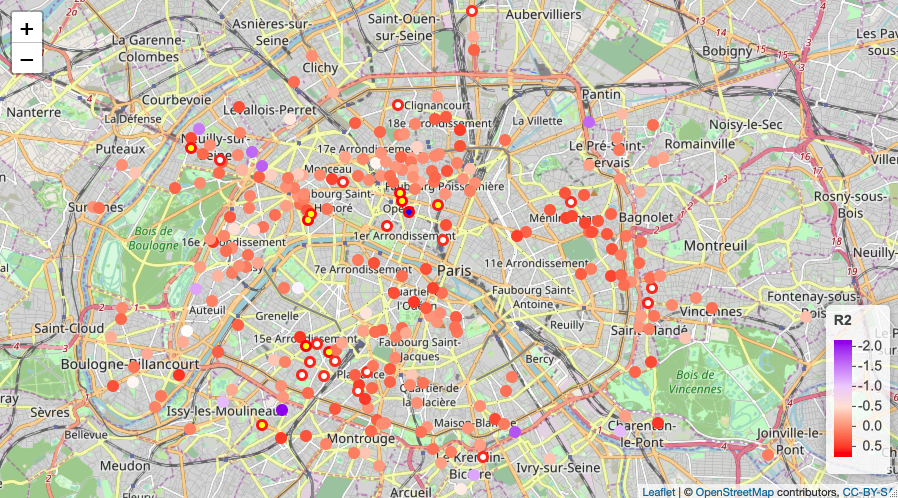}
    \caption{\small \textit{Representation of learned predictability in Paris network} using the $R^2$ scores of all sensors from GCN ($H = 0$, dropout method, run 2). The first $1$, $10$, and $27$ selected sensors with highest scores are in blue, yellow, and white respectively.}
    \label{fig:learned representation of predictability}
\end{figure}

We report the performance of the  training strategy to build a selection network described in Subsection \ref{sec:slcNetDP}. Figure \ref{fig:training selection net} shows training curves of the two runs of GCN, $H = 0$, in Table \ref{tab:reconstruction error paris}. 
For Paris dataset, which is cleaner, we did not use large batch for each model $I$'s learning, while employing a coarser search with a large learning rate. The two trainings finish after learning $3750$ and $2625$ prediction models respectively. The numbers of models explored are not tremendous considering the network size, which are however still infeasible in reasonable time for sufficient and independent training. On the other hand, Figure \ref{fig:output of selection net} shows the output of trained selection net $f'^{\GG}_{\Theta}$, testing on validation data without dropout. As an aggregated reconstruction model, even with insufficient training, it is already able to recover the signals of predictable sensors well, as shown in the subfigures corresponding to the first selected sensors. However, the prediction of last selected ($p$-th) sensor is bad in both runs, which proves the previous statement that the network will not learn to copy the input as output.
We visualize the learned  predictability of all sensors on map in Figure \ref{fig:learned representation of predictability} using the $R^2$ measure to perform scoring.



\begin{figure}[!htb]
    \centering
    \includegraphics[width=0.8\textwidth]{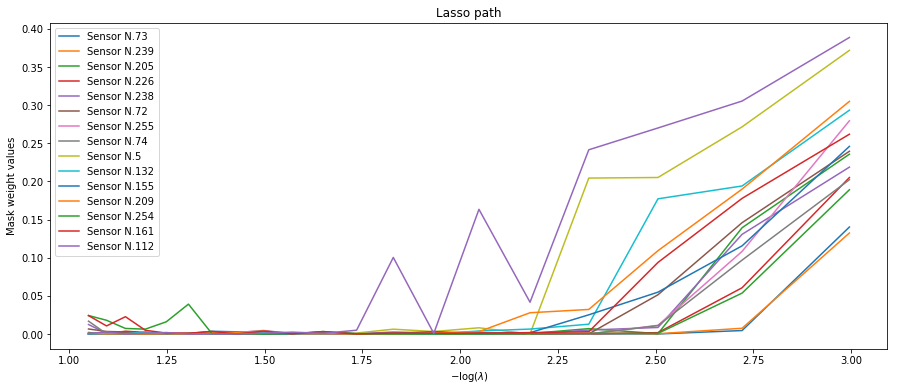}
    \caption{\small \textit{Lasso path of the first 15 selected sensors by masking with $\ell_1$ regularization reported in Table \ref{tab:selection results of GCN (MSE) and GCN (Masking)}}}
    \label{fig:lasso_path}
\end{figure}

Lastly, in Table \ref{tab:selection results of GCN (MSE) and GCN (Masking)}, we show the results of GCN, $H = 0$, transformed by dropout method using MSE, and by the masking with $\ell_1$ regularization. We can find that linear model, GCN (masking with $\ell_1$ regularization) and GCN (MSE) have a lot of overlap in the best turned-off set. For one thing, both linear model and GCN (masking with $\ell_1$ regularization) rely heavily on training set, for the other, MSE-based criteria can select more noisy signals compared to $R^2$ score. We will revisit these two reasons in Subsection \ref{sec:toulouse network}. 
Figure \ref{fig:lasso_path} shows the \textit{Lasso path} of masking with $\ell_1$ regularization.

\begin{minipage}[!htb]{\linewidth}
\bigskip
\captionof{table}{\small Sensor selection results of GCN, $H = 0$, dropout run 2 using MSE, and GCN, $H = 0$, masking with $\ell_1$ regularization, also their overlaps with the results of best models in Table \ref{tab:selection results of 3 best models}.} \label{tab:selection results of GCN (MSE) and GCN (Masking)} 
\begin{tabular}{ C{0.90in} C{1.30in} C{0.80in} *3{C{0.55in}}}
\toprule[1.5pt]
 & \bf \scriptsize Selected sensors (in order) & \bf \scriptsize Overlap w. linear & \bf \scriptsize Overlap w. kernel & \bf \scriptsize Overlap w. GCN ($R^2$) & \bf \scriptsize Overlap of the two \\
\hline
\bf \scriptsize GCN (MSE): 21.13 (29.65) & \scriptsize 155, 38, 132, 17, 61, 157, 164, 131, 98, 225, 87, 97, 112, 34, 159,  41, 151, 205, 103, 193, 186, 203, 72, 5, 204, 198, 58. & \tiny 34, 132, 5, 72, 205, 112, 17, 186, 155, 61. & \scriptsize 193, 34, 225, 5, 112, 186, 155. & \scriptsize 225, 34, 5, 112, 58, 61. & \multirow{2}{3em}{\scriptsize 34, 132, 5, 72, 205, 112, 17, 61, 155, 157. }\\\cline{2-2}
\bf \scriptsize GCN (mask): 24.95 (29.65) & \scriptsize 73, 239, 205, 226, 238, 72, 255, 74, 5, 132, 155, 209, 254, 161, 112, 227, 17, 224, 117, 30, 84, 61, 257, 233, 157, 34, 247. & \tiny 224, 257, 34, 226, 132, 5, 227, 72, 73, 74, 205, 239, 112, 17, 209, 117, 155, 61, 30, 255. & \scriptsize 34, 5, 74, 238, 112, 117, 155, 30. & \scriptsize 257, 34, 5, 112, 117, 61, 30. &\\
\bottomrule[1.25pt]
\end{tabular}\par
\end{minipage}

\subsubsection{Toulouse Network}\label{sec:toulouse network}

\begin{minipage}[h]{\linewidth}
\bigskip
\captionof{table}{\small Reconstruction error over the test set of the $18$ (10\% of $185$) selected sensors in Toulouse network. For each method, the values in parenthesis are the mean reconstruction error from $100$ random sampling sets $I$ of cardinality $p=18$.} \label{tab:reconstruction error toulouse} 
\begin{tabular}{ C{1.50in} *4{C{0.90in}}}\toprule[1.5pt]
$H$& $0$ & $1$ & $5$ & $10$ \\
\midrule
\bf Linear models &  32.33 (36.00)   & 29.95 (27.55)  &  30.61 (28.09) & 30.36 (28.41) \\
\bf Kernel models &  23.34 (28.40)  & 18.13 (20.59)  & 18.08 (20.62)  &  18.63 (20.46)\\
\bf GCNs ($R^2$, run $1$) &  16.72 (21.92) &  / &  17.74 (21.70) & / \\
\bf GCNs ($R^2$, run $2$) &  17.97 (21.92) & /  &  18.24 (21.70) & / \\
\bottomrule[1.25pt]
\end {tabular}\par
\bigskip
\end{minipage}

\begin{figure}[!htb]
    \centering
    \includegraphics[width=0.8\textwidth]{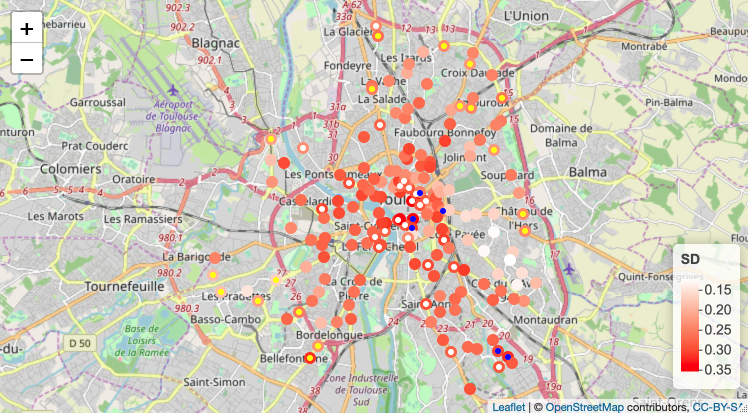}
    \caption{\small \textit{Display of the selected sensors from results in Table \ref{tab:selection results of 3 best models, toulouse}.} The selected sensors by linear model are marked by circles with inside yellow. The intersection and union of kernel and GCN ($R^2$) results are marked with blue and white, respectively.}
    \label{fig:sensor_selection_results_Toulouse}
\end{figure}

Table \ref{tab:reconstruction error toulouse} reports the  reconstruction error of three model classes. We can see that the errors of selected sensors do not decrease from the random sampling results as much as in Paris dataset, especially for the negative performance of linear models. This can be explained by the fact that the network time series from Toulouse dataset are less stationary after data preprocessing. 
We also compare the sensor selection results of the best models\footnote{The best models are determined by the reconstruction error on validation sets.} of three classes in Table \ref{tab:selection results of 3 best models, toulouse}, which are linear ridge regression ($H = 5, \lambda = 0.839$), kernel ridge regression ($H = 10, \lambda = 1.036, \gamma = 0.012$), ChebNet ($H = 0$, dropout method with $R^2$, run 1). While there are some sensors selected by the graph-based methods, kernel and GCN with dropout, the linear model returns completely different sensors. We visualize the selected sensors on a map in Figure \ref{fig:sensor_selection_results_Toulouse}. We observe that linear model selects sensors in suburb while kernel model and GCN tend to find those sensors in city center with higher standard deviation. Figure \ref{fig:recovered signal lin, toulouse}, \ref{fig:recovered signal ker, toulouse} and \ref{fig:recovered signal gcn, toulouse} show  recovered signals of the $1$st and $18$th selected sensors on the test set. 

We can see that,  even after a previous scaling of time series, the linear model cannot distinguish noise and regular signals very well. 
 For bike-sharing networks, a typical noisy signal appears at less used stations. We found that these signals are not well predictable, and they usually have a low standard deviation in the original scale as shown in Figure \ref{fig:output of selection net} and \ref{fig:output of selection net, toulouse}. 

\begin{minipage}[h]{\linewidth}
\bigskip
\captionof{table}{\small Sensor selection results of the best models in three classes of reconstruction functions. The sensors are indexed from $0$ to $184$ as the order of their columns in \textit{X\_hour\_toulouse.csv} file.}
\label{tab:selection results of 3 best models, toulouse} 
\begin{tabular}{ C{0.50in} C{1.80in} *3{C{0.80in}}} 
\toprule[1.5pt]
 & \bf \scriptsize Selected sensors (in order) & \bf \scriptsize Overlap of linear, kernel & \bf \scriptsize Overlap of kernel, GCN & \bf \scriptsize Overlap of linear, GCN \\
\hline
\bf Linear & \scriptsize 171, 170, 163, 177, 168, 137, 180, 165, 99, 135, 162, 103, 181, 169, 159, 164, 167, 175. &  & & \multirow{2}{4em}{} \multirow{3}{2em}{}\\ \cline{2-2}
\bf Kernel & \scriptsize 139, 6, 8, 140, 1, 2, 12, 26, 28, 13, 4, 138, 24, 45, 11, 34, 43, 86. &  & \multirow{2}{4em}{\small 6, 11, 140, 139, 24, 26. } & \multirow{2}{4em}{}\\ \cline{2-2}
\bf GCN & \scriptsize 11, 74, 26, 77, 139, 24, 72, 60, 140, 41, 0, 63, 151, 15, 147, 166, 64, 6. \\
\bottomrule[1.25pt]
\end{tabular}\par
\end{minipage}

\begin{figure}[!htb]
    \centering
    \includegraphics[width=0.8\textwidth]{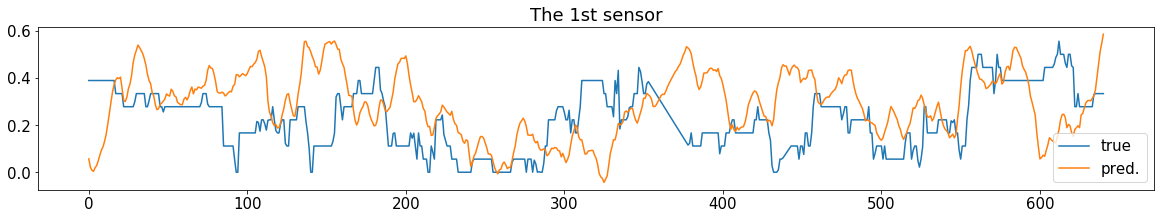}
    \includegraphics[width=0.8\textwidth]{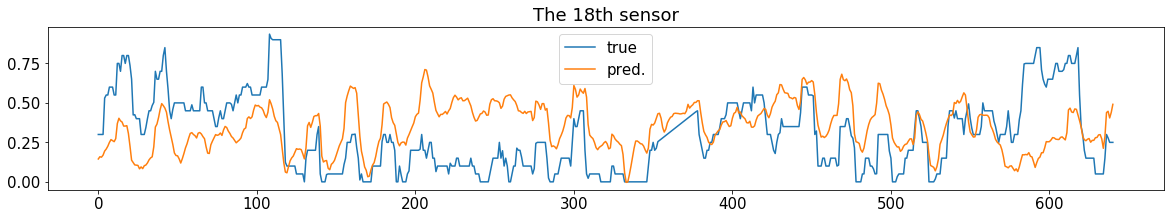}
    \caption{\small \textit{Reconstruction of selected sensors, linear ridge regression ($H = 5, \lambda = 0.839$).}}
    \label{fig:recovered signal lin, toulouse}
\end{figure}

\begin{figure}[!htb]
    \centering
    \includegraphics[width=0.8\textwidth]{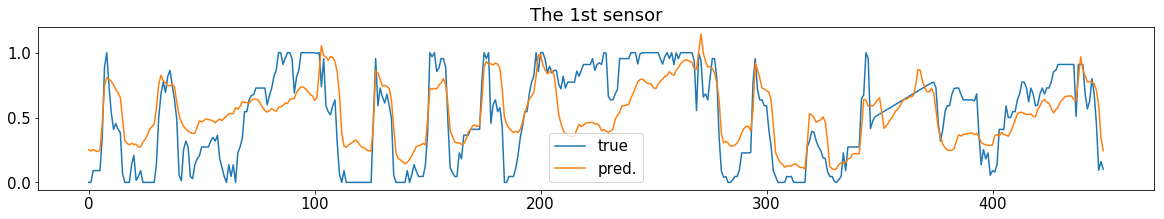}
    \includegraphics[width=0.8\textwidth]{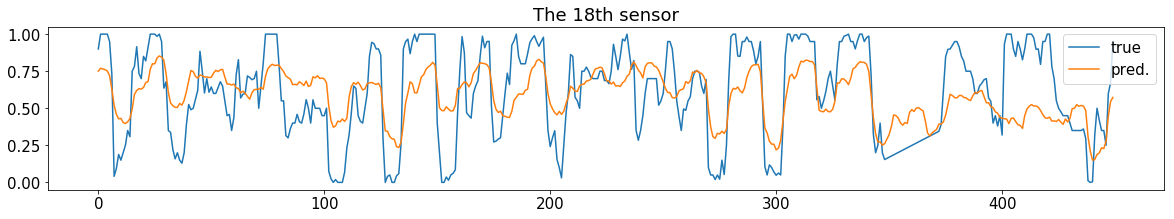}
    \caption{\small \textit{Reconstruction of selected sensors, kernel ridge regression ($H = 10, \lambda = 1.036, \gamma = 0.012$).}}
    \label{fig:recovered signal ker, toulouse}
\end{figure}

\begin{figure}[!htb]
    \centering
    \includegraphics[width=0.8\textwidth]{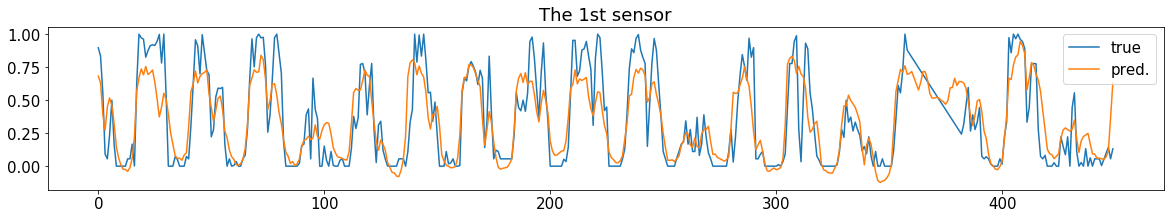}
    \includegraphics[width=0.8\textwidth]{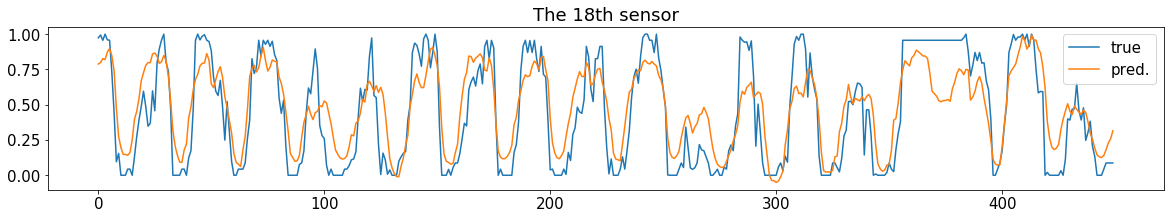}
    \caption{\small \textit{Reconstruction of selected sensors, ChebNet ($H = 0$, dropout method with $R^2$, run 1).}}
    \label{fig:recovered signal gcn, toulouse}
\end{figure}

Figure \ref{fig:output of selection net, toulouse} shows the outputs of trained selection net (dropout) of the best-predictable and worst-predictable sensors in Toulouse network. For this dataset, we employ a finer learning process with the larger batch size and much smaller learning rate. The training finishes at epoch $255$ with $4080$ missing sets searched. We also visualize the learned  predictability of the sensors on the map displayed in Figure \ref{fig:learned representation of predictability, toulouse}.

\begin{figure}[!htb]
    \centering
    \includegraphics[width=0.65\textwidth]{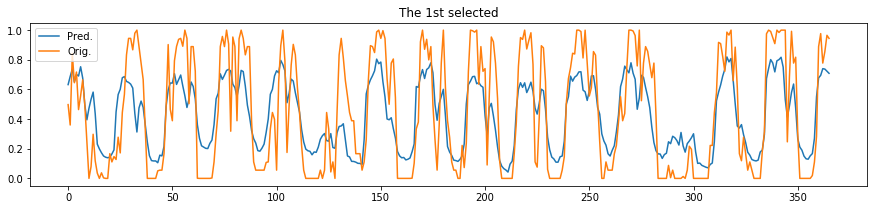}
    \includegraphics[width=0.65\textwidth]{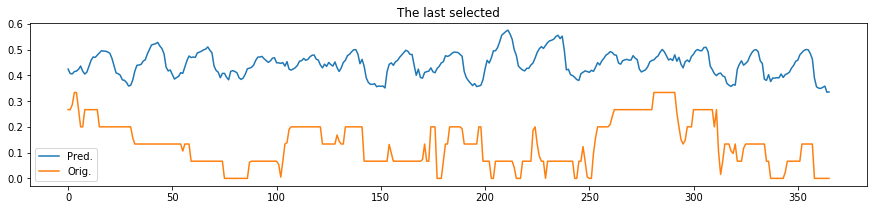}
    \caption{\small \textit{Testing of selection net, dropout method, $H = 0$, run 1.} Figures show the output of the first and last ($N$-th) selected sensors from trained selection net, testing on validation set. The number of missing set $I$'s explored in the training is $4080$.}
    \label{fig:output of selection net, toulouse}
\end{figure}

\begin{figure}[!htb]
    \centering
    \includegraphics[width=0.7\textwidth]{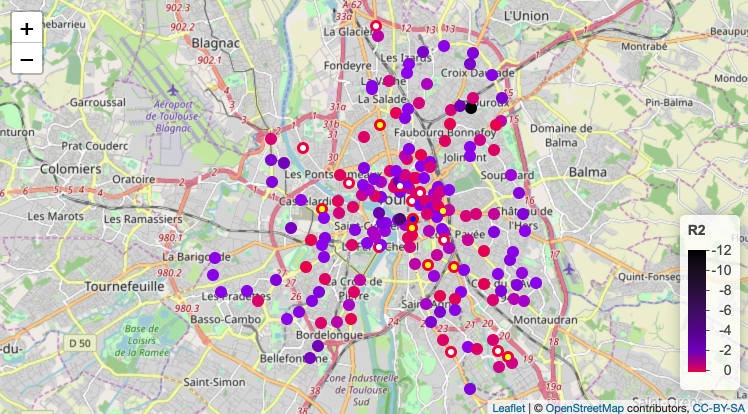}
    \caption{\small \textit{Representation of learned predictability in Toulouse's network} using the $R^2$ scores of all sensors from GCN ($H = 0$, dropout method, run 1). The first $1$, $8$, and $18$ selected sensors with highest scores are in blue, yellow, and white respectively.}
    \label{fig:learned representation of predictability, toulouse}
\end{figure}

\begin{figure}[!htb]
    \centering
    \includegraphics[width=0.8\textwidth]{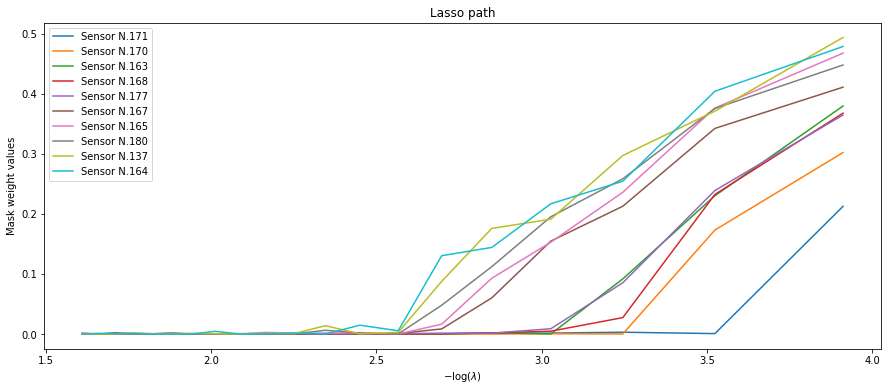}
    \caption{\small \textit{Lasso path of the first 10 selected sensors by GCN with masking and $\ell_1$ regularization reported in Table \ref{tab:selection results of GCN (MSE) and GCN (Masking), toulouse}}}
    \label{fig:lasso_path_toulouse}
\end{figure}

Lastly, we report the results of masking with $\ell_1$ regularization in Table \ref{tab:selection results of GCN (MSE) and GCN (Masking), toulouse}. It shows that linear model, GCN (masking with $\ell_1$ regularization) and GCN (MSE) still have the most overlap. On the other hand, compared across the reconstruction errors associated with GCN class from Table \ref{tab:reconstruction error paris}, \ref{tab:selection results of GCN (MSE) and GCN (Masking)}, \ref{tab:reconstruction error toulouse},  \ref{tab:selection results of GCN (MSE) and GCN (Masking), toulouse}, we can see that the masking transformation are not very effective, and the dropout method using $R^2$ score yields better results than using MSE. Figure \ref{fig:lasso_path_toulouse} shows the \textit{Lasso path} of masking with $\ell_1$ regularization.

\begin{minipage}[h]{\linewidth}
\bigskip
\captionof{table}{\small Sensor selection results of GCN, $H = 0$, dropout run 1 using MSE and GCN, $H = 0$, masking with $\ell_1$ regularization, also their overlaps with the results of best models in Table \ref{tab:selection results of 3 best models, toulouse}.} \label{tab:selection results of GCN (MSE) and GCN (Masking), toulouse} 
\begin{tabular}{C{0.90in} C{1.30in} C{0.70in} C{0.50in} C{0.70in} C{0.70in}} 
\toprule[1.5pt]
 & \bf \scriptsize Selected sensors (in order) & \bf \scriptsize Overlap w. linear & \bf \scriptsize Overlap w. kernel & \bf \scriptsize Overlap w. GCN ($R^2$) & \bf \scriptsize Overlap of the two \\
\hline
\bf \scriptsize GCN (MSE): 20.66 (21.92) & \scriptsize 163 ,177, 167, 166, 125, 184, 168, 120,  51,  66,  31, 129, 170, 124, 6, 139, 128, 143. & \scriptsize 163, 167, 168, 170, 177. & \scriptsize 139, 6. & \scriptsize 139, 166, 6. & \multirow{2}{3em}{\scriptsize 128, 163, 167, 168, 170, 177. }\\\cline{2-2}
\bf \scriptsize GCN (mask): 20.71 (21.92) & \scriptsize 171, 170, 163, 168, 177, 167, 165, 180, 137, 164, 127, 128, 162, 181, 99, 100, 104, 159. & \tiny 162, 99, 163, 164, 165, 167, 168, 137, 170, 171, 177, 180, 181, 159. & \scriptsize  & \scriptsize  & \\
\bottomrule[1.25pt]
\end{tabular}\par
\end{minipage}

\section{Conclusion}

In this paper, we propose several data-driven sensor selection strategies for three classes of reconstruction models. For linear class, we interpret the reconstruction errors and selection criteria through partial variance, and induce  greedy algorithms to select the best sampling set, which are related to the literature on this topic. We analyze the kernel class which is a generalization of the linear class. For graph convolutional neural network class, we propose two ways to transform a prediction network into a selection network using either $\ell_1$ regularization or the dropout technique. Especially, the proposed training strategy using dropout brings GCN the promising reconstruction performance in the numerical experiments. 

Lastly, we make Python codes (used to carry out the numerical experiments reported in this paper) available at \url{https://github.com/yiyej} for reproducibility.

\section*{Appendices}

\begin{pf}(\textbf{Proposition} \ref{prop:miniemperror})

\medskip

We rewrite problem (\ref{eq:recpbLin}) in a more concrete matrix form with the help of Frobenius Norm, which gives
\begin{equation}\label{empirical i.i.d. case S(I)}
    \FF(I,\hat{\Theta}(I)) = \mathop{\rm min}\limits_{\Theta_1} \frac{1}{T_0} \|\Tilde{\mathbf{x}}_{I} - \Tilde{\mathbf{x}}_{I^c}\Theta_1\|^2_F,
\end{equation}
where $\Tilde{\mathbf{x}}_{I} = \left[\,  \mathbf{x}_{I,1}, \mathbf{x}_{I,2}, ..., \mathbf{x}_{I,T_0} \, \right]^t \in \R^{T_0 \times p}$ and $\Tilde{\mathbf{x}}_{I^c} = \left[\, \mathbf{x}_{I^c,1}, \mathbf{x}_{I^c,2}, ..., \mathbf{x}_{I^c,T_0} \, \right]^t \in \R^{T_0 \times (N - p)}$. 
Through linear algebra, we know the minimizer $\Theta_1^* = (\Tilde{\mathbf{x}}_{I^c}^t\Tilde{\mathbf{x}}_{I^c})^{-1}\Tilde{\mathbf{x}}_{I^c}^t\Tilde{\mathbf{x}}_{I}$. Plug $\Theta_1^*$ into Equation (\ref{empirical i.i.d. case S(I)}), we have
$$
\begin{aligned}
    \FF(I,\hat{\Theta}(I)) &= \frac{1}{T_0} \|\Tilde{\mathbf{x}}_{I} - \Tilde{\mathbf{x}}_{I^c}\Theta_1^*\|^2_F \\
    &= \frac{1}{T_0} \|\Tilde{\mathbf{x}}_{I} - \Tilde{\mathbf{x}}_{I^c}[\Tilde{\mathbf{x}}_{I^c}^t\Tilde{\mathbf{x}}_{I^c}]^{-1}\Tilde{\mathbf{x}}_{I^c}^t\Tilde{\mathbf{x}}_{I}\|^2_F \\
    &\stackrel{ (a)}{=} \frac{1}{T_0} tr(\left[ \Tilde{\mathbf{x}}_{I} - \Tilde{\mathbf{x}}_{I^c}[\Tilde{\mathbf{x}}_{I^c}^t\Tilde{\mathbf{x}}_{I^c}]^{-1}\Tilde{\mathbf{x}}_{I^c}^t\Tilde{\mathbf{x}}_{I} \right]^t\left[ \Tilde{\mathbf{x}}_{I} - \Tilde{\mathbf{x}}_{I^c}[\Tilde{\mathbf{x}}_{I^c}^t\Tilde{\mathbf{x}}_{I^c}]^{-1}\Tilde{\mathbf{x}}_{I^c}^t\Tilde{\mathbf{x}}_{I} \right]) \\
    &= \frac{1}{T_0} tr(\Tilde{\mathbf{x}}_{I}^t\Tilde{\mathbf{x}}_{I} - \Tilde{\mathbf{x}}_{I}^t\Tilde{\mathbf{x}}_{I^c}[\Tilde{\mathbf{x}}_{I^c}^t\Tilde{\mathbf{x}}_{I^c}]^{-1}\Tilde{\mathbf{x}}_{I^c}^t\Tilde{\mathbf{x}}_{I}) \\
    &= tr(\frac{\Tilde{\mathbf{x}}_{I}^t\Tilde{\mathbf{x}}_{I}}{T_0} - \frac{\Tilde{\mathbf{x}}_{I}^t\Tilde{\mathbf{x}}_{I^c}}{T_0}[\frac{\Tilde{\mathbf{x}}_{I^c}^t\Tilde{\mathbf{x}}_{I^c}}{T_0}]^{-1}\frac{\Tilde{\mathbf{x}}_{I^c}^t\Tilde{\mathbf{x}}_{I}}{T_0}) = tr(\hat{\bSigma}_{I} -  \hat{\bSigma}_{II^c}\hat{\bSigma}_{
    I^cI^c}^{-1}\hat{\bSigma}_{I^cI}).
\end{aligned}
$$
where, Equation $(a)$ above is due to the fact that $\|A\|^2_F = tr(A^tA)$.

\qed
\end{pf}

\begin{pf}(\textbf{Proposition} \ref{prop:miniemperrorH>0})

\medskip

\begin{equation}\label{empirical ts case S(I)}
    \FF(I,\hat{\Theta}(I)) = \argmin_{\Theta_1} \frac{1}{T_0} \|\Tilde{\mathbf{x}}_{I}(0) - \Tilde{\mathbf{x}}_{I^c}\Theta_1\|^2_F,
\end{equation}
where $\Tilde{\mathbf{x}}_{I}(0) = (\, \mathbf{x}_{I,H + 1}, \mathbf{x}_{I,H + 2}, ..., \mathbf{x}_{I,T_0} \, ) \in \R^{(T_0 - H)\times p}$ and
$$
\begin{aligned}
\Tilde{\mathbf{x}}_{I^c} & = 
\begin{pmatrix}
    \mathbf{x}_{I^c,H + 1}^t & \mathbf{x}_{I^c,H}^t & \hdots & \mathbf{x}_{I^c,1}^t\\
    \mathbf{x}_{I^c,H + 2}^t & \mathbf{x}_{I^c,H + 1}^t & \hdots & \mathbf{x}_{I^c,2}^t\\ 
    \vdots & \vdots &  & \vdots\\
    \mathbf{x}_{I^c,T_0}^t & \mathbf{x}_{I^c,T_0 - 1}^t & \hdots & \mathbf{x}_{I^c,T_0 - H}^t
\end{pmatrix} \\
:&= \left[\, \Tilde{\mathbf{x}}_{I^c}(0),  \Tilde{\mathbf{x}}_{I^c}(-1), ..., \Tilde{\mathbf{x}}_{I^c}(-H) \, \right]  \in \R^{(T_0 - H)\times [(N-p)(H+1)]}.
\end{aligned}
$$

Through linear algebra, we know $\Theta_1^* = (\Tilde{\mathbf{x}}_{I^c}^t\Tilde{\mathbf{x}}_{I^c})^{-1}\Tilde{\mathbf{x}}_{I^c}^t\Tilde{\mathbf{x}}_{I}(0)$. Plug $\Theta_1^*$ into equation (\ref{empirical ts case S(I)}) have
$$
\begin{aligned}
    \FF(I,\hat{\Theta}(I)) & = \frac{1}{T_0} \|\Tilde{\mathbf{x}}_{I}(0) - \Tilde{\mathbf{x}}_{I^c}\Theta_1\|^2_F \\
    & = \frac{1}{T_0} \|\Tilde{\mathbf{x}}_{I}(0) - \Tilde{\mathbf{x}}_{I^c}[\Tilde{\mathbf{x}}_{I^c}^t\Tilde{\mathbf{x}}_{I^c}]^{-1}\Tilde{\mathbf{x}}_{I^c}^t\Tilde{\mathbf{x}}_{I}(0)\|^2_F \\
    & = \frac{1}{T_0} tr(\left[ \Tilde{\mathbf{x}}_{I}(0) - \Tilde{\mathbf{x}}_{I^c}[\Tilde{\mathbf{x}}_{I^c}^t\Tilde{\mathbf{x}}_{I^c}]^{-1}\Tilde{\mathbf{x}}_{I^c}^t\Tilde{\mathbf{x}}_{I}(0) \right]^t\left[ \Tilde{\mathbf{x}}_{I}(0) - \Tilde{\mathbf{x}}_{I^c}[\Tilde{\mathbf{x}}_{I^c}^t\Tilde{\mathbf{x}}_{I^c}]^{-1}\Tilde{\mathbf{x}}_{I^c}^t\Tilde{\mathbf{x}}_{I}(0) \right]) \\
    &= tr(\frac{\Tilde{\mathbf{x}}_{I}^t(0)\Tilde{\mathbf{x}}_{I}(0)}{T_0} - \frac{\Tilde{\mathbf{x}}_{I}^t(0)\Tilde{\mathbf{x}}_{I^c}}{T_0}[\frac{\Tilde{\mathbf{x}}_{I^c}^t\Tilde{\mathbf{x}}_{I^c}}{T_0}]^{-1}\frac{\Tilde{\mathbf{x}}_{I^c}^t\Tilde{\mathbf{x}}_{I}(0)}{T_0}) \\
    &= tr(\mathbf{\hat{\Sigma}}_{I} - [\hat{\beta}^H_{II^c}][\hat{\alpha}^H_{I^c}]^{-1}[\hat{\beta}^H_{II^c}]^t).
\end{aligned}
$$
where $\frac{\Tilde{\mathbf{x}}_{I}^t(0)\Tilde{\mathbf{x}}_{I^c}}{T_0} = \frac{1}{T_0}\left[\, \Tilde{\mathbf{x}}_{I}^t(0)\Tilde{\mathbf{x}}_{I^c}(0),  \Tilde{\mathbf{x}}_{I}^t(0)\Tilde{\mathbf{x}}_{I^c}(-1), ..., \Tilde{\mathbf{x}}_{I}^t(0)\Tilde{\mathbf{x}}_{I^c}(-H) \, \right] = \hat{\beta}^H_{II^c}$ and 
$$
\frac{\Tilde{\mathbf{x}}_{I^c}^t\Tilde{\mathbf{x}}_{I^c}}{T_0} =
\frac{1}{T_0}
\begin{pmatrix}
\Tilde{\mathbf{x}}_{I^c}^t(0)\Tilde{\mathbf{x}}_{I^c}(0) & \Tilde{\mathbf{x}}_{I^c}^t(0)\Tilde{\mathbf{x}}_{I^c}(-1) & \cdots & \Tilde{\mathbf{x}}_{I^c}^t(0)\Tilde{\mathbf{x}}_{I^c}(-H)\\
\Tilde{\mathbf{x}}_{I^c}^t(-1)\Tilde{\mathbf{x}}_{I^c}(0) & \Tilde{\mathbf{x}}_{I^c}^t(-1)\Tilde{\mathbf{x}}_{I^c}(-1) & \cdots & \Tilde{\mathbf{x}}_{I^c}^t(-1)\Tilde{\mathbf{x}}_{I^c}(-H)\\
\vdots & \vdots & \ddots & \vdots\\
\Tilde{\mathbf{x}}_{I^c}^t(-H)\Tilde{\mathbf{x}}_{I^c}(0) & \Tilde{\mathbf{x}}_{I^c}^t(-H)\Tilde{\mathbf{x}}_{I^c}(-1) & \cdots & \Tilde{\mathbf{x}}_{I^c}^t(-H)\Tilde{\mathbf{x}}_{I^c}(-H)
\end{pmatrix}
= \hat{\alpha}^H_{I^c}.
$$

\qed
\end{pf}

\begin{fact}\label{fact:prodkernel} (Products of kernels are kernels.)
Given $k_1$ on $\X_1$ and $k_2$ on $\X_2$, then the mapping $k_1k_2$ whose value is defined by
$$
k_1k_2\left[(x_1,y_1) , (x_2,y_2) \right] = k_1(x_1,x_2)k_2(y_1,y_2)
$$
is a kernel on $\X_1 \times \X_2$.
\end{fact}

The general proof see \cite{christmann2008support}, Lemma 4.6 p.114. Here, we give a proof from the Gram matrix's point of view. 

\begin{pf}(\textbf{Fact} \ref{fact:prodkernel})

\medskip

We show that $k_1 k_2$ validates two points in the kernel definition. 

\begin{enumerate}
    \item $k_1k_2$ is symmetric. Because $\forall \, x_1, x_2 \in \X_1$ and $\forall \, y_1, y_2 \in \X_2$, we have $k_1k_2\left[(x_1,y_1) , (x_2,y_2) \right] = k_1(x_1,x_2)k_2(y_1,y_2) = k_1(x_2,x_1)k_2(y_2,y_1) = k_1k_2\left[(x_2,y_2) , (x_1,y_1) \right]$
    \item Any finite Gram matrix is PSD. For $\forall \, (x_1, y_1), (x_2,y_2), ..., (x_n,x_n) \in \X_1 \times \X_2$, their Gram matrix K writes as
    $$
    \begin{pmatrix}
        k_1(x_1,x_1)k_2(y_1,y_1) & k_1(x_1,x_2)k_2(y_1,y_2) & \cdots & k_1(x_1,x_n)k_2(y_1,y_n)\\
        k_1(x_2,x_1)k_2(y_2,y_1) & k_1(x_2,x_2)k_2(y_2,y_2) & \cdots & k_1(x_2,x_n)k_2(y_2,y_n)\\
        \vdots & \vdots & \ddots & \vdots\\
        k_1(x_n,x_1)k_2(y_n,y_1) & k_1(x_n,x_2)k_2(y_n,y_2) & \cdots & k_1(x_n,x_n)k_2(y_n,y_n)
    \end{pmatrix},
    $$
    which equals to the following Hadamard product of matrices.
    $$
    \begin{pmatrix}
        k_1(x_1,x_1) & k_1(x_1,x_2) & \cdots & k_1(x_1,x_n)\\
        k_1(x_2,x_1) & k_1(x_2,x_2) & \cdots & k_1(x_2,x_n)\\
        \vdots & \vdots & \ddots & \vdots\\
        k_1(x_n,x_1) & k_1(x_n,x_2) & \cdots & k_1(x_n,x_n)
    \end{pmatrix}
    \begin{pmatrix}
        k_2(y_1,y_1) & k_2(y_1,y_2) & \cdots & k_2(y_1,y_n)\\
        k_2(y_2,y_1) & k_2(y_2,y_2) & \cdots & k_2(y_2,y_n)\\
        \vdots & \vdots & \ddots & \vdots\\
        k_2(y_n,y_1) & k_2(y_n,y_2) & \cdots & k_2(y_n,y_n)
    \end{pmatrix}
    $$
    $$
    := K_1 \odot K_2.
    $$
    Because $K_1$ and $K_2$ are Gram matrices of kernels $k_1$ and $k_2$ respectively, they are both PSD. From  \textit{Schur Product Theorem}, we know $K$ is also PSD.
\end{enumerate}

\qed
\end{pf}

\begin{fact}\label{prop:0bestlinearriddge} The best theoretical reconstruction function of linear ridge regression family $\hat{\mathbf{x}}_{I,t}(\lambda) = \hat{\bSigma}_{II^c}(\hat{\bSigma}_{I^c} + \lambda \Id)^{-1}\mathbf{x}_{I^c,t}$ is the least squares estimator $\hat{\mathbf{x}}_{I,t}(\lambda^*) = \hat{\bSigma}_{II^c}\hat{\bSigma}_{I^c}^{-1}\mathbf{x}_{I^c,t}$ from ordinary linear regression in Subsection \ref{sec:linH=0}, where $\lambda^* = 0$ brings the minimal reconstruction error.
\end{fact}
\begin{pf}
(\textbf{Fact} \ref{prop:0bestlinearriddge})

\medskip

When $I$ is fixed, the reconstruction error is
$$ 
\begin{aligned}
&\frac{1}{T_0} \sum\limits_{t = 1}^{T_0} \|\mathbf{x}_{I,t} -  K_{II^c}(K_{I^c} + \lambda \Id)^{-1}\mathbf{x}_{I^c,t} \|_{\ell_2}^2 \\
= \quad & tr(\hat{\bSigma}_{I} - 2\hat{\bSigma}_{II^c}(K_{I^c} + \lambda \Id)^{-1}K_{I^cI} + K_{II^c}(K_{I^c} + \lambda \Id)^{-1}\hat{\bSigma}_{I^c}(K_{I^c} + \lambda \Id)^{-1}K_{I^cI})
\end{aligned}
$$
which is a real-valued function of $\lambda$. Thus we only need to calculate the derivative, which is
$$
tr(2\hat{\bSigma}_{II^c}(K_{I^c} + \lambda \Id)^{-2}K_{I^cI} - 2K_{II^c}(K_{I^c} + \lambda \Id)^{-2}\hat{\bSigma}_{I^c}(K_{I^c} + \lambda \Id)^{-1}K_{I^cI}),
$$
where $\cdot^{-m}$ denotes $\left[\cdot^{-1}\right]^m$. When the kernel is the linear one, $K_{I^c} = \hat{\bSigma}_{I^c}$ and $K_{II^c} = \hat{\bSigma}_{II^c}$. Then the derivative simplifies to 
$$
\begin{aligned}
& tr(2\hat{\bSigma}_{II^c}(\hat{\bSigma}_{I^c} + \lambda \Id)^{-2}\hat{\bSigma}_{I^cI} - 2\hat{\bSigma}_{II^c}(\hat{\bSigma}_{I^c} + \lambda \Id)^{-2}\hat{\bSigma}_{I^c}(\hat{\bSigma}_{I^c} + \lambda \Id)^{-1}\hat{\bSigma}_{I^cI}) \\
= \quad & tr(2\hat{\bSigma}_{II^c}\left[ (\hat{\bSigma}_{I^c} + \lambda \Id)^{-2} - (\hat{\bSigma}_{I^c} + \lambda \Id)^{-2}\hat{\bSigma}_{I^c}(\hat{\bSigma}_{I^c} + \lambda \Id)^{-1} \right] \hat{\bSigma}_{I^cI}).
\end{aligned}
$$
We denote the eigendecomposition of PD matrix $\hat{\bSigma}_{I^c}$ as $\hat{\bSigma}_{I^c} = \mathbf{Q}\Delta\mathbf{Q}^t$. Then the previous trace writes as
\begin{equation}\label{eq_appendix:derivative}
\begin{aligned}
& tr(2\hat{\bSigma}_{II^c}\mathbf{Q}\left[(\Delta + \lambda\Id)^{-2} - \Delta(\Delta + \lambda\Id)^{-3}\right]\mathbf{Q}^t \hat{\bSigma}_{I^cI}) \\
\quad & = tr(2\hat{\bSigma}_{II^c}\mathbf{Q}\left[\lambda(\Delta + \lambda\Id)^{-3} \right]\mathbf{Q}^t \hat{\bSigma}_{I^cI}) \\
\quad & = \lambda tr(2\hat{\bSigma}_{II^c}\mathbf{Q}(\Delta + \lambda\Id)^{-3} \mathbf{Q}^t \hat{\bSigma}_{I^cI}).
\end{aligned}
\end{equation}
Note that $2\hat{\bSigma}_{II^c}\mathbf{Q}(\Delta + \lambda\Id)^{-3} \mathbf{Q}^t \hat{\bSigma}_{I^cI}$ is a PD matrix. Thus when $\lambda \geq 0$, derivative (\ref{eq_appendix:derivative}) is sitrictly positive except $0$. Hence, the reconstruction error in linear ridge regression is monotone increasing on $[0, + \infty)$, where it has the unique minimizer $0$.

\qed
\end{pf}

\begin{pf} (\textbf{Proposition} \ref{prop:bastlambda}) 

\medskip

When $I$ is fixed, the reconstruction error is
$$
\begin{aligned}
& \frac{1}{T_0} \sum\limits_{t = H+1}^{T_0} \|\mathbf{x}_{I,t} -  K^H_{II^c}(K_{I^c}^H + \lambda \Id)^{-1}\mathbf{x}_{I^c,t}^H \|_{\ell_2}^2 \\
= \quad & tr(\hat{\bSigma}_{I} - 2\hat{\mathbf{\beta}}^H_{II^c}(K_{I^c}^H + \lambda \Id)^{-1}[K^H_{II^c}]^t + K^H_{II^c}(K_{I^c}^H + \lambda \Id)^{-1}\hat{\mathbf{\alpha}}_{I^c}^H(K_{I^c}^H + \lambda \Id)^{-1}[K^H_{II^c}]^t).
\end{aligned}
$$
which is a real-valued function of $\lambda$. Its derivative w.r.t. to $\lambda$ is
$$
tr(2\hat{\mathbf{\beta}}^H_{II^c}(K_{I^c}^H + \lambda \Id)^{-2}[K^H_{II^c}]^t - 2K_{II^c}^H(K_{I^c}^H + \lambda \Id)^{-2}\hat{\mathbf{\alpha}}^H_{I^c}(K_{I^c}^H + \lambda \Id)^{-1}[K^H_{II^c}]^t).
$$
where $\cdot^{-m}$ denotes $\left[\cdot^{-1}\right]^m$. When the kernel is taken as the sample autocovariance, $K^H_{I^c} = \hat{\mathbf{\alpha}}_{I^c}^H$ and  $K^H_{II^c} = \hat{\mathbf{\beta}}_{II^c}^H$. Then the derivative simplifies to 
$$
\begin{aligned}
& tr(2\hat{\mathbf{\beta}}^H_{II^c}(\hat{\mathbf{\alpha}}_{I^c}^H + \lambda \Id)^{-2}[\hat{\mathbf{\beta}}^H_{II^c}]^t - 2\hat{\mathbf{\beta}}_{II^c}^H(\hat{\mathbf{\alpha}}_{I^c}^H + \lambda \Id)^{-2}\hat{\mathbf{\alpha}}^H_{I^c}(\hat{\mathbf{\alpha}}_{I^c}^H + \lambda \Id)^{-1}[\hat{\mathbf{\beta}}^H_{II^c}]^t) \\
 = \quad & tr(2\hat{\mathbf{\beta}}^H_{II^c}\left[ (\hat{\mathbf{\alpha}}_{I^c}^H + \lambda \Id)^{-2} - (\hat{\mathbf{\alpha}}_{I^c}^H + \lambda \Id)^{-2}\hat{\mathbf{\alpha}}^H_{I^c}(\hat{\mathbf{\alpha}}_{I^c}^H + \lambda \Id)^{-1} \right] [\hat{\mathbf{\beta}}^H_{II^c}]^t) 
\end{aligned}
$$
We denote the eigendecomposition of the positive definite (PD) matrix $\hat{\mathbf{\alpha}}^H$ as $\hat{\mathbf{\alpha}}^H = \mathbf{Q}\Delta\mathbf{Q}^t$. Then the previous trace writes as
\begin{equation}\label{eq_appendix:derivativeH>0}
\lambda tr(2\hat{\mathbf{\beta}}^H_{II^c}\mathbf{Q}(\Delta + \lambda\Id)^{-3} \mathbf{Q}^t [\hat{\mathbf{\beta}}^H_{II^c}]^t).
\end{equation}
Since $\hat{\mathbf{\beta}}^H_{II^c}\mathbf{Q}(\Delta + \lambda\Id)^{-3} \mathbf{Q}^t [\hat{\mathbf{\beta}}^H_{II^c}]^t$ is a PD matrix for any $\lambda \geq 0$, the derivative (\ref{eq_appendix:derivativeH>0}) is strictly positive except for $\lambda = 0$. Therefore, the reconstruction error is monotone increasing on $[0, + \infty)$, and its unique minimizer is $\lambda = 0$.

\qed

\end{pf}

\noindent
\textbf{List of selected sensors and the hyperparameter values used in Table \ref{tab:reconstruction error paris}}. 

\begin{itemize}
    \item Linear, $H = 0$: 5, 209, 73, 239, 155, 112, 74, 132, 30, 224, 226, 238, 229, 205, 255, 260, 117, 72, 227, 17, 34, 257, 114, 61, 42, 31, 254.
    
    \item Linear, $H = 1 (\lambda = 0.740)$: 5, 209, 112, 30, 73, 61, 224, 155, 117, 239, 257, 34, 229, 31, 42, 114, 240, 205, 74, 227, 186, 17, 72, 255, 113, 132, 226.
    
    \item Linear, $H = 5 (\lambda = 2.030)$: 5, 209, 112, 73, 30, 239, 155, 61, 74, 224, 117, 227, 17, 42, 34, 72, 226, 31, 229, 255, 257, 205, 114, 248, 84, 33, 238.

    \item Linear, $H = 10 (\lambda = 3.199)$: 73, 5, 209, 239, 112, 227, 74, 30, 155, 61, 224, 17, 72, 226, 117, 205, 34, 33, 255, 42, 238, 229, 248, 254, 260, 199, 31.

    \item Kernel, $H = 0 (\lambda = 0.203)$: 34, 197, 186, 3, 112, 229, 42, 74, 5, 189, 224, 113, 20, 190, 238, 30, 232, 114, 11, 70, 18, 228, 267, 193, 2, 93, 225.
    
    \item Kernel, $H = 1 (\lambda = 0.428, \gamma = 0.693)$: 34, 197, 186, 112, 229, 42, 74, 189, 3, 113, 225, 232, 193, 5, 20, 238, 114, 30, 11, 190, 70, 18, 228, 267, 117, 93, 155.
    
    \item Kernel, $H = 5 (\lambda = 1.917, \gamma = 0.028)$: 197, 34, 112, 193, 42, 74, 229, 117, 190, 232, 238, 189, 225, 113, 20, 11, 186, 114, 70, 33, 3, 256, 228, 93, 72, 18, 155.
    
    \item Kernel, $H = 10 (\lambda = 3.581, \gamma = 0.007)$: 197, 42, 112, 74, 193, 34, 117, 227, 238, 190, 229, 189, 110, 232, 256, 114, 186, 72, 33, 11, 70, 155, 113, 97, 18, 20, 93.
    
    \item GCN, dropout method, $H = 0$, run 1
    
    \textit{$R^2$:} 61, 34,  5, 229, 112, 49, 110, 2, 114, 108, 225, 189, 58, 221, 42, 209, 117, 71, 30, 220, 113, 266, 102, 92, 12, 169, 3.
        
    \textit{MSE:} 157, 155 , 38 , 61,  17, 164, 159 ,132 , 97, 131 , 98 ,193 ,204, 186 , 34 , 87 ,103, 209 ,71 ,185 ,220, 110, 184 ,245, 225 , 72, 112.
    
    \item GCN, dropout method, $H = 0$, run 2
    
    \textit{$R^2$:} 5, 34,  61, 112, 58, 108, 225, 221, 49, 42, 229, 3, 257, 117, 113, 169, 110, 266, 240, 30, 109, 2, 189, 95, 114, 248, 92.
        
    \textit{MSE:} 155, 38, 132, 17, 61, 157, 164, 131, 98, 225, 87, 97, 112, 34, 159,  41, 151, 205, 103, 193, 186, 203, 72, 5, 204, 198, 58.
    
    \item GCN, dropout method, $H = 5$, run 1
        
    \textit{$R^2$:} 61, 5, 34, 229, 225, 221, 112, 114,  58, 110, 3, 266, 2, 257, 209, 108, 49, 117, 30, 113, 42, 92, 20, 95, 169, 189, 102.
        
    \textit{MSE:} 97, 61, 17, 131, 155, 38, 225, 164, 193, 132, 209, 34, 204, 103, 157, 198, 110, 98, 112, 5, 252, 245, 41, 111, 126, 266, 226.
    
    \item GCN, dropout method with, $H = 5$, run 2
    
    \textit{$R^2$:} 5, 61, 229, 34, 112, 92, 225, 3, 30, 108, 58, 136, 266, 117, 20, 169, 189, 49, 2, 113, 110, 209, 257, 114, 42, 109, 31.
    
    \textit{MSE:} 97, 17, 164, 38, 155, 61, 98, 157, 204, 225, 41, 103, 207, 131, 87, 252, 132, 198, 212, 126, 209, 184, 185, 5, 112, 151, 193.

    \item GCN, masking method, $H = 0$
        
    73, 239, 205, 226, 238, 72, 255, 74, 5, 132, 155, 209, 254, 161, 112, 227, 17, 224, 117, 30, 84, 61, 257, 233, 157, 34, 247.
    
\end{itemize}

\noindent
\textbf{List of selected sensors and the hyperparameter values used in Table \ref{tab:reconstruction error toulouse}}. 

\begin{itemize}
    \item Linear, $H = 0$: 171, 163, 168, 177, 137, 170, 180, 167, 181, 99, 135, 162, 159, 165, 104, 164, 169, 97.
    
    \item Linear, $H = 1 (\lambda = 0.302)$: 171, 163, 177, 170, 168, 137, 180, 165, 99, 135, 162, 103, 181, 140, 159, 169, 167, 175.
    
    \item Linear, $H = 5 (\lambda = 0.839)$: 171, 170, 163, 177, 168, 137, 180, 165, 99, 135, 162, 103, 181, 169, 159, 164, 167, 175.

    \item Linear, $H = 10 (\lambda = 1.374)$: 171, 170, 168, 163, 177, 137, 180, 99, 165, 162, 103, 135, 181, 164, 169, 167, 159, 119.

    \item Kernel, $H = 0 (\lambda = 0.011)$: 46, 0, 98, 6, 24, 139, 11, 114, 71, 8, 4, 140, 28, 116, 15, 137, 142, 25.
    
    \item Kernel, $H = 1 (\lambda = 0.149, \gamma = 1.204)$: 139, 6, 8, 1, 140, 26, 2, 28, 0, 13, 138, 4, 11, 12, 7, 45, 24, 34.
    
    \item Kernel, $H = 5 (\lambda = 0.549, \gamma = 0.048)$: 139, 6, 8, 1, 140, 2, 26, 28, 13, 4, 138, 11, 12, 0, 45, 24, 7, 34.
    
    \item Kernel, $H = 10 (\lambda = 1.036, \gamma = 0.012)$: 139, 6, 8, 140, 1, 2, 12, 26, 28, 13, 4, 138, 24, 45, 11, 34, 43, 86.
    
    \item GCN, dropout method, $H = 0$, run 1
    
    \textit{$R^2$:} 11,  74,  26,  77, 139,  24,  72,  60, 140,  41,   0,  63, 151, 15, 147, 166,  64,   6.
        
    \textit{MSE:} 163, 177, 167, 166, 125, 184, 168, 120,  51,  66,  31, 129, 170, 124, 6, 139, 128, 143.
    
    \item GCN, dropout method, $H = 0$, run 2
    
    \textit{$R^2$:} 4, 8, 85, 11, 41, 43, 107, 151, 26, 69, 166, 91, 116, 102, 134, 28, 35, 24.
        
    \textit{MSE:} 163, 167, 166, 177, 125, 184, 66, 168, 120, 128, 31, 51, 170, 129, 6, 143, 75, 87.
    
    \item GCN, dropout method, $H = 5$, run 1
        
    \textit{$R^2$:} 50,  56,   6,  74,  28,   4, 146, 127,  17,  69, 107,  60, 106, 166, 133, 141,  26,   7.
        
    \textit{MSE:} 163, 177, 167, 125, 166, 184, 168,  66, 120,  31, 128,  67,  51, 6, 173,  50,  53,  87.
    
    \item GCN, dropout method, $H = 5$, run 2
    
    \textit{$R^2$:} 24, 43, 1, 69, 41, 107, 39, 98, 27, 6, 40, 80, 85, 11, 4, 26, 150, 74.
    
    \textit{MSE:} 163, 167, 177, 184, 166, 168, 125, 31, 66, 51, 120, 129, 87, 6, 173, 180, 139, 136.
    
    \item GCN, masking method, $H = 0$
        
    171, 170, 163, 168, 177, 167, 165,180, 137, 164, 127, 128, 162, 181, 99, 100, 104, 159.

\end{itemize}

\bibliographystyle{acm}
\bibliography{sensor_selection}

\end{document}